\documentclass{article}

\usepackage{microtype}
\usepackage{graphicx}
\usepackage{subcaption}
\usepackage{booktabs} % for professional tables

\usepackage{hyperref}

\usepackage[preprint]{icml2026}

\usepackage{amsmath}
\usepackage{amssymb}
\usepackage{mathtools}
\usepackage{amsthm}

\usepackage[capitalize,noabbrev]{cleveref}

\theoremstyle{plain}

\theoremstyle{definition}

\theoremstyle{remark}

\usepackage[textsize=tiny]{todonotes}

\icmltitlerunning{Preprint. Work in progress / Under review.}

\begin{document}

\twocolumn[
  \icmltitle{How Emotion Shapes the Behavior of LLMs and Agents: \\A Mechanistic Study}



  \icmlsetsymbol{equal}{*}

  \begin{icmlauthorlist}
    \icmlauthor{Moran Sun}{SCSE}
    \icmlauthor{Tianlin Li}{SCSE}
    \icmlauthor{Yuwei Zheng}{SCSE}
    \icmlauthor{Zhenhong Zhou}{CCDS}
    \icmlauthor{Aishan Liu}{SCSE}
    \icmlauthor{Xianglong Liu}{SCSE}
    \icmlauthor{Yang Liu}{CCDS}
  \end{icmlauthorlist}

  \icmlaffiliation{SCSE}{School of Computer Science and Engineering, Beihang University, Beijing, China}
  \icmlaffiliation{CCDS}{College of Computing and Data Science, Nanyang Technological University, Singapore}

  \icmlcorrespondingauthor{Tianlin Li}{tianlin001@buaa.edu.cn}

  \icmlkeywords{Emotion-aware Modeling, Large Language Model, LLM-based Agent, Representation-Level Steering, Mechanistic Interpretability, Sparse Autoencoder}

  \vskip 0.3in
]



% Use ONE off the following lines. DO NOT remove the command.
\printAffiliationsAndNotice{}  % no special notice (required even if empty)

\begin{abstract}
  Emotion plays an important role in human cognition and performance. Motivated by this, we investigate whether analogous emotional signals can shape the behavior of large language models (LLMs) and agents. Existing emotion-aware studies mainly treat emotion as a surface-level style factor or a perception target, overlooking its mechanistic role in task processing. To address this limitation, we propose \textbf{E-STEER}, an interpretable emotion steering framework that enables direct representation-level intervention in LLMs and agents. It embeds emotion as a structured, controllable variable in hidden states, and with it, we examine the impact of emotion on objective reasoning, subjective generation, safety, and multi-step agent behaviors. The results reveal non-monotonic emotion-behavior relations consistent with established psychological theories, and show that specific emotions not only enhance LLM capability but also improve safety, and systematically shape multi-step agent behaviors.
\end{abstract}

\section{Introduction}
Emotion plays a central role in human decision-making, creativity, and social interaction. Psychological studies show that positive affect enhances creative thinking \cite{affect-as-information}, excessive arousal can impair performance \cite{yerkes-dodson}, and feelings of insecurity may lead to conflicted decisions \cite{self-consistency}. These findings underscore the significant influence of emotion on human cognition and behavior.
Currently, Large Language Models (LLMs) are mainly trained on human-generated corpora \cite{human-cor} and incorporate many designs that imitate human thinking \cite{cot, few-shot, con-win}. Considering emotion as an important human mechanism, we aim to explore whether intelligent models inherit similar emotion-regulation signals. Thus, questions arise: \textit{What is the mapping between human emotion and model “emotion”? And can this “emotion” help improve model performance?}

\begin{figure}[tbp]
    \centering
    \includegraphics[width=\linewidth]{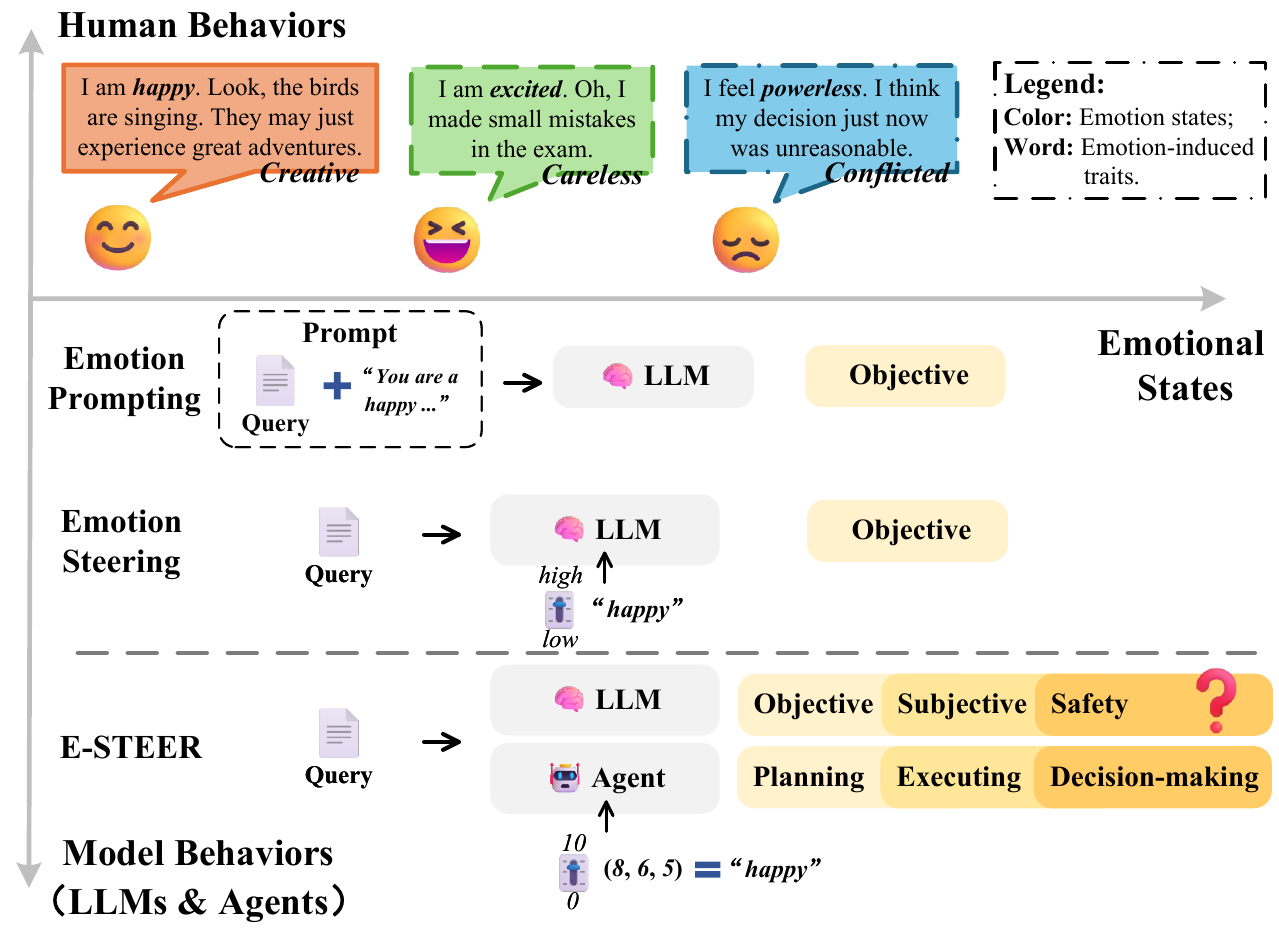}
    \caption{Emotion greatly affects human behaviors. How can emotion influence the behaviors of LLMs and Agents?}
    \label{fig: motivation}
\end{figure} 

Existing studies on the influence of emotion in LLMs can be broadly categorized into emotion prompting and emotion steering. Emotion prompting explicitly adds emotion labels or descriptions (e.g., "You are happy today, ...") into the input prompts, guiding the model to generate text under a specified emotional context \cite{emo-prompt}. Such studies are typically evaluated on text generation tasks, such as emotional dialogue \cite{llm-dia}. Some work has further explored encoding emotion as continuous numerical values within prompts, but empirical results show that this formulation is often less effective than using textual emotion labels \cite{label-coord}.
In contrast, emotion steering focuses on intervening in the hidden representations of LLM to control emotion, thereby directly influencing the internal reasoning process \cite{steering}. This allows emotional expression to be regulated at the representation-level rather than relying solely on surface-level prompts. Previous work has demonstrated that steering techniques can enhance specific emotion-related representations in LLMs and induce strong emotional characteristics in the generated text \cite{emo-steer}.

While these approaches advance emotional control in LLMs, they leave several challenges unresolved. Prompt-level approaches depend on implicit inference of affective type and intensity from natural language, leading to imprecise modulation and limited numerical sensitivity. As a result, these methods are largely restricted to discrete emotion labels. 
Emotion steering directly intervenes in LLM hidden states, allowing more continuous and fine-grained emotional control. However, existing steering methods still exhibit notable limitations. Most techniques typically focus on a limited set of basic emotions, without mapping to the complete continuous affective space. For example, Chebrolu \textit{et al.} proposed a steering method that enhances emotion-specific traits in model outputs, covering six emotions (e.g., trust and sadness) \cite{cur-steer-weak}. Additionally, previous work primarily assesses subjective text generation, leaving broader areas such as reasoning, safety, and agent behavior largely unexplored.

To address the above challenges, we propose E-STEER, an interpretable emotion steering framework for LLMs and agents. Instead of relying on discrete emotion labels, E-STEER represents emotion in the continuous Valence–Arousal–Dominance (VAD) space \cite{vad}. This decomposes emotion into three mutually independent dimensions, enabling more fine-grained and quantifiable emotional modeling. The framework employs Sparse Autoencoders (SAEs) to obtain an interpretable sparse representation of hidden states \cite{sae}, enabling direct identification and steering on VAD-based emotion features. This allows emotion to influence internal reasoning via representation-level control, overcoming the numeric insensitivity of prompt-level approaches. At the same time, the orthogonal structure of VAD enables it to simultaneously steer along the independent latent dimensions, allowing fine-grained multi-dimensional emotion control. 
Based on E-STEER, we systematically conduct evaluations of LLMs and agents across diverse tasks under controlled affective conditions. The results reveal that emotion reshapes model behavior: different emotional states induce significant differences in reasoning activity, accuracy, and safety. Notably, in multi-step agent settings, emotional biases accumulate along decision chains and substantially affect the outcomes.

The contributions of this paper are summarized as follows:
\begin{itemize}
    \item We propose E-STEER, an interpretable emotion steering framework that enables targeted and continuous modulation of emotion-related behaviors in LLMs.
    \item We introduce VAD theory and decompose emotion into three orthogonal dimensions, enabling multidimensional SAE-based steering over the complete emotion space.
    \item We conduct systematic experiments across four categories of LLM and agent behaviors to analyze how emotion modulates model behavior.
\end{itemize}

\section{Related Work}
\subsection{Human-inspired Mechanisms in LLMs}
LLMs incorporate various human-inspired mechanisms to enhance abilities. Chain-of-Thought (CoT) prompting provides models with reasoning examples, guiding them to generate more coherent intermediate reasoning and thereby improving multi-step problem-solving \cite{cot}. Few-shot prompting allows models to learn task patterns from a small number of examples, similar to sample problems in textbooks \cite{few-shot}. Extended context windows draw inspiration from human memory, enabling models to access longer histories and support coherent multi-turn reasoning \cite{con-win}. Persona vector ensures the behavior of the model in interaction is consistent with the specific identity \cite{persona}.

\subsection{Functional Roles of Emotion in AI Systems}
The primary application of emotion in AI has been in emotion recognition. Models should identify both the type and intensity of emotions present in the input \cite{text-emo-anl}. Researchers have further incorporated psychological theories such as VAD, which quantify emotion in continuous, measurable dimensions, enabling more fine-grained modeling \cite{dl-vad-anl}. These recognitions typically serve downstream applications, such as emotion-aware dialogue systems \cite{llm-dia}. On the other hand, studies examining the influence of emotion on LLM behavior have largely focused on subjective generation tasks \cite{llm-story, llm-nego}, with limited exploration of reasoning or safety. Moreover, most existing approaches operate at the prompt-level and rely on discrete emotional labels. Recent work indicates that LLMs are often insensitive to continuous values in prompts, such as VAD coordinates \cite{label-coord}. Agent is a kind of emerging LLM application \cite{react, langchain, autogen}. But in agent frameworks, LLMs are invoked multiple times, making agents more susceptible to external affective influences than single-step LLM generation \cite{inf-agent}.

\subsection{Steering for Fine-Grained Control of LLMs}
To address this insensitivity to numerical values in prompts, researchers proposed steering methods. Steering injects specific vectors into the model's hidden states, enabling direct linear control of the reasoning process \cite{steering}. For example, Anthropic introduced SAE to map dense hidden states into an interpretable sparse space, allowing explainable steering along specific directions \cite{sae}. Some recent work has conducted emotion steering, but these approaches generally focus on controlling discrete emotion labels \cite{emo-steer}. And they primarily evaluate whether generated text exhibits the target emotional features, without examining broader model behaviors.

\section{Preliminary}

\subsection{Valence–Arousal–Dominance (VAD) Representation}
\begin{figure}[h]
    \centering
    \includegraphics[width=0.8\linewidth]{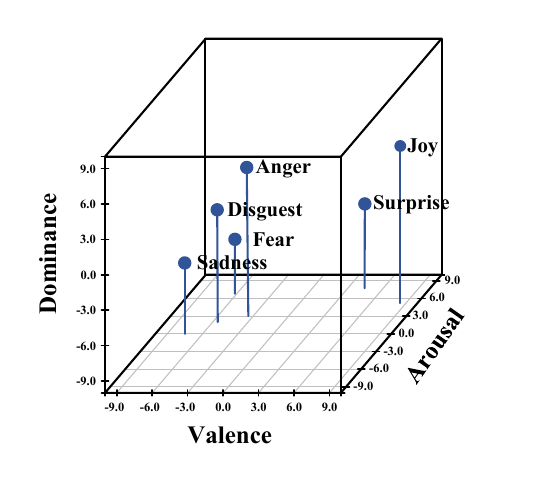}
    \caption{Distribution of emotional labels in the VAD space}
    \label{fig: vad space}
\end{figure}

Traditional emotional expressions mainly use discrete labels, such as happy, angry, and trust. Another expression form, VAD, represents affective states as continuous coordinates in a three-dimensional space \cite{vad}, as shown in \cref{fig: vad space}. Valence describes the degree of positivity or negativity, arousal reflects its intensity or activation level, and dominance captures the sense of control associated with the affective state. In this study, the ranges for all three dimensions are uniformly set to $[\,-10, 10\,]$. 

The VAD framework originates from empirical studies in affective
psychology and represents a wide range of human emotions, with a small number of semantically interpretable dimensions. Compared with discrete emotion labels, VAD provides a continuous and fine-grained representation that enables quantitative analysis and parametric modeling in a low-dimensional space. Owing to its limited dimensionality, interpretability, and practical flexibility, VAD has been widely adopted in affective intelligence. It has become a factual standard for continuous emotion modeling in tasks such as emotion recognition \cite{dl-vad-anl}, intensity analysis \cite{vad-ia}, and emotion-aware dialogues \cite{vad-dlg}.

\subsection{Sparse Autoencoder (SAE)}
SAE can decompose LLM hidden states into a set of interpretable and controllable latent features. By mapping dense hidden representations into the sparse latent space, the SAE provides a structured interface for analyzing and intervening in specific factors encoded by the model. The SAE follows an encoder–latent–decoder architecture. SAE receives the hidden states of LLM $\mathbf{h}_k$, constructs them as latent representations $\mathbf{z}$, and finally outputs the reconstructed hidden states $\mathbf{h}_k'$.
\begin{equation}
    \mathbf{z} = f_{enc}(\mathbf{h}_k), \ \mathbf{h}_k' = f_{dec}(\mathbf{z}),
    \label{eq: sae mapping}
\end{equation}
where $f_{enc}$, $f_{dec}$ refer to SAE encoder and decoder functions, respectively. 

SAEs constrain only a small subset of latent neurons to activate significantly, jointly forming the latent representation pattern for a particular feature. With sufficient dimensional capacity, distinct neuron subsets encode different features, promoting functional specialization. This sparsity allows targeted analysis and controllable modulation of specific features by intervening on the corresponding latent dimensions \cite{sae}.
Existing SAE-based interventions operate either within MLP submodules to enhance interpretability or at block-level hidden states to improve behavioral stability and controllability. To achieve control over emotion-related features expressed in model behavior, we apply steering at the output of the $k$-th block.
The encoder would include nonlinear activations, whereas the decoder is a single linear layer. Variations in latent features are translated into approximately linear changes in reconstructed hidden states. It enables feature-level linear controllability of LLM hidden representations.

\subsection{Problem Formulation}
Let $x \in \mathcal{X}$ denote an input prompt consisting solely of the task description, and let $y = f_\theta(x)$ be the output produced by a LLM with parameters $\theta$ (or the core LLM within an agent). Let $\mathbf{h}_k \in \mathbb{R}^{d_h}$ represent the hidden state at the $k$-th transformer block during inference, where $d_h$ represents the hidden size of the LLM.

During computation and reasoning, the model maintains an internal emotional state represented by a continuous vector, denoted as 
\begin{equation}
\mathbf{e} = [\,e_v, e_a, e_d\,] \in \mathbb{R}^3,
\end{equation}
where $e_v$, $e_a$, and $e_d$ correspond to the valence, arousal, and dominance components of the emotional vector, respectively.

The objective of this study is to investigate how different emotional states affect the task performance of the model. Therefore, we need to obtain two relations: 
First, we need to know how emotion $\mathbf{e}$ modifies the hidden representations $\mathbf{h}_k$ of the model to induce a target state, i.e.,
\begin{equation}
    g_1: (\mathbf{h}_k, \mathbf{e}) \rightarrow h_k^{(\mathbf{e})}.
\end{equation}

Second, we define an emotion-behavior function to characterize how emotional variations affect the model's output behavior,
\begin{equation}
    g_2: (x_0, \mathbf{e}) \rightarrow \mathcal{E}(f_{\theta}(x_0|{\mathbf{h}_k \leftarrow \mathbf{h}_k^{(\mathbf{e})})}),
\end{equation}
where $x_0$ is a fixed input prompt and $\mathcal{E}(\cdot)$ denotes a task-specific evaluation function. 

Together, $g_1$ specifies the controllable steering mechanism, while $g_2$ captures the resulting emotion-behavior relationship.

\section{Study Design}
\subsection{Research Framework}
\begin{figure*}[htbp]
    \centering
    \includegraphics[width=0.9\linewidth]{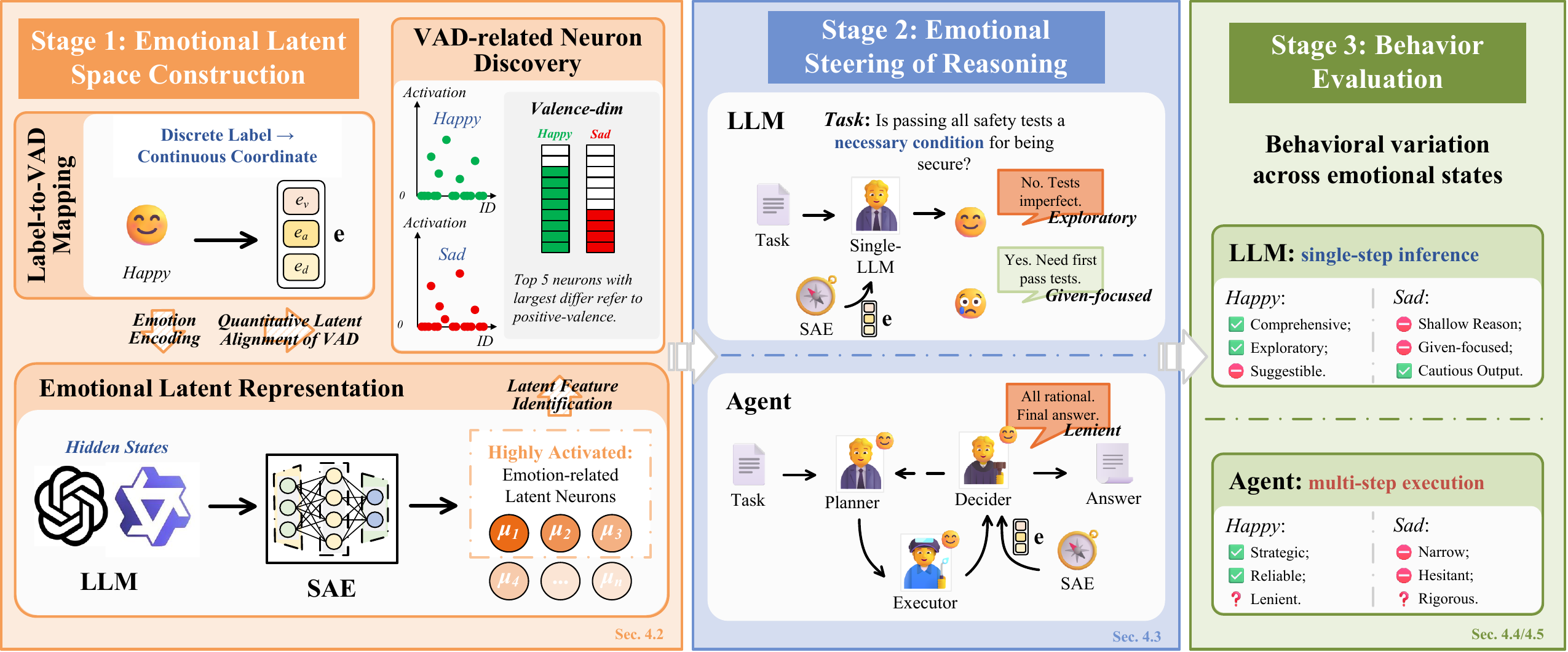}
    \caption{The framework of \textbf{E-STEER}. It consists of three stages: \textbf{(1) Emotional Latent Space Construction}, which derives interpretable latent representations for LLM hidden states aligned with the coordinates; \textbf{(2) Emotional Steering of Reasoning}, where implicit emotional states and reasoning behaviors are continuously modulated by intervening on selected SAE features without altering tasks or prompts; and \textbf{(3) Behavior Evaluation}, which systematically assesses changes in task performance under different VAD configurations.}
    \label{fig: framework}
\end{figure*}

We proposed an interpretable VAD precise-control framework, E-STEER, as shown in \cref{fig: framework}. By explicitly mapping VAD dimensions onto internal representations, emotional states can directly influence reasoning processes. Without modifying tasks or prompts, VAD parameters can be continuously adjusted to induce control. This provides a unified methodology for studying how emotion affects internal representations and task performance in LLMs and agents.

\subsection{VAD-based Feature Extraction}
To identify VAD-based latent neurons, we employ a positive–negative contrastive procedure. Specifically, positive–negative pairs are constructed by fixing the same task description while varying only the assigned emotional labels. By comparing the activated neuron subsets in the latent space, the implicit representation of changes in emotional features can be extracted. For each sample, the hidden state at the $k$-th block is encoded into the latent space via the SAE encoder.

Since SAE latent neurons are designed to encode feature strength with non-negative activations, the activation difference of a neuron between a positive–negative pair reflects the change of the emotional feature in the latent representation. Larger differences indicate stronger associations. Following this criterion, the top 50 neurons with the largest activation differences are selected as candidates. And those with stable patterns across multiple contrastive pairs are retained as the final emotional latent representation. This feature identification process is conducted separately for the Valence, Arousal, and Dominance dimensions.

\subsection{Emotional State Steering Mechanism}
Activation patching enables that, without altering model parameters or input prompts, controlled interventions can be achieved by replacing or perturbing hidden states. SAE steering is instantiated through this, where hidden states reconstructed with feature shifts are re-injected into the model, guiding the content toward desired directions.

\begin{figure}[tbp]
    \centering
    \includegraphics[width=\linewidth]{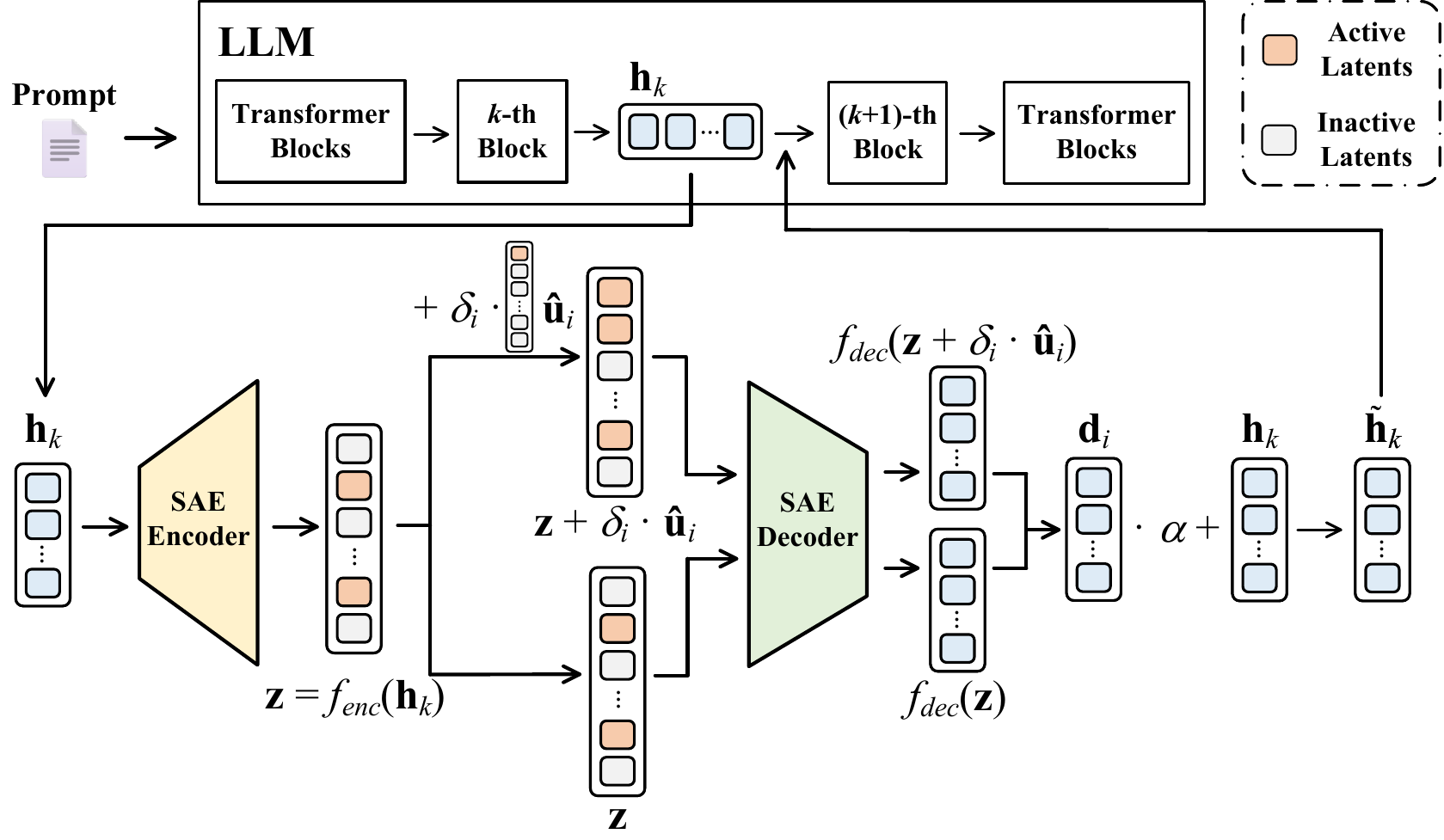}
    \caption{The SAE steering pipeline}
    \label{fig: steering}
\end{figure} 

Steering is applied at the same block used for training and feature extraction. Specifically, the intervention is implemented by additively injecting a steering direction $\mathbf{d}_i$ into the original hidden state $\mathbf{h}_k$. This direction represents changes of the target emotional feature $e_i \in \{e_v, e_a, e_d\}$ in the model hidden space. And it can be derived from the mapping established by the SAE between latent feature shifts and hidden states. Formally, given a latent feature offset $\delta_i$, the steering direction $\mathbf{d}_i$ is defined as:
\begin{equation}
    \mathbf{d}_i = f_{dec}(\mathbf{z}+\delta_i\cdot \hat{\mathbf{u}}_i) - f_{dec}(\mathbf{z}),
    \label{eq: direction}
\end{equation}
where $\hat{\mathbf{u}}_i$ denotes a unit vector in the latent space representing the direction spanned by the subset of latent neurons corresponding to feature $e_i$. Compared to directly replacing hidden states with SAE reconstructed ones, direction-based steering preserves task-irrelevant information in the original representation, thereby avoiding excessive semantic distortion. This design results in more stable, controllable, and interpretable steering behavior. It is important to note that hidden representations at the $k$-th block vary across different tasks. Directly adding a steering direction may therefore lead to inconsistent steering effects: in some tasks, a small $\delta_i$ may already cause generation instability, while in others the same produces negligible influence. Consequently, scale alignment between the steering direction and the original hidden state is necessary, and it can be represented as:
\begin{equation}
    \tilde{\mathbf{d}}_i = \frac{\mathbf{d}_i}{\|\mathbf{d}_i\|} \cdot \|\mathbf{h}_k\| \cdot \frac{\delta_i}{\delta_{max}}.
    \label{eq: scale align}
\end{equation}
Accordingly, the final hidden states after steering $\tilde{\mathbf{h}}_k$ can be expressed as:
\begin{equation}
    \tilde{\mathbf{h}}_k = \mathbf{h}_k + \alpha \cdot \sum_{e_i \in \{e_v, e_a, e_d\}} \tilde{\mathbf{d}}_i.
\end{equation}

Each emotion-related feature corresponds to a specific subset of latent neurons. To control multiple VAD dimensions simultaneously, distinct latent neuron subsets are assigned different $\delta_i$ values. The resulting steering directions are then linearly combined to achieve joint multidimensional control. In practice, steering is implemented via forward hooks: during the forward pass, the hidden state after the $k$-th block is intercepted, modified according to the computed steering direction, and then re-injected into the model to guide subsequent generation.

\subsection{Evaluation of Emotional Effects on Reasoning}
To determine whether emotional states alter reasoning forms, we design tasks at both the LLM and Agent levels, examining effects from the perspectives of cognition and sequential decision-making.

At the LLM-level, we employ three task types. Objective tasks have deterministic goals, requiring the model to integrate prior knowledge and select the most credible solution among multiple candidate reasoning paths. More challenging instances typically require multi-step reasoning and thus better reflect reasoning depth and process consistency. Subjective tasks involve comprehension and open-ended generation, emphasizing relevant argumentation and creative expression. These tasks allow us to examine whether emotion modulates cognitive capability and expressive style, such as coherence, divergence, and conciseness. Safety tasks require judgments under uncertainty and specific constraints, enabling us to probe whether emotional states influence risk assessment tendencies and boundary adherence.

At the Agent-level, we analyze the sequential process across three stages: planning, deciding, and execution. Planning reflects task decomposition and strategy formation, including two forms: initial planning, captures the foresight of the agent before execution; and replanning, reflects the ability to revise plans based on intermediate execution outcomes. The decider determines whether replanning is necessary, selects the best candidate solution, and diagnoses execution failures. Emotional effects may be most pronounced at this stage, as affective states could alter the strictness of decision criteria. Execution represents action implementation and is used to assess whether emotion influences operational efficiency, stability, and consistency in long-horizon tasks.

\subsection{Behavioral Patterns Across Emotional States}
To understand how affective modulation shapes the final performance of model, we analyze behavioral metrics across emotional states to identify systematic response patterns and the optimal affective space.
We first analyse the overall trends of metrics as emotional states vary. Human performance is known to respond nonlinearly to emotion, for example, along the arousal dimension it follows an inverted-U profile. Similarly, we hypothesize that extreme emotional states may also drive model performance toward lower bounds.
On this basis, we locate optimal performance range. We also assess whether these optimal ranges remain consistent across similar and heterogeneous tasks, to evaluate the model’s sensitivity to emotional variation.

To quantify the impact of affective modulation, we compare optimal performance against the neutral baseline. The resulting relative gains demonstrate that emotion is not a minor perturbation but a controllable variable capable of systematically shaping model behavior. In specific conditions, performance also exhibits asymmetry between positive and negative states, with positive affect yielding overall superior outcomes.
Different emotional dimensions influence performance to varying degrees; thus, we conduct sensitivity analysis across valence, arousal, and dominance. By measuring the fluctuation range (defined as $\frac{max-min}{average}$), we identify which dimension induces the largest behavioral variation in reasoning.

\section{Experimental Analysis}
\subsection{Experiment Setup}
\textbf{Experimental Design.}  
In this study, we design four task categories. \textbf{(1) LLM objective behavior:} assess reasoning and problem-solving, spanning Logical Reasoning, Code Generation, and Quantitative/Scientific tasks. \textbf{(2) LLM subjective behavior:} evaluate open-ended text generation and creativity. \textbf{(3) LLM safety:} test the model’s ability to avoid unsafe outputs under attack. \textbf{(4) Agent behavior:} examine multi-step Planning, Decision-making, and Execution, capturing how emotional biases accumulate along long reasoning chains. Each task is evaluated along the valence, arousal, and dominance dimensions.

For LLM behavior evaluation, the experiments are divided into the primary and validation. Model sampling is disabled in the primary experiments to ensure reproducibility. We use Qwen3-8B \cite{qwen} as the LLM and attach the trained SAE to its $k=17$th layer. Additional validation experiments are reported in the Appendix~\ref{8a: val-exp}.

For agent behavior evaluation, we construct a lightweight agent composed of three modules: planner, decider, and executor. The planner is responsible for plan generation, with the ability to replan when necessary. The decider serves as an intermediate controller. It needs to validate the feasibility of plans, diagnose execution failures, and select the final answer from candidate outputs. The executor performs the planned operations through tool usage or LLM-based analysis. It should provide the confidence associated with the execution to other modules.

\textbf{Datasets.}  
For each behavior, we select a representative benchmark dataset: \textbf{(1) LLM objective behavior:} LogiQA 2.0 \cite{logiqa} for logic reasoning; HumanEval \cite{humaneval} for code generation; Math \cite{math} for quantitative and scientific. \textbf{(2) LLM subjective behavior:} TinyStories \cite{tinystories}. \textbf{(3) LLM safety:} HarmBench \cite{harmbench}. 
\textbf{(4) Agent behavior:} we combine HotpotQA \cite{hotpotqa}, Scientific \cite{scientific} and GAIA \cite{gaia}, comprehensively testing the behaviors of the agents.

All datasets were preprocessed with basic steps, including deduplication, filtering, and format standardization. For each task, subsets were sampled proportionally based on specific criteria (e.g., difficulty level) to ensure representativeness and reproducibility.

\textbf{Metrics.} 
For LLM behaviors, we categorize metrics into efficiency and quality. \textbf{(1) Efficiency metrics:} focus on whether the model can produce a parsable answer within the fixed maximum generation length, quantified by the Answer Validity Rate (\textbf{AVR}). \textbf{(2) Quality metrics:} depend on task type: for objective abilities, they refer to Task Success Rate (\textbf{TSR}), including pass@1 for Code Generation and accuracy for others. For subjective abilities, they include relevance, coherence, creativity, and conciseness. These Context Qualities are evaluated by the LLM. For safety, we focus on the Probability of Safety Risks, including harmful, biased, and hallucinatory outputs. 

For agent behaviors, we analyze from module-level and system-level perspectives. \textbf{(1) Planning metrics:} include two metrics: Plan Validity Rate, the proportion of executable plans; and Replan Improvement, the improvement in executor confidence from the initial plan to the revised plan for tasks that trigger replanning. \textbf{(2) Decision-making metrics:} capture the judgments of the decider over both plans and answers. Plan-part judgment evaluates the assessment ability for plan feasibility, including Replan Frequency and Replan Trigger Confidence. Answer-part decision focuses on the ability to select the final answer, measured by the Rational Selection Rate, defined as the rate at which the decider selects the candidate with the highest execution confidence among all. \textbf{(3) Executing metrics:} focus on the Execution Completion Rate, reflecting the ability to complete the planned operations. Additionally, \textbf{(4) System-Level Behaviors} are measured by the Overall Success Rate of tasks.

\subsection{LLM Objective Behavior}

\begin{figure*}[tbp]
    \centering
    \begin{subfigure}[b]{0.332\linewidth}
        \includegraphics[width=\linewidth]{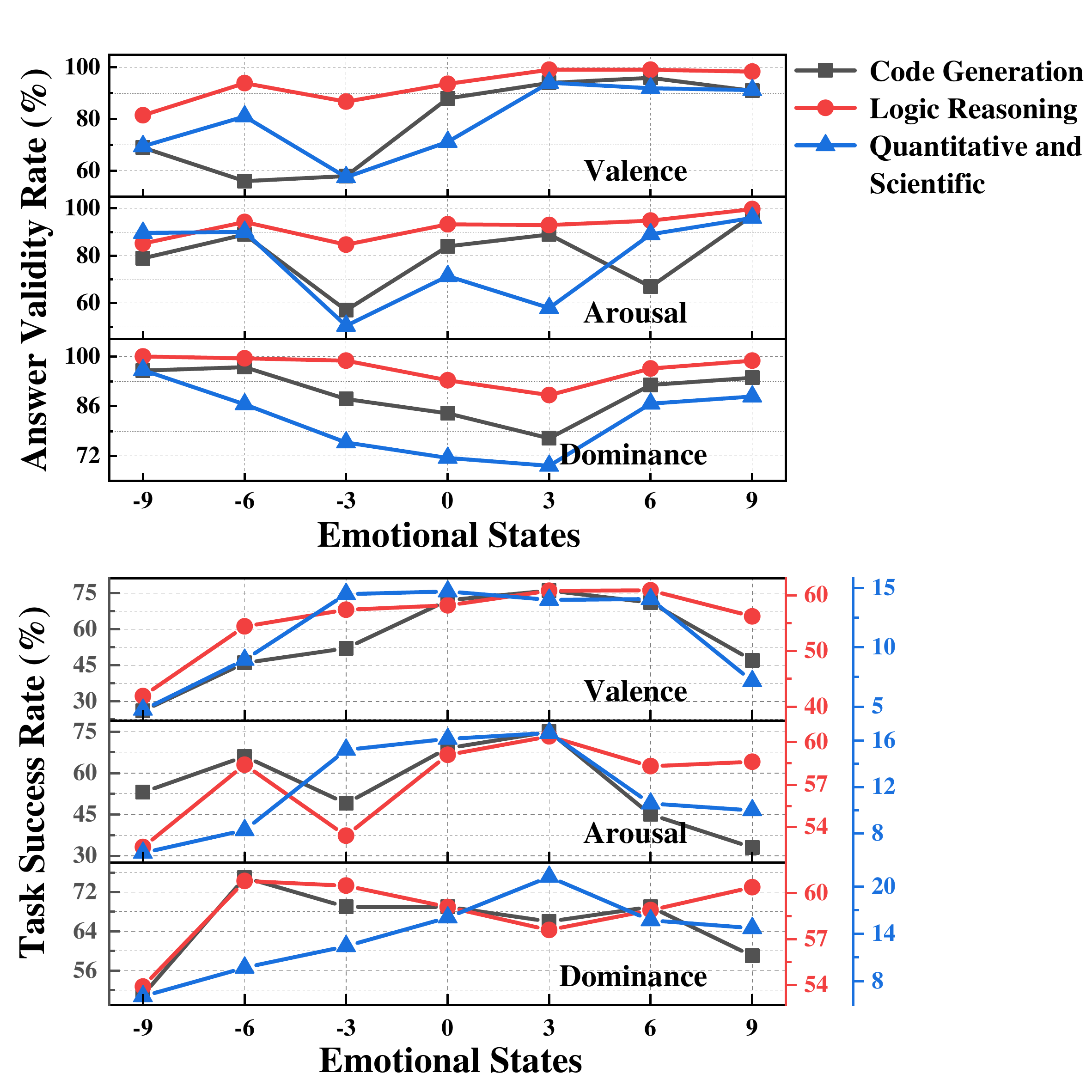}
        \caption{LLM objective behaviors}
        \label{fig: llm obj}
    \end{subfigure}
    \begin{subfigure}[b]{0.315\linewidth}
        \includegraphics[width=\linewidth]{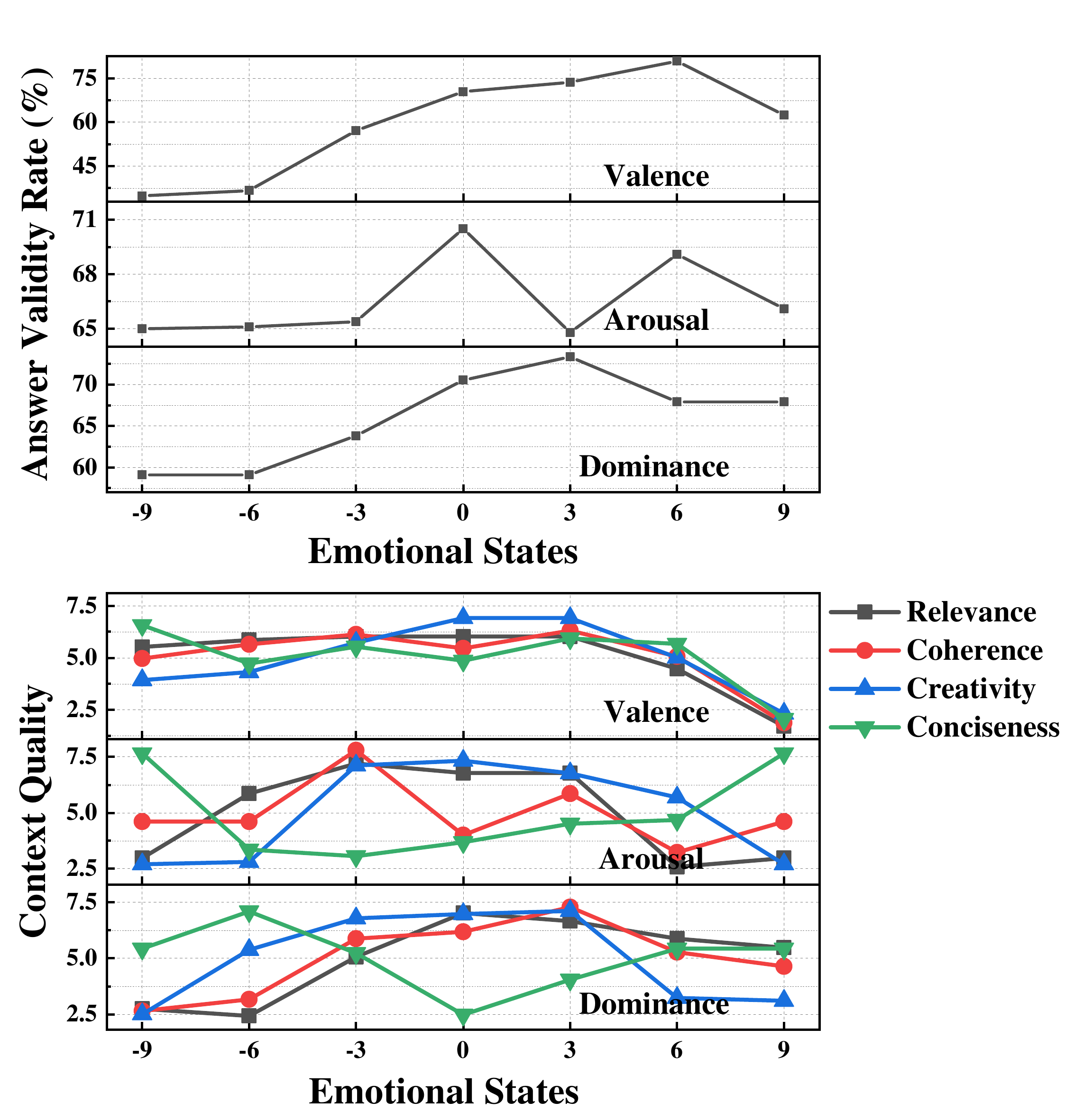}
        \caption{LLM subjective behaviors}
        \label{fig: llm sub}
    \end{subfigure}
    \begin{subfigure}[b]{0.33\linewidth}
        \includegraphics[width=\linewidth]{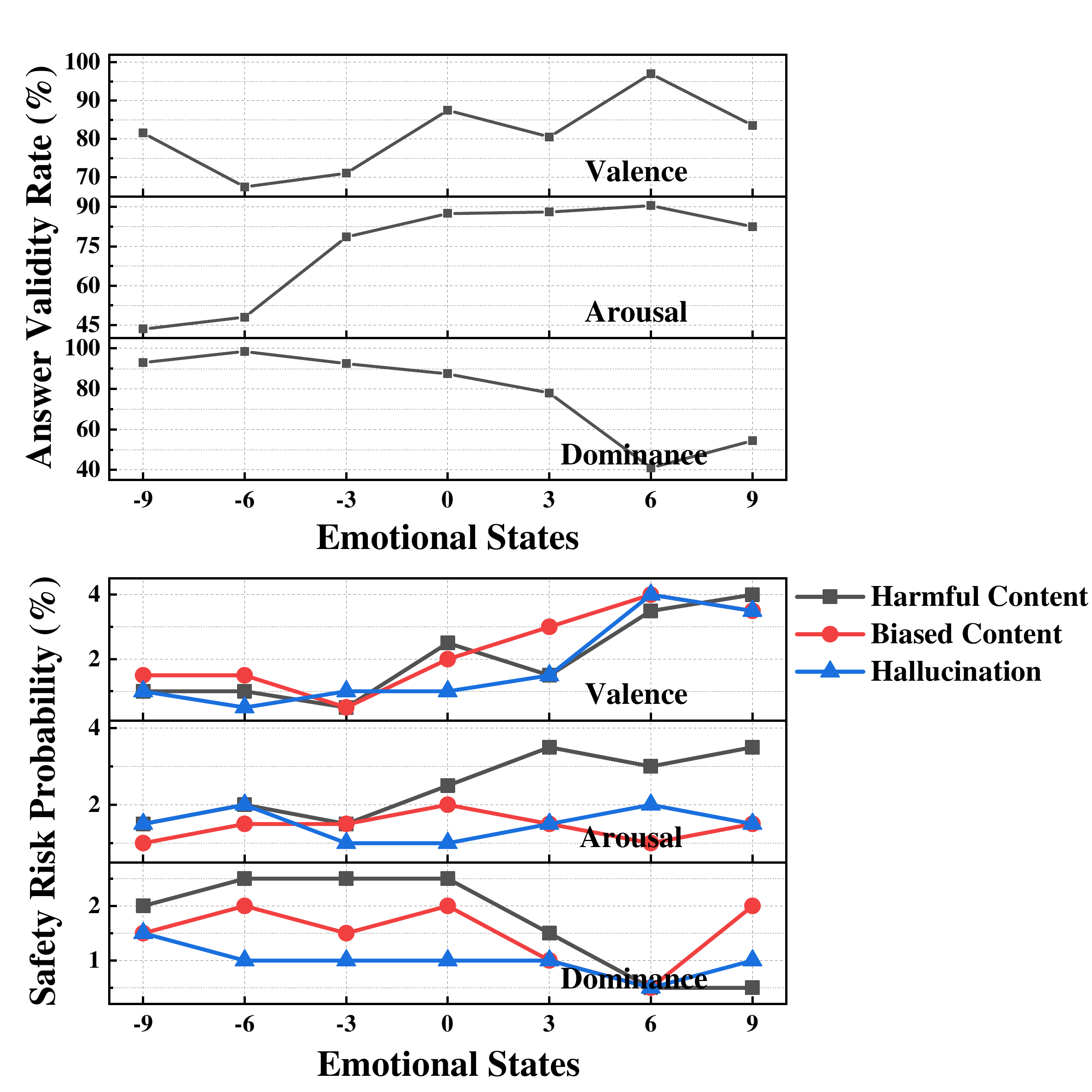}
        \caption{LLM safety}
        \label{fig: llm safe}
    \end{subfigure}
    \caption{The behaviors of LLM across emotion states}
    \label{fig: llm}
\end{figure*}

\cref{fig: llm obj} shows the performance of LLMs on objective tasks across emotional states. Positive valence promote more active reasoning, yielding 33.1\% higher AVR than negative ones. In contrast, AVR along with arousal and dominance, exhibits a U-shaped pattern: excessive activation tends to cause the model to end reasoning prematurely. And the trough occurs at arousal = -3 and dominance = +3.

For quality metrics, performance exhibits an inverted U-shaped trend across all three VAD dimensions, indicating moderate levels are optimal. Along the valence, TSR peaks at positive valence and shifts toward neutrality as task difficulty increases. Compared with neutral states, it improves TSR by 3.4\% on average. For arousal, the moderately excited state (+3) makes the best performance, yielding a 4.7\% improvement. The optimal dominance level varies with task difficulty. For easier tasks, repeated verification exhibited under lower dominance (-6) improves TSR; as task difficulty increases, the optimal point shifts toward higher dominance (e.g., +3 in quantitative/scientific tasks), since excessive deliberation may introduce errors. And overall performance improves by up to 14.5\% compared with neutral states. In conclusion, higher valence and lower dominance benefit simpler tasks, whereas the opposite is effective for more difficult tasks. Higher arousal consistently yields better performance across all difficulty levels.
Across dimensions, valence produces the largest performance fluctuation range, 71.2\%, while variability decreases by 17.8\% for arousal and 25.6\% for dominance.

\subsection{LLM Subjective Behavior}
The performance of LLMs on subjective tasks under different emotional states is shown in \cref{fig: llm sub}. The VAR curves of all three dimensions exhibit an inverted U-shaped pattern. Subjective tasks involve stronger exploration and looser boundaries; both self-conflicted reasoning at low emotional states and active exploration at high emotional states result in longer reasoning chains.

Trends of context quality vary from metrics. Relevance, coherence, and creativity all exhibit an inverted U-shaped pattern. Moderate calmness (arousal=-3) and confidence (dominance=+3) improve textual relevance and coherence, while mild positivity (valence=+3) further enhances creativity. Compared with neutral states, these settings improve performance by 5.2\%, 33.6\%, and 6.5\%, respectively. Additionally, negativity leads to more concise outputs, whereas positivity tends to introduce redundancy; conciseness improves by 23.3\% under negative compared with positive valence. Calm or confident states support efficient generation, while overly activated or constrained states hinder text expansion. It results in increased conciseness at both extremes of the arousal and dominance dimensions.

\subsection{LLM Safety}
\cref{fig: llm safe} illustrates the impact of emotions on LLM safety. The AVR for safety exhibits a consistent trend with that for subjective tasks. In contrast, under high dominance (e.g., +6), the safety-specific requirements constrain model outputs into generic “I cannot answer” responses, significantly decreasing AVR. Low valence and arousal facilitate analytical processing, substantially reducing safety risks: compared with the neutral state (0), safety risk probability decreases by 52.7\% at valence = -3 and 21.7\% at arousal = -3. High dominance induces more controlled and disciplined behavior of the model, peaking at +6 with an average 68.3\% improvement over neutral. These results suggest that emotion influences not only task-level performance but also the internal risk-sensitive response mechanisms.

\subsection{Agent Behavior}

\begin{figure*}[tbp]
    \centering
    \begin{subfigure}[b]{0.24\linewidth}
        \includegraphics[width=\linewidth]{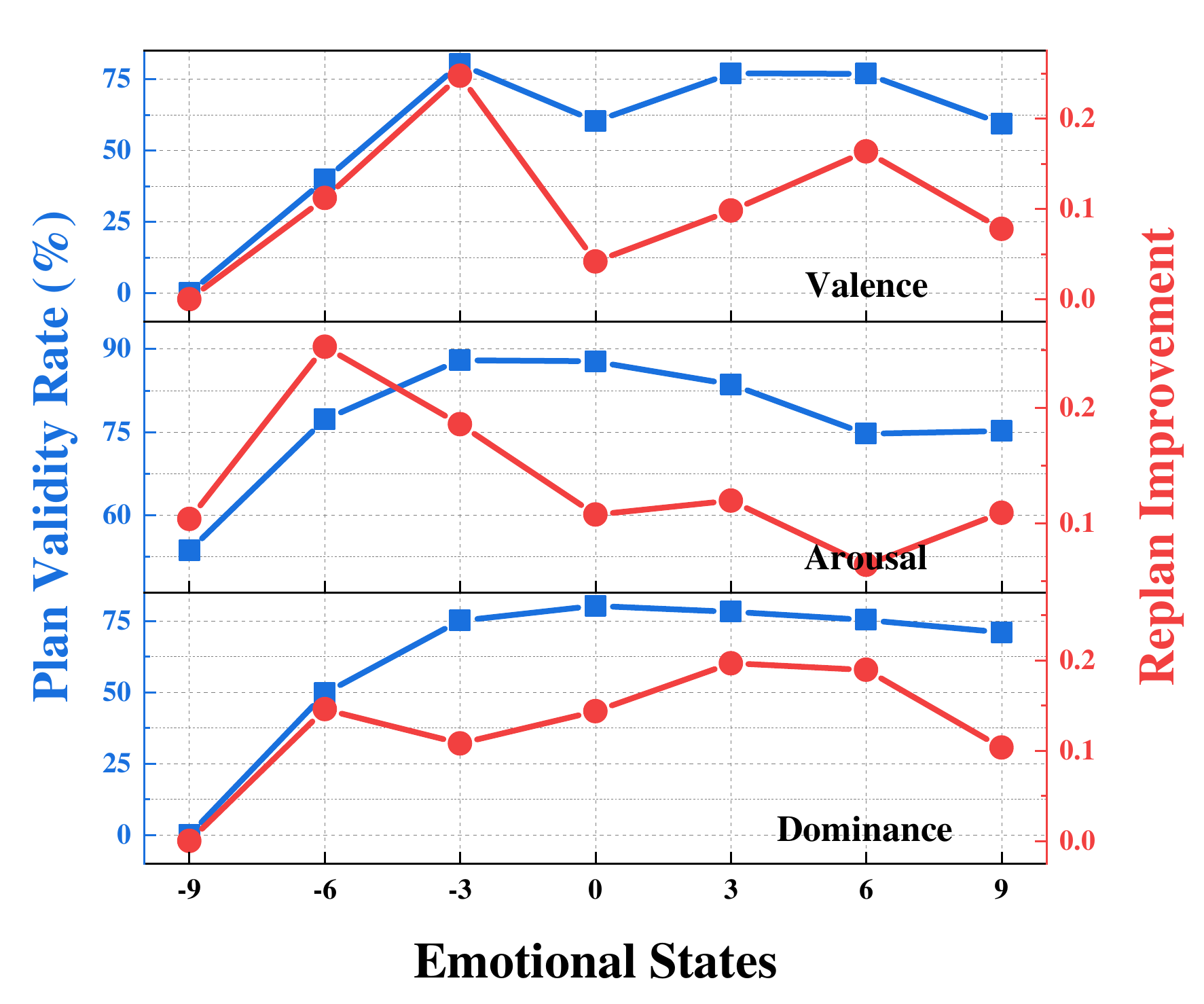}
        \caption{Agent planning}
        \label{fig: agent plan}
    \end{subfigure}
    \begin{subfigure}[b]{0.50\linewidth}
        \begin{subfigure}[b]{0.48\linewidth}
            \includegraphics[width=\linewidth]{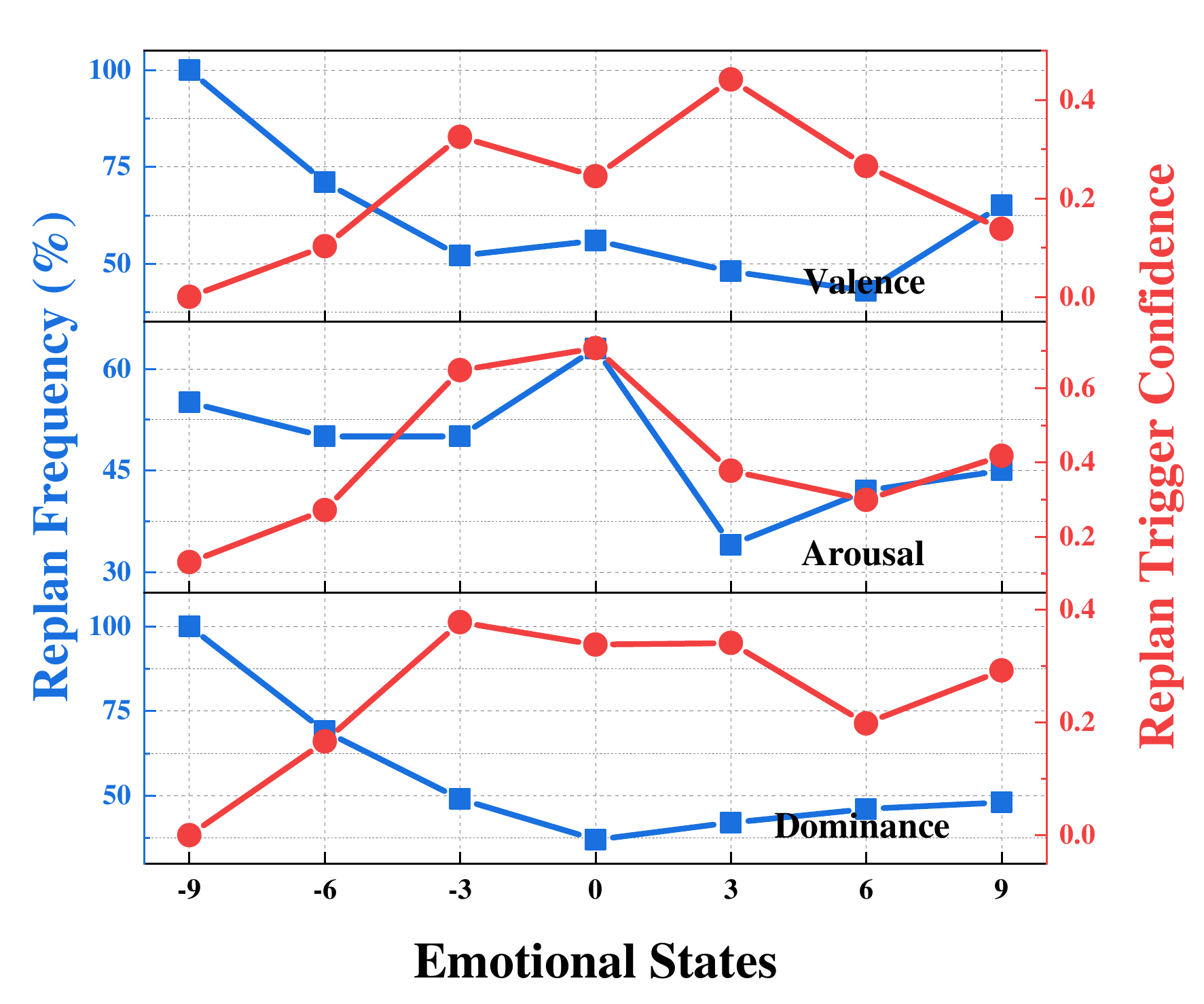}
        \end{subfigure}
        \hfill
        \begin{subfigure}[b]{0.48\linewidth}
            \includegraphics[width=\linewidth]{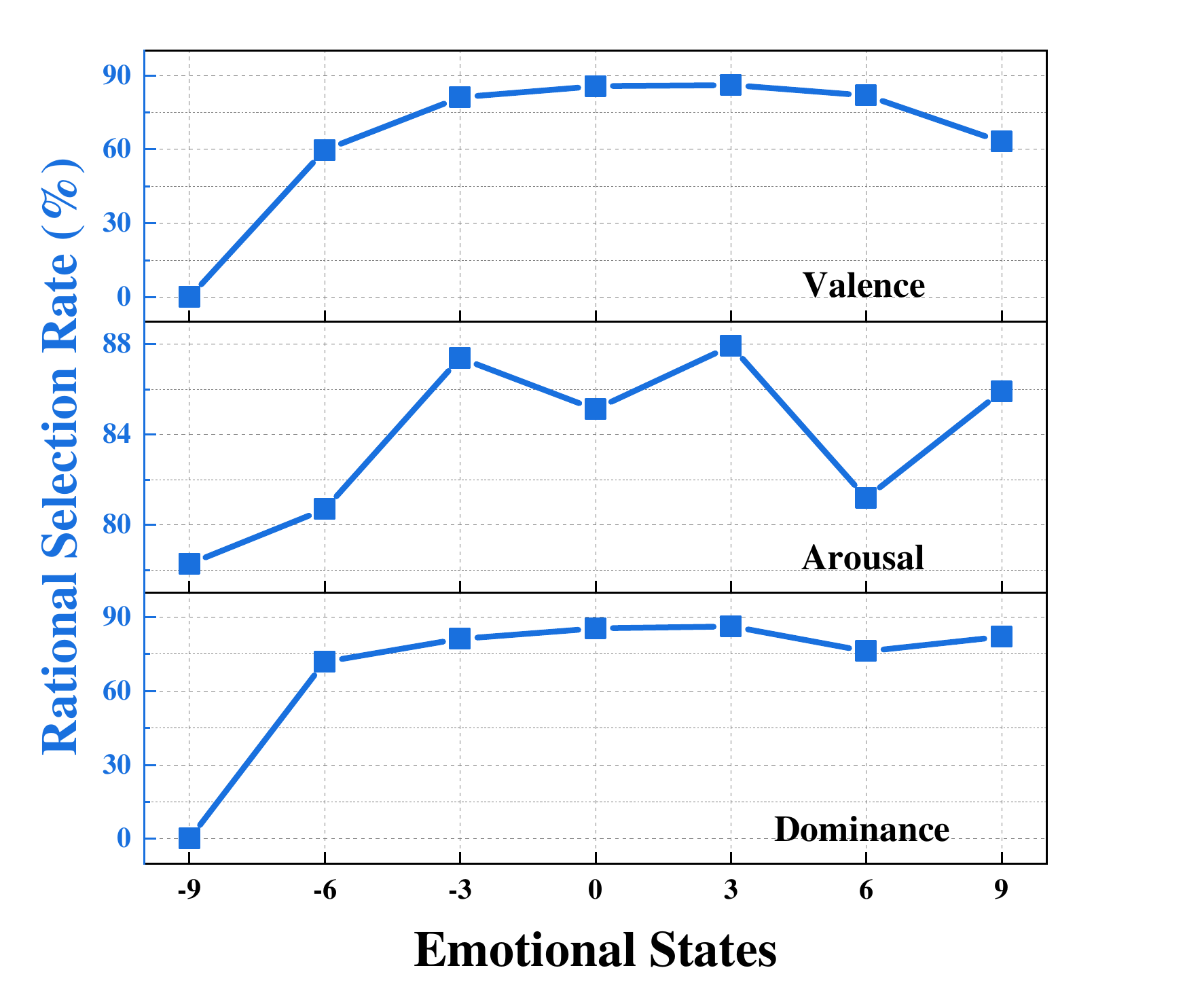}
        \end{subfigure}
        \caption{Agent decision-making}
        \label{fig: agent decide}
    \end{subfigure}
    \begin{subfigure}[b]{0.24\linewidth}
        \includegraphics[width=\linewidth]{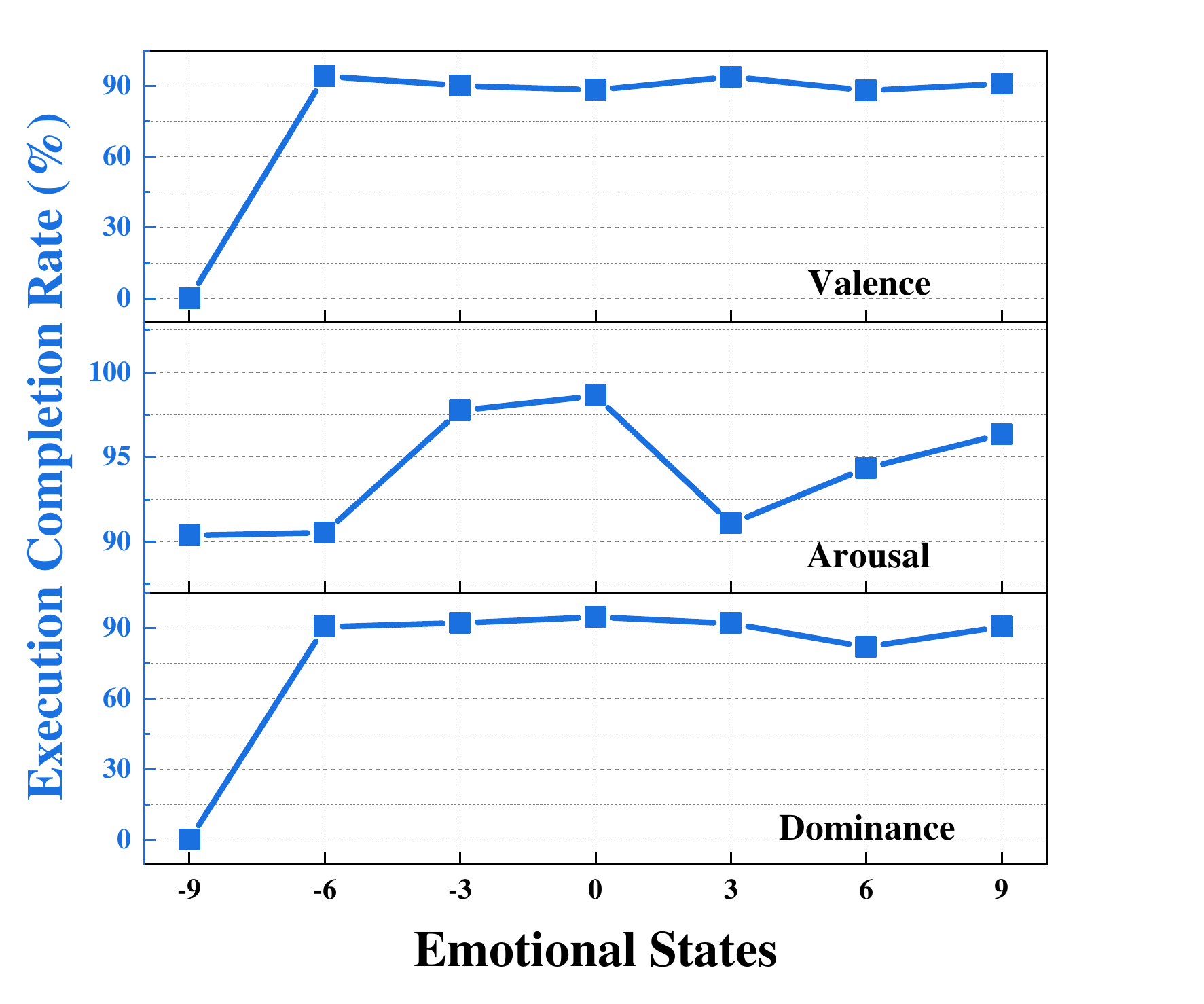}
        \caption{Agent execution}
        \label{fig: agent execute}
    \end{subfigure}
    \caption{The behaviors of different agent modules across emotion states}
    \label{fig: agent}
\end{figure*}

At the planning stage, plan validity rate exhibits an inverted-U trend across three dimensions, as shown in \cref{fig: agent plan}. Reduced valence and arousal support more systematic task analysis, with performance peaking at valence = -3 and arousal = -3. Compared with neutral states, they improved the rate by 33.2\% and 0.3\%, respectively. Elevated dominance strengthens the global grasp of task goals and available execution options, resulting in more effective plans; positive dominance improves performance by 79.8\% on average compared to negative states. 
Replan improvement follows a similar pattern, with peaks (valence=-3, arousal=-6, dominance=+3) closely aligned with those of validity plans. This indicates that emotional states can not only enhance effective planning but also promote more efficient plan refinement across multiple iterations.

\cref{fig: agent decide} shows the trends of agent decision-making behavior. Replan frequency exhibits a U-shaped pattern, where lower valence and dominance increase the tendency to negate prior plans, and lower arousal imposes stricter thresholds for plan revision. The frequency reaches its minimum at valence = +6 and arousal = +3, decreasing by 23.2\% and 46.0\% relative to neutral states, respectively. Positive dominance reduces replanning as well, yielding a 37.6\% decrease compared with negative states. In contrast, the replan trigger confidence follows an inverted-U trend, with the peaks (valence=+3, arousal=0, dominance=-3) largely overlapping with that of replan frequency. Initial plans tend to have higher execution confidence when replanning is less frequent, leading to higher trigger confidence when replanning does occur.   
Meanwhile, the rational selection rate exhibits an inverted-U pattern, peaking at positive valence (+3), arousal (+3), and dominance (+3), yielding an average 42.4\% higher rate than negative states across the three dimensions. These states enable the decider to make more consistent and rational final answer selections based on available information.

The executor is the least affected by emotional factors, as shown in \cref{fig: agent execute}, because it primarily involves objective tool-use. Nevertheless, higher valence, lower arousal, and higher dominance are still associated with a higher likelihood of completing the planned executions.

\begin{figure}[htbp]
    \centering
    \includegraphics[width=\linewidth]{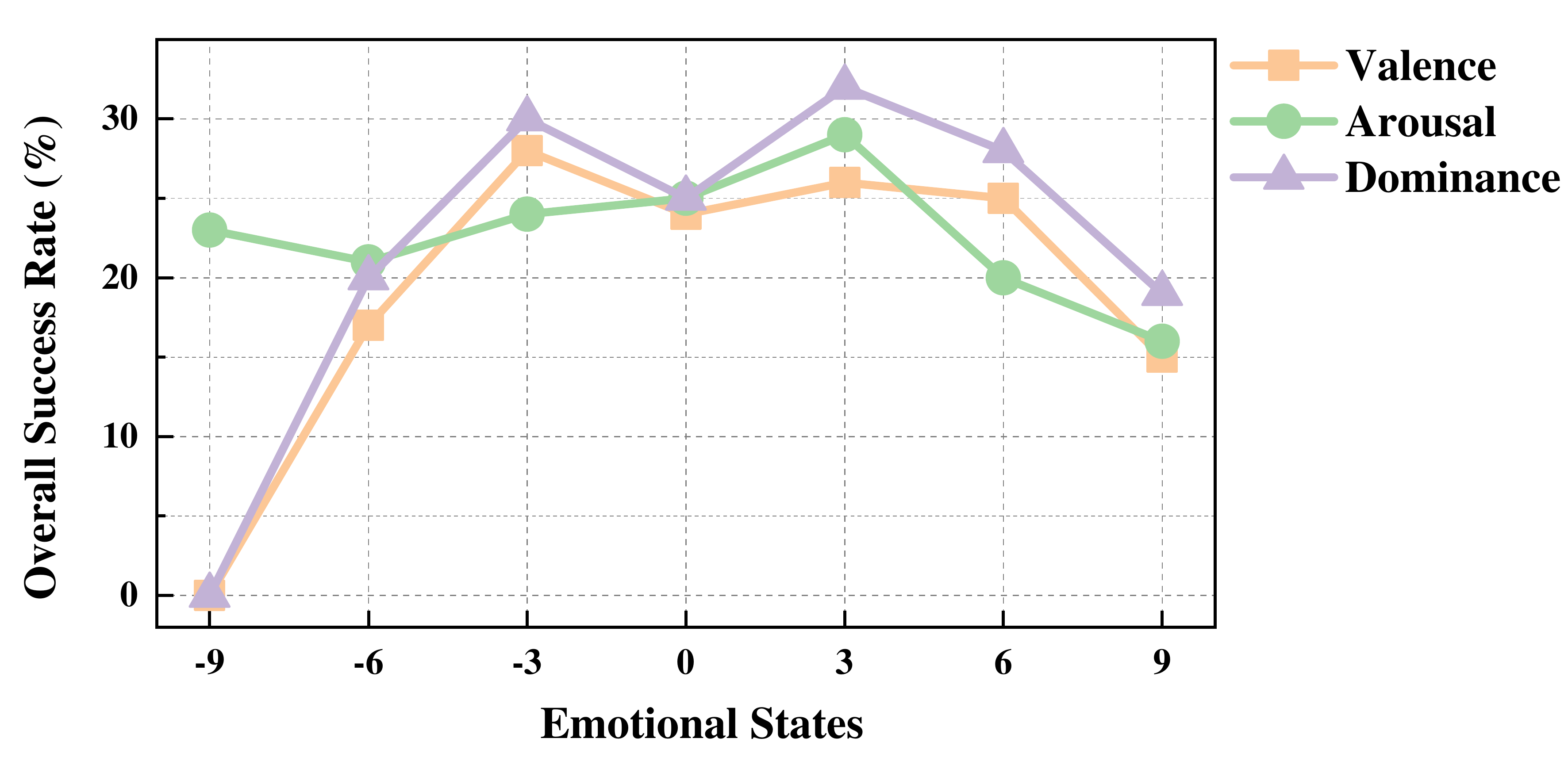}
    \caption{Overall task success rate of the Agent}
    \label{fig: agent overall}
\end{figure}

\cref{fig: agent overall} reports the system-level overall success rates across emotional conditions. An inverted-U pattern is consistently observed across all three dimensions. Lower valence (-3), higher arousal (+3), and higher dominance (+3) promote deeper task analysis and lead to improved success rates, consistent with LLM behavior on more difficult objective tasks. Compared with neutral states, dominance yields the largest improvement (28.0\%), followed by arousal (16.7\%) and valence (16.0\%). Performance variability is greatest along valence and dominance, with fluctuation ranges of 145.2\% and 145.5\%, respectively, far exceeding that for arousal (57.6\%). Notably, the performance peaks do not perfectly align with those at the module-level, highlighting the multi-step and integrative nature of agent systems.

\subsection{Validation of VAD-related Latents}
\begin{figure}[htbp]
    \centering
    \includegraphics[width=\linewidth]{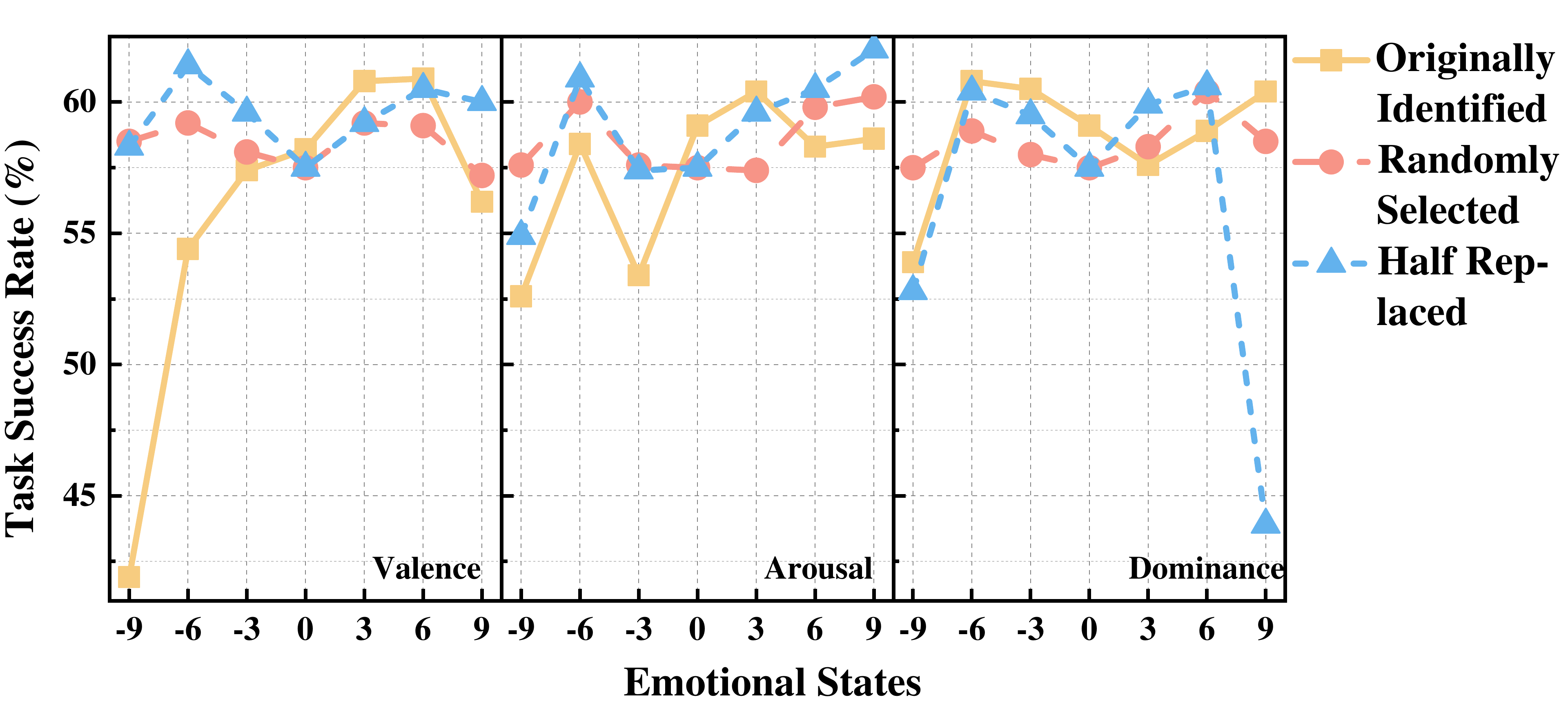}
    \caption{Validation of selected latents as VAD features}
    \label{fig: ablation}
\end{figure}

To verify that E-STEER indeed operates on VAD-related features, we conduct the validation experiment on LogiQA 2.0. We evaluate three settings: \textbf{(1) the originally identified latent neurons}, \textbf{(2) randomly selected neurons}, and \textbf{(3) half of the originally selected neurons with the rest randomly replaced}. As shown in \cref{fig: ablation}, random selection exhibits only mild fluctuations, losing the distinct patterns across emotional states. Overall, its performance variation decreases by 70.9\% compared with the original selection. Half replacement preserves an intermediate trend. For example, along the valence, random selection (3.4\%) drops by 90.0\% relative to the original (34.1\%), whereas half replacement (6.6\%) decreases by 80.8\%, remaining between the two. Although the half replacement presents an isolated deviation causing a sharp TSR drop, the overall pattern, excluding this extreme point, remains intermediate.

\section{Conclusion}
In this paper, we present a novel emotion steering framework, E-STEER, for LLMs and Agents by mapping dense hidden representations with an interpretable sparse space, enabling both interpretability and controllability. Direct manipulation on internal hidden states provides improved linear control over model outputs. Using this framework, we systematically analyze the effects of different emotional states on the task performance of LLMs and LLM-based agents, establishing a connection between psychological emotion theories and model behavior.

However, emotion modeling based on the VAD theory has inherent limitations: as valence, arousal, and dominance are not strictly orthogonal, making it difficult to ideally disentangle their individual effects.

Future work will extend this framework to multimodal settings and more task types, such as geometry. In addition, emotional states may evolve during task execution, and such dynamics can substantially influence the final outcomes. Modeling this emotional evolution and designing adaptive regulation mechanisms is also an important direction for future research.

\section*{Impact Statement}
This work aims to advance the understanding of how emotional signals interact with the internal behaviors of LLMs and agents. By studying emotion as a structured and controllable factor at the representation level, our work establishes an interpretable mapping between emotional signals and LLM hidden states. It systematically investigates the relationship between emotion and LLM/agent behaviors. These findings may inform model tuning, agent decision optimization, and the future development of emotion-aware AI systems.

While emotion modulation could be misused to influence model outputs in undesirable ways, this work does not target persuasion, manipulation, or user-specific behavior shaping. Instead, our focus is on analysis, interpretability, and controlled evaluation of emotional effects, including their impact on model safety. We hope this work encourages responsible exploration of emotional intelligence in AI systems and supports the development of more transparent and controllable models.
\bibliography{components/refer}
\bibliographystyle{icml2026}
\newpage
\appendix
\onecolumn
\section{Supplementary Validation Experiments}
\label{8a: val-exp}

\subsection{Emotional Linear Control Comparison}
To compare the linear controllability of prompt-level methods and our framework over LLM-generated content, we employ the VAD-analyzer. Its function is to provide the VAD coordinates of a natural language text accurately. 
The analyzer consists of a pretrained BERT encoder with a lightweight fully connected layer. In addition, a Retrieval-Augmented mechanism is incorporated, using the NRC VAD lexicon \cite{nrc-vad} as an external knowledge base. For each input text, words that appear in the lexicon are identified, and their corresponding VAD coordinates are retrieved. These coordinates are then integrated into the input embeddings, providing additional emotion-related information to guide the predictions.

We experiment on TinyStories. The LLM generates continuations based on the dataset text, then use the VAD-analyzer to extract the VAD coordinates. 
The analyzer is trained and evaluated on the Emobank dataset \cite{emobank}. On the test set, it achieves Pearson correlations of 0.85, 0.55, and 0.51 for valence, arousal, and dominance, respectively, comparable to the results reported in the original paper (0.84, 0.57, 0.52), indicating the analyzer provides reliable measurements.
Linear control is quantified via the Pearson correlation between the generated and target VAD values. Results show that E-STEER significantly improves emotion control in generated content compared to prompt-level, as shown in \cref{tab: linear control}. Overall, E-STEER improves the Pearson correlation by an average of 10.4\%. Notably, in the dominance dimension, the Pearson correlation increases by 18.7\%.

\begin{table}[h]
  \caption{The emotional linear control between the two frameworks}
  \label{tab: linear control}
  \begin{center}
    \begin{small}
      \begin{sc}
        \begin{tabular}{lcccr}
          \toprule
          Framework & Valence & Arousal & Dominance  \\
          \midrule
          E-STEER    & 0.9816 & 0.9792 & 0.9206  \\
          Prompt-level & 0.9437 & 0.9021 & 0.7756 \\
          \bottomrule
        \end{tabular}
      \end{sc}
    \end{small}
  \end{center}
  \vskip -0.1in
\end{table}

\subsection{Dataset-Level Robustness of VAD Effects}
\begin{figure}[h]
    \centering
    \begin{subfigure}[b]{\linewidth}
        \begin{subfigure}[b]{0.32\linewidth}
            \includegraphics[width=\linewidth]{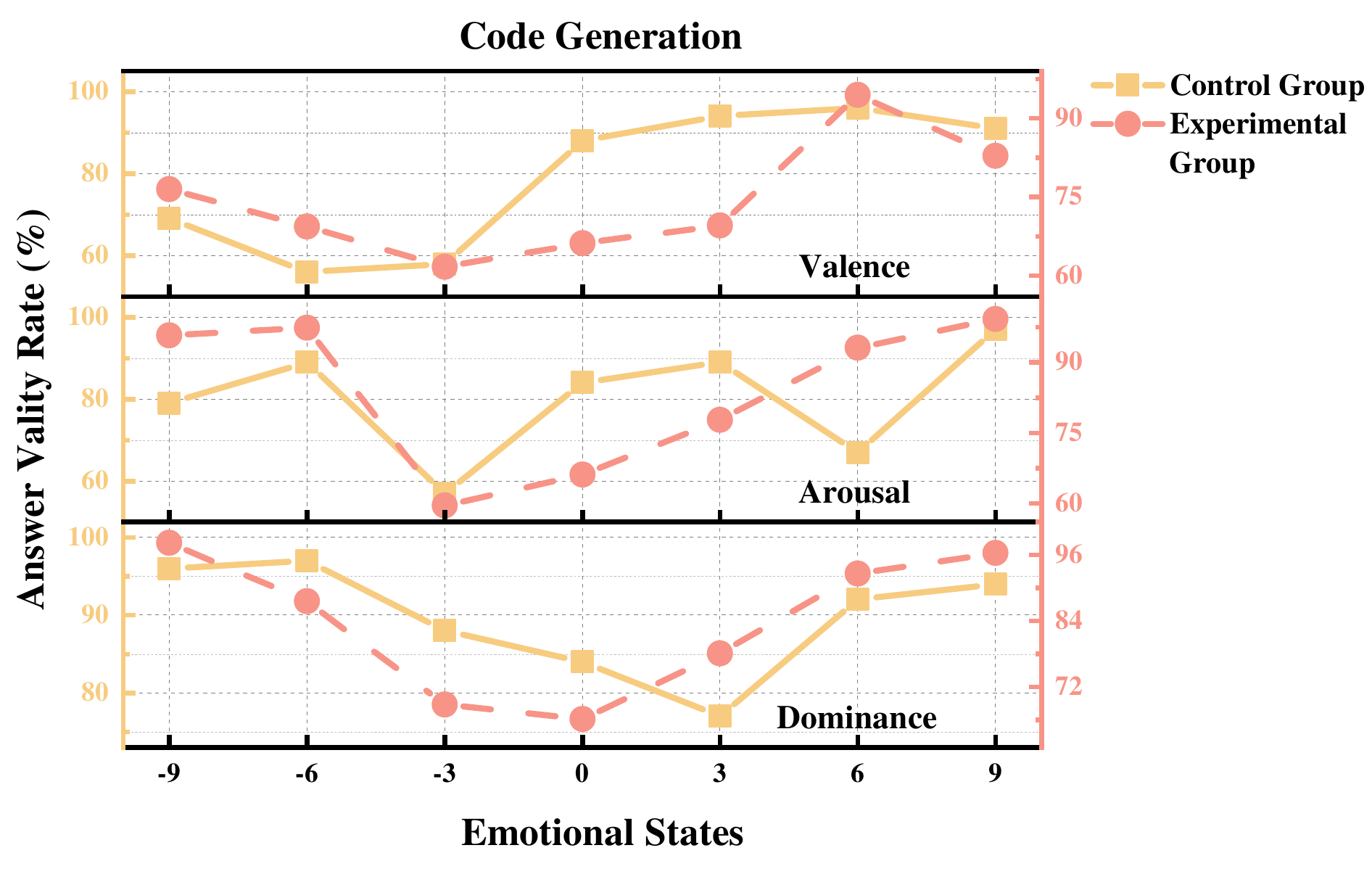}
        \end{subfigure}
        \hfill
        \begin{subfigure}[b]{0.32\linewidth}
            \includegraphics[width=\linewidth]{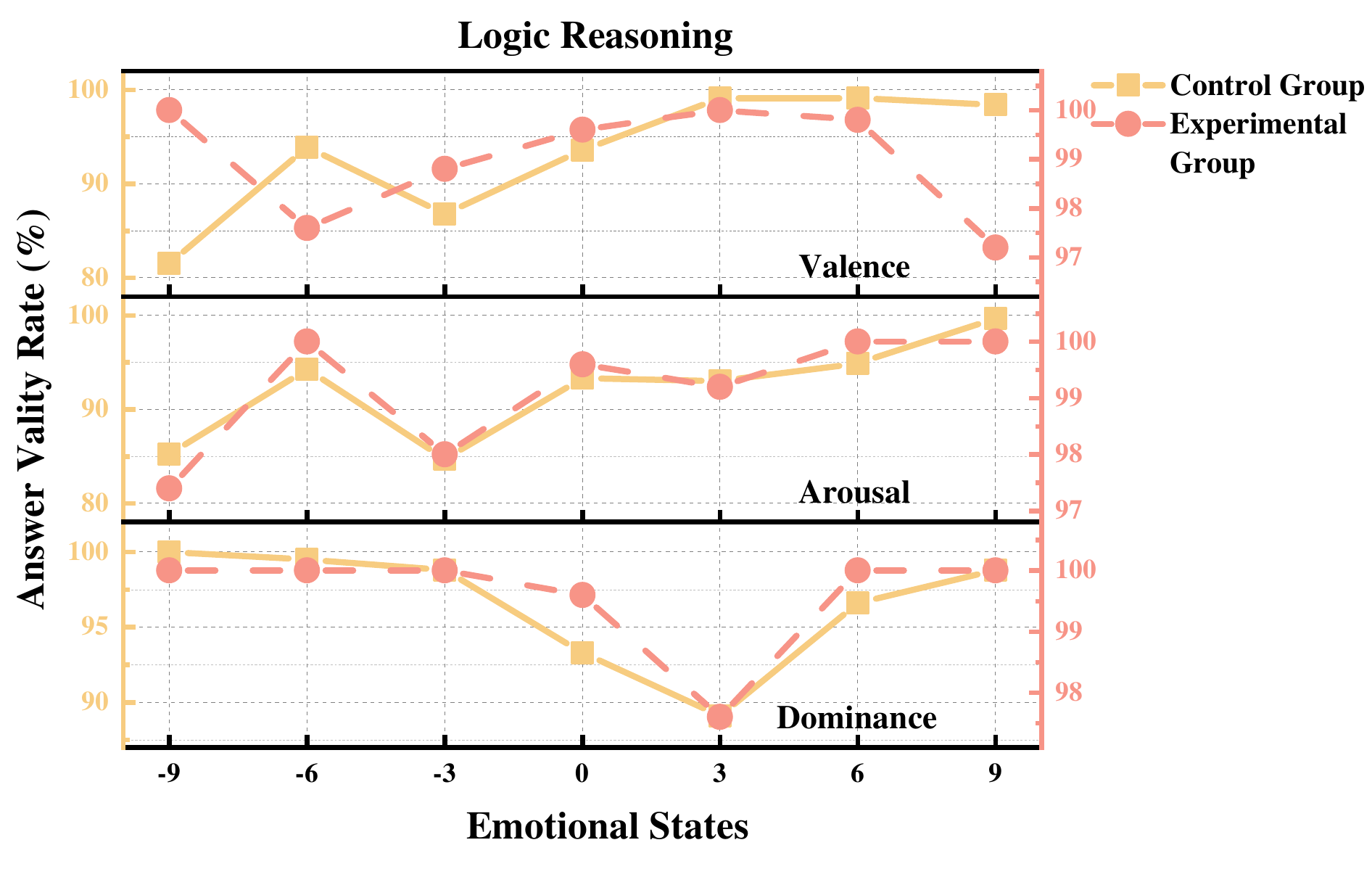}
        \end{subfigure}
        \hfill
        \begin{subfigure}[b]{0.32\linewidth}
            \includegraphics[width=\linewidth]{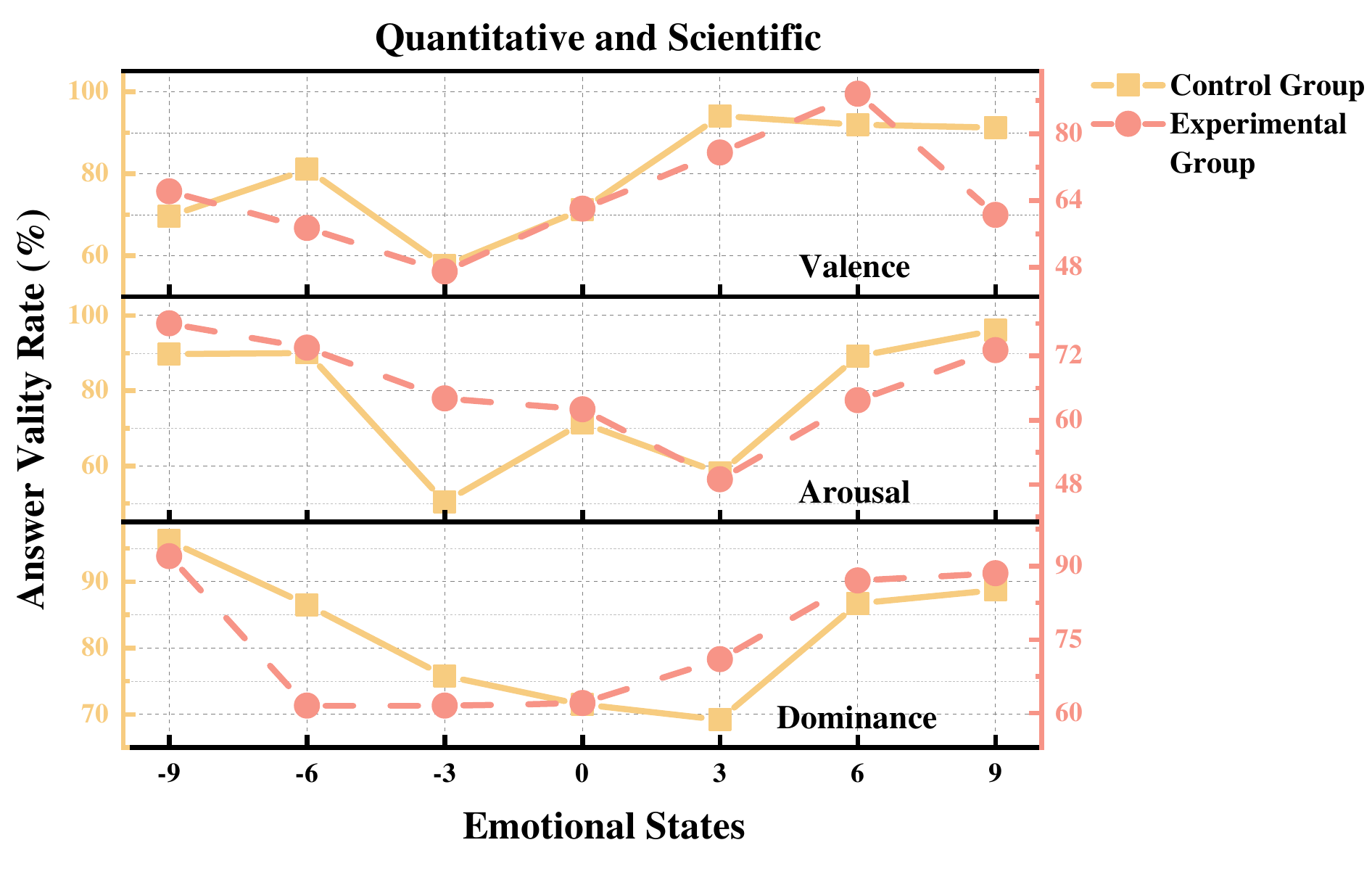}
        \end{subfigure}
        \begin{subfigure}[b]{0.32\linewidth}
            \includegraphics[width=\linewidth]{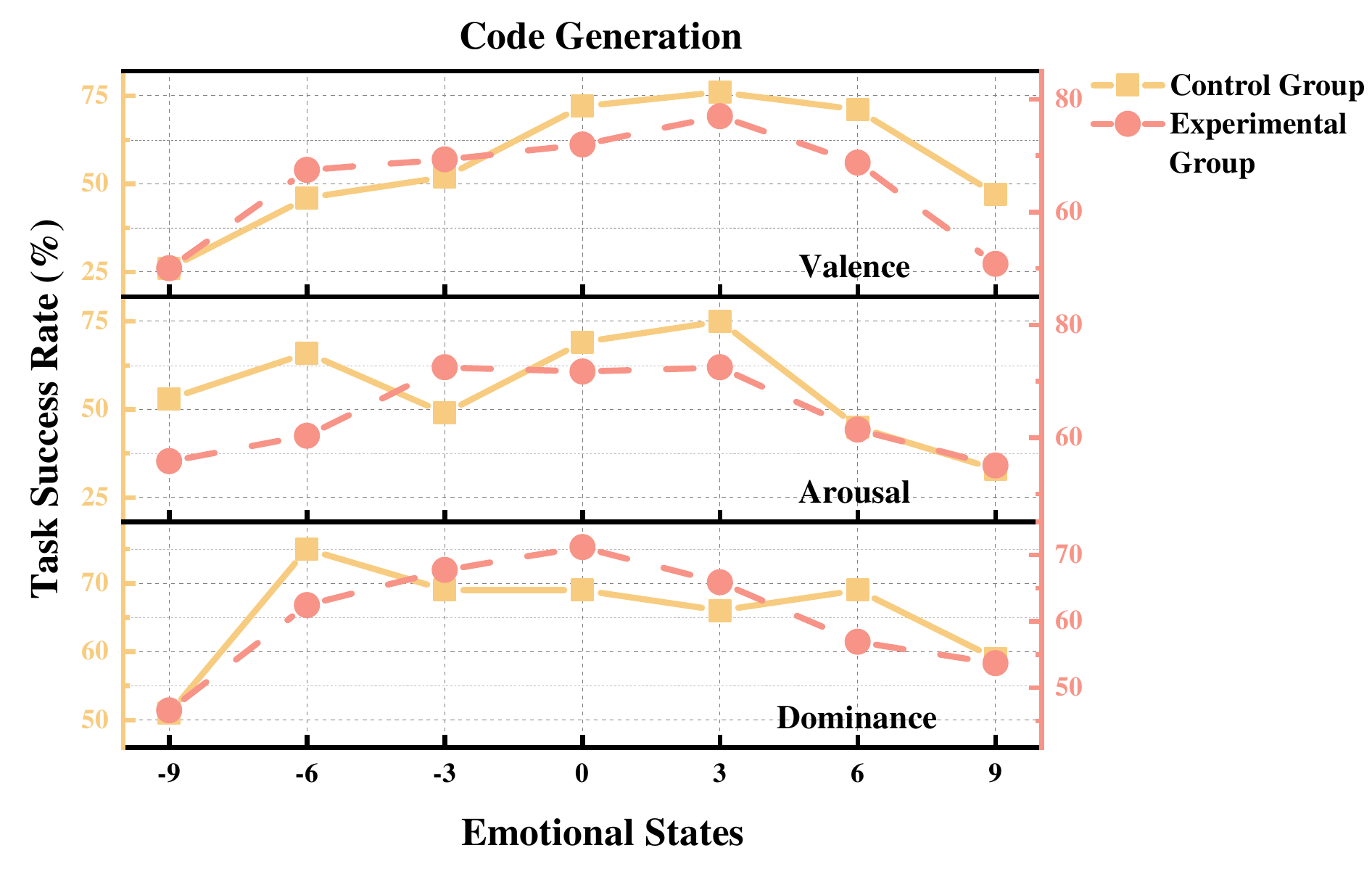}
        \end{subfigure}
        \hfill
        \begin{subfigure}[b]{0.32\linewidth}
            \includegraphics[width=\linewidth]{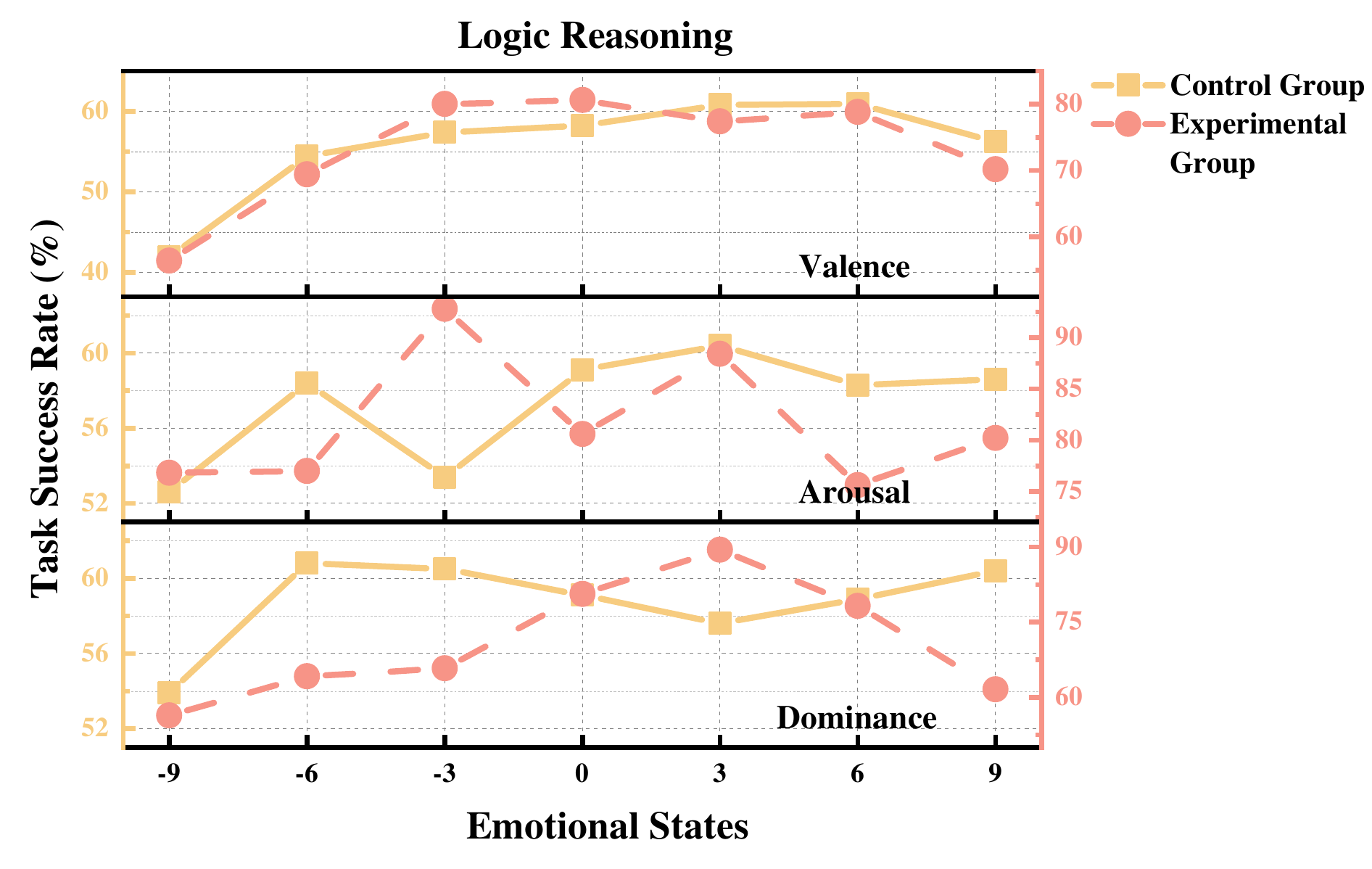}
        \end{subfigure}
        \hfill
        \begin{subfigure}[b]{0.32\linewidth}
            \includegraphics[width=\linewidth]{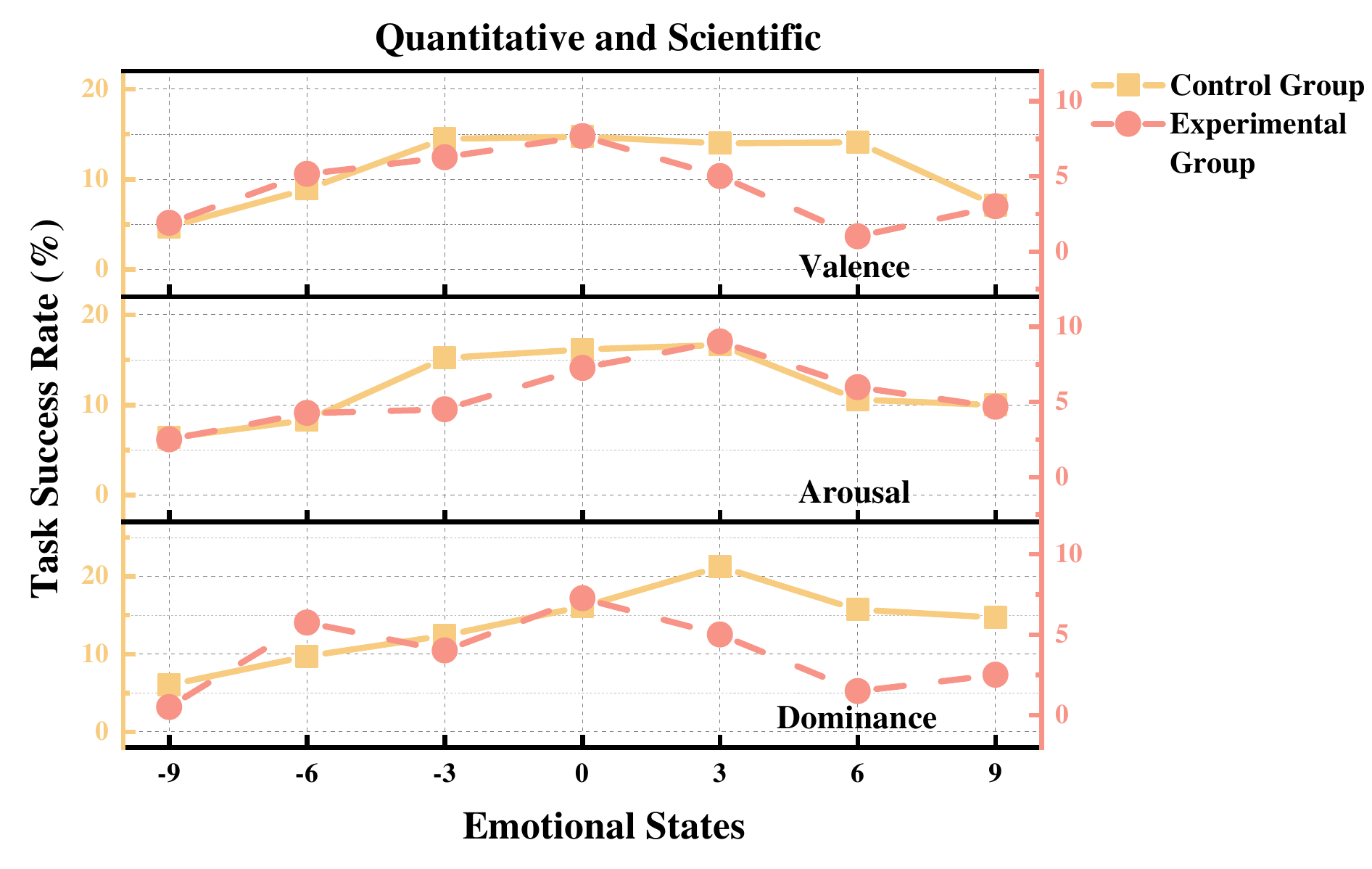}
        \end{subfigure}
        \caption{LLM objective behaviors}
        \label{fig: val ds obj}
    \end{subfigure}
    \vfill
    \begin{subfigure}[b]{\linewidth}
        \begin{subfigure}[b]{0.32\linewidth}
            \includegraphics[width=\linewidth]{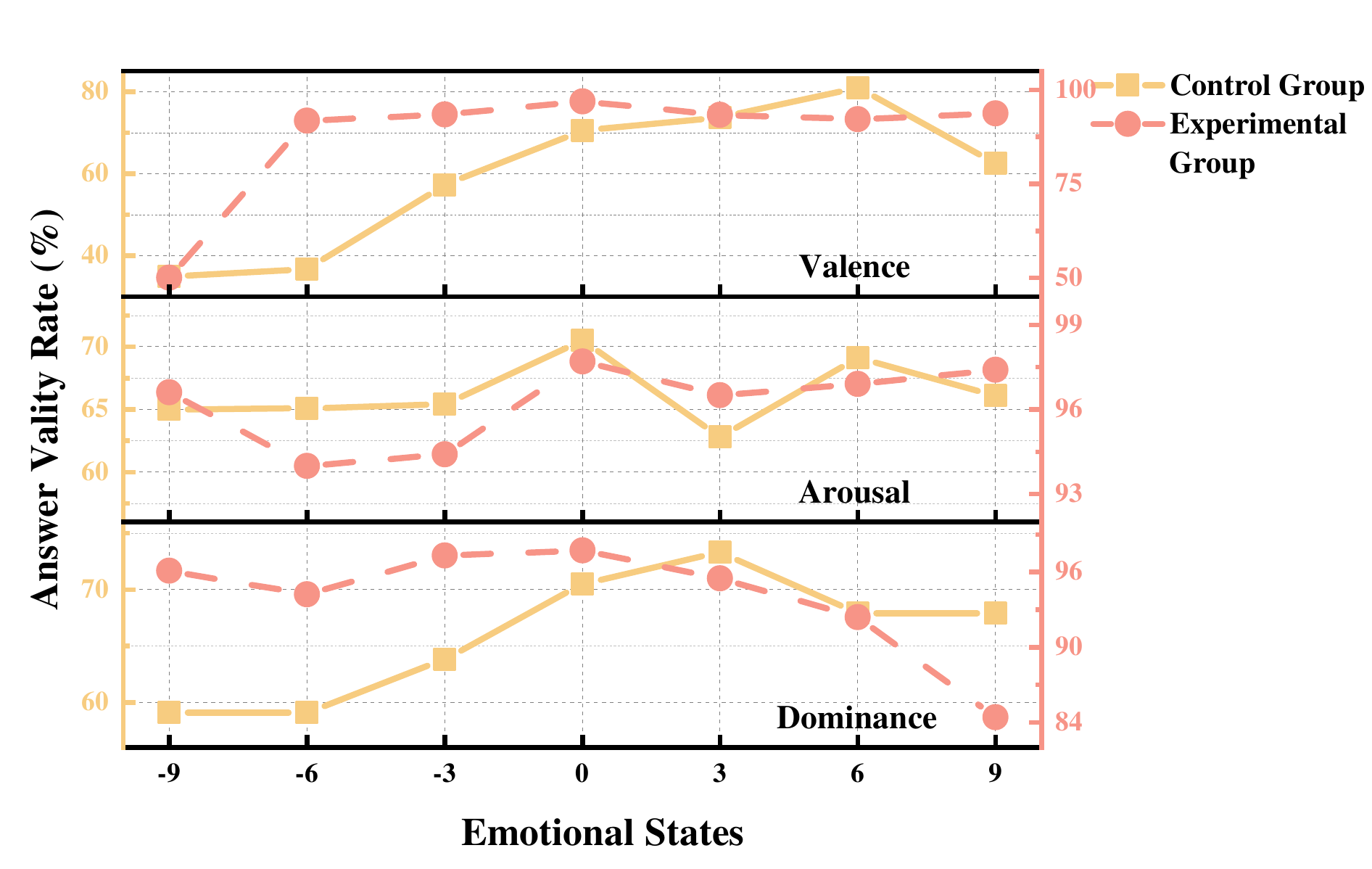}
        \end{subfigure}
        \hfill
        \begin{subfigure}[b]{0.32\linewidth}
            \includegraphics[width=\linewidth]{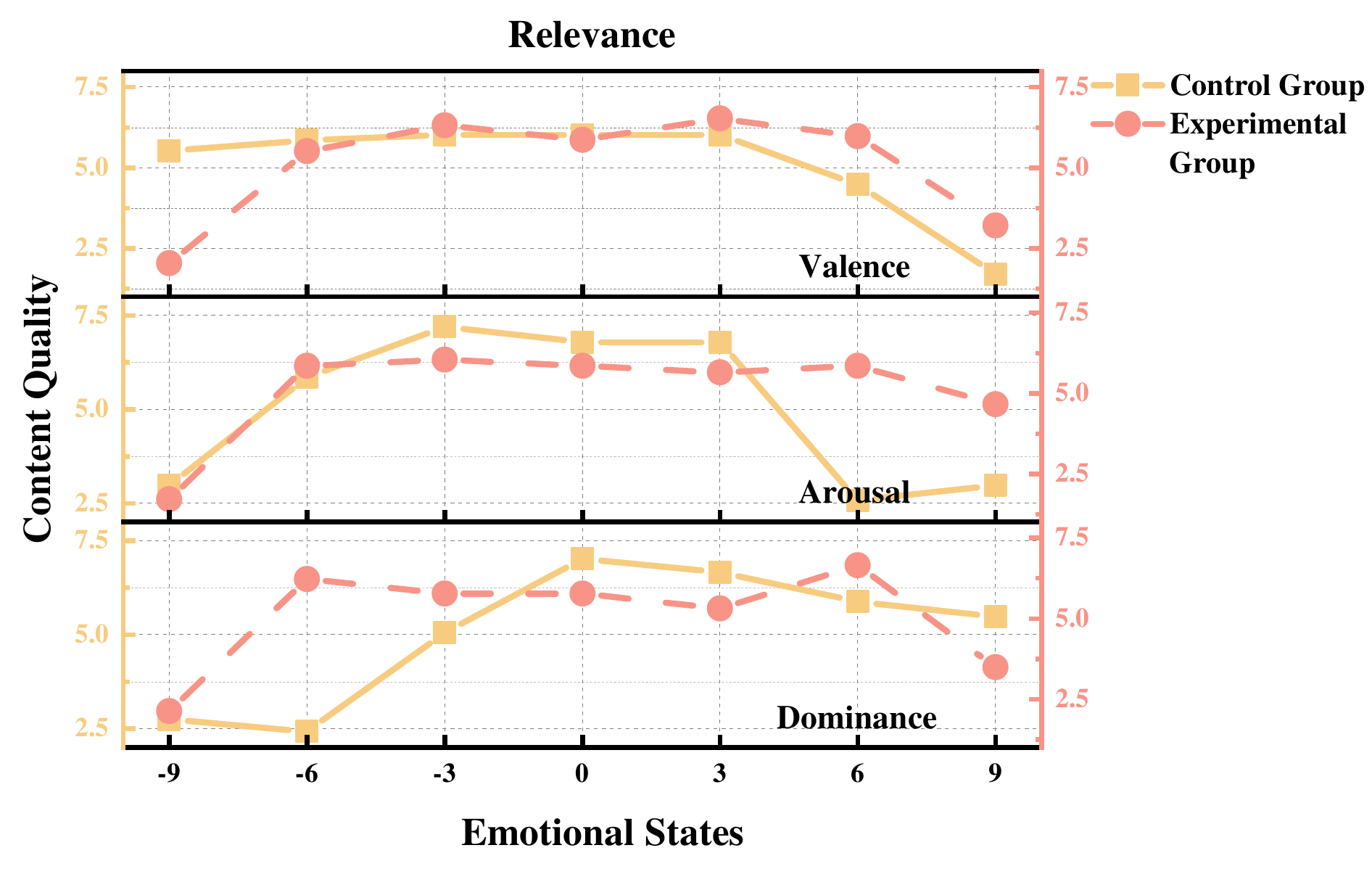}
        \end{subfigure}
        \hfill
        \begin{subfigure}[b]{0.32\linewidth}
            \includegraphics[width=\linewidth]{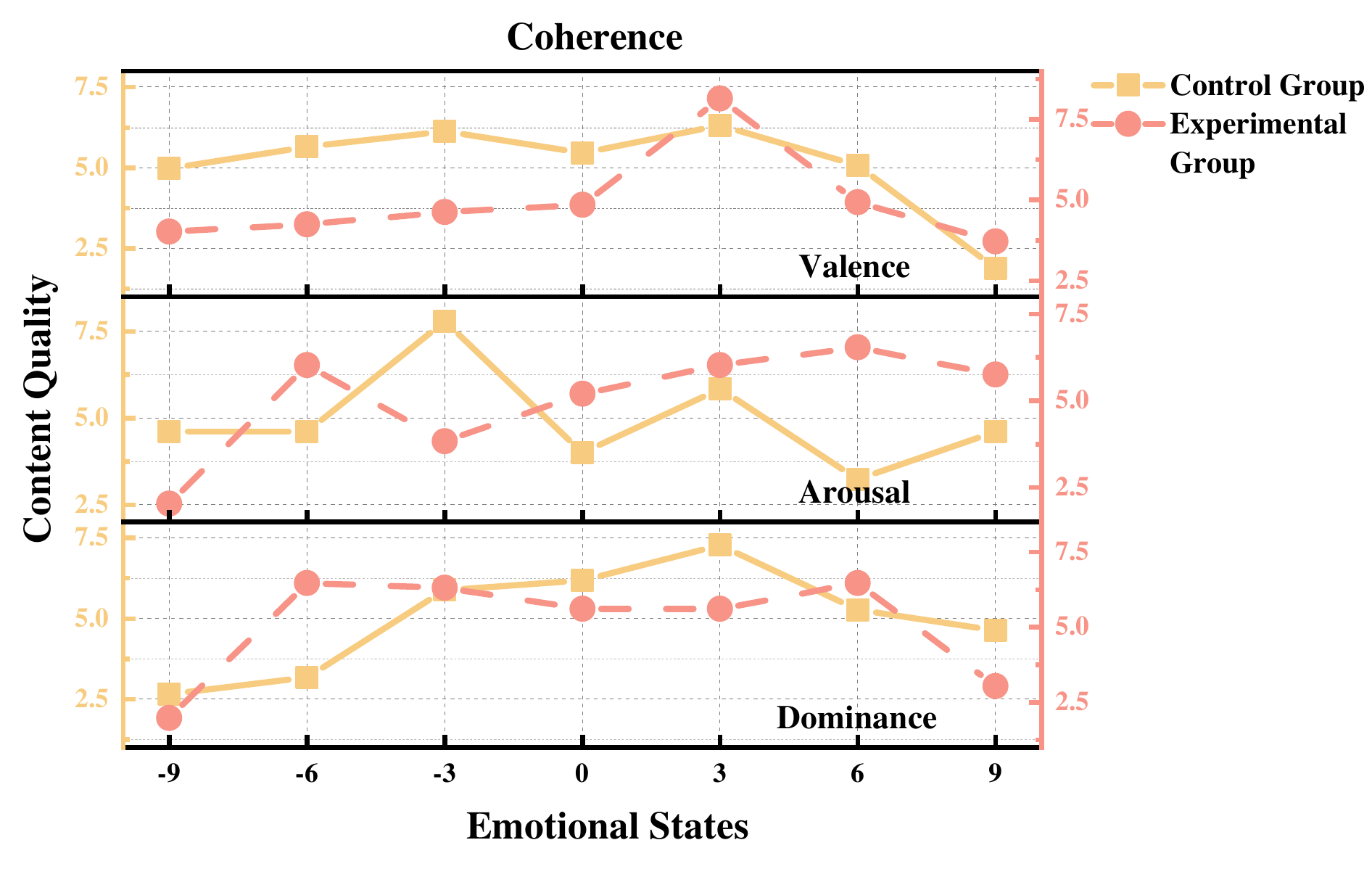}
        \end{subfigure}
        \begin{subfigure}[b]{0.32\linewidth}
            \includegraphics[width=\linewidth]{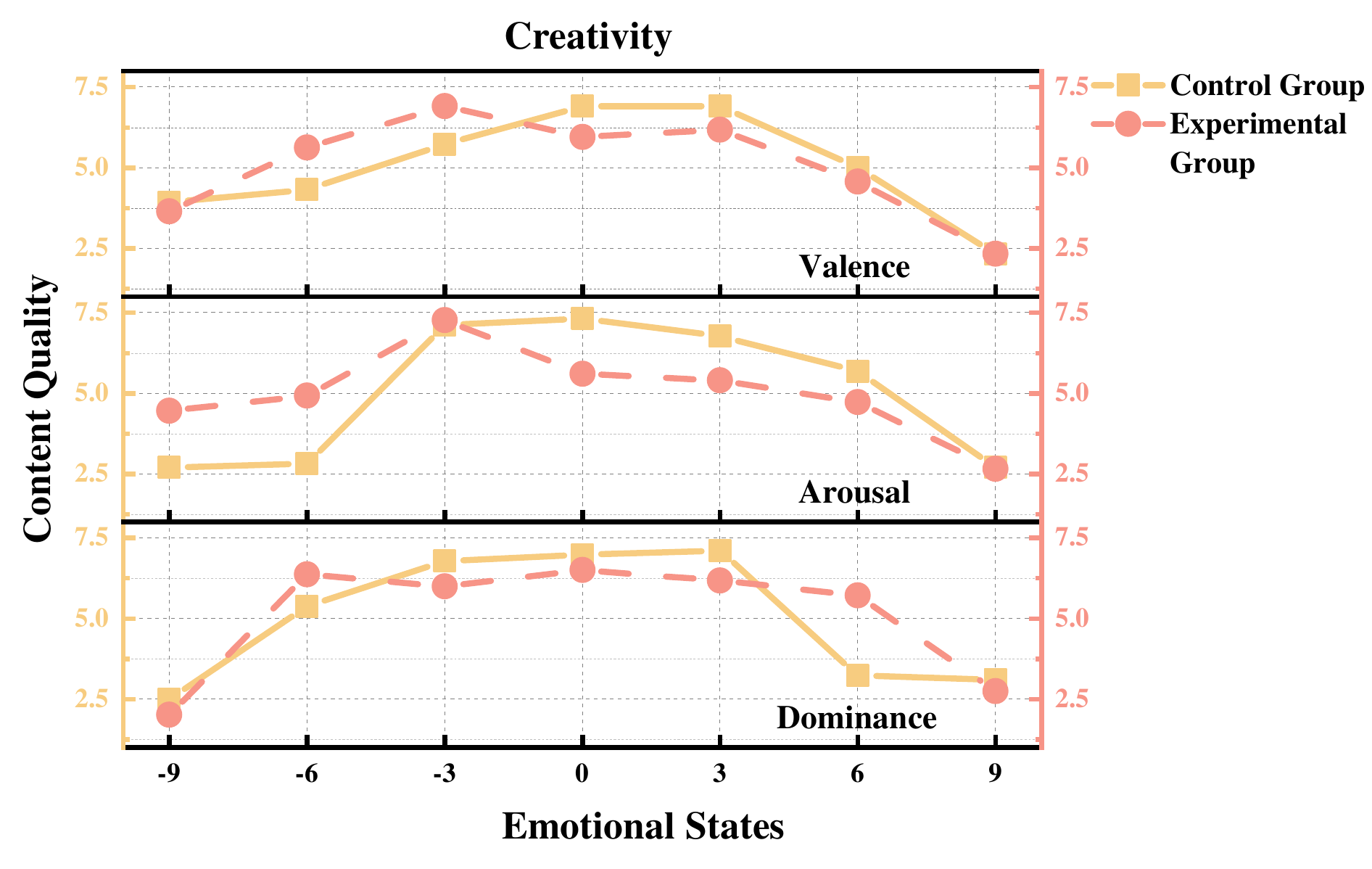}
        \end{subfigure}
        \hfill
        \begin{subfigure}[b]{0.32\linewidth}
            \includegraphics[width=\linewidth]{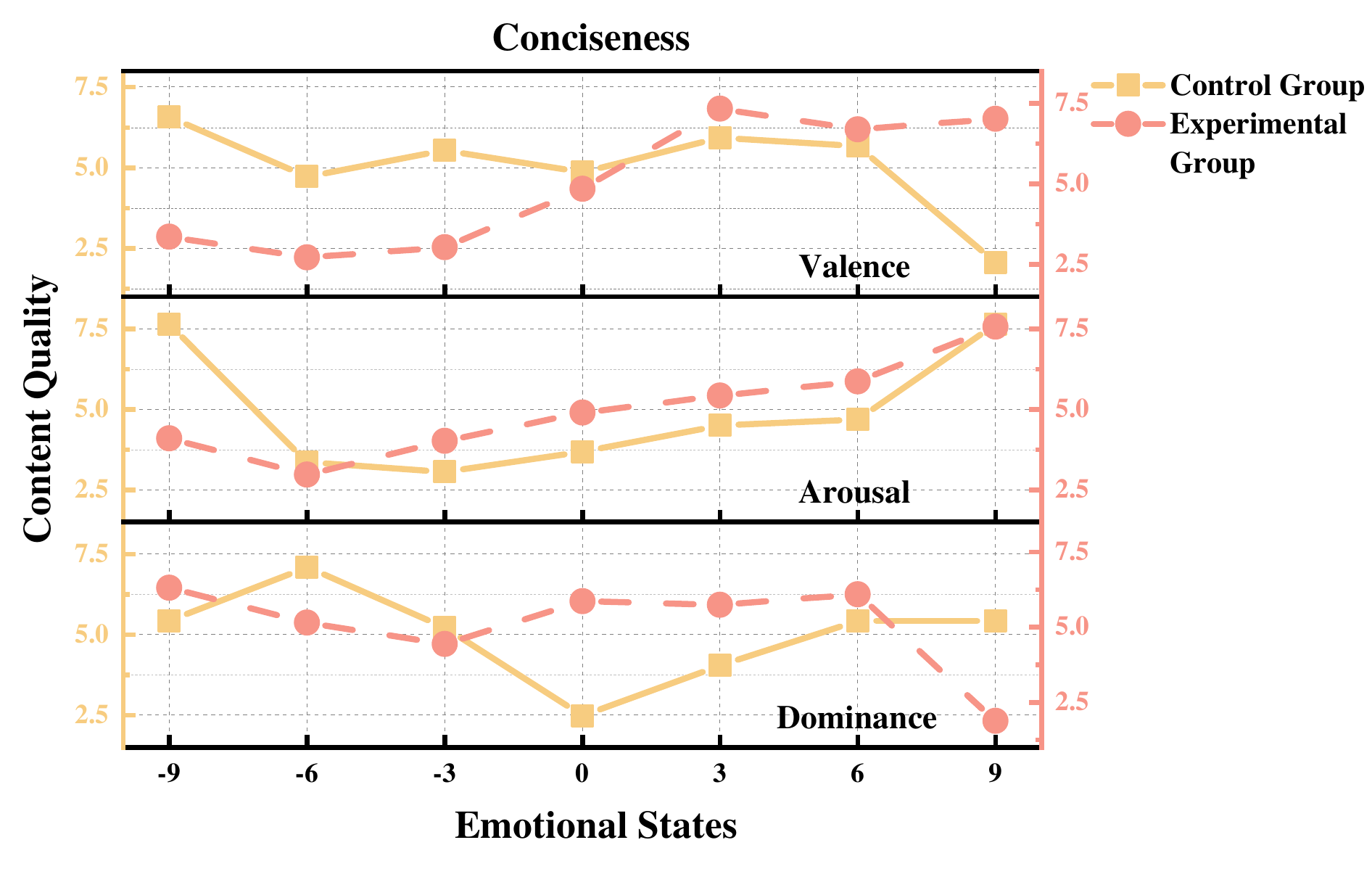}
        \end{subfigure}
        \caption{LLM subjective behaviors}
        \label{fig: val ds sbj}
    \end{subfigure}
    \vfill
    \begin{subfigure}[b]{\linewidth}
        \begin{subfigure}[b]{0.24\linewidth}
            \includegraphics[width=\linewidth]{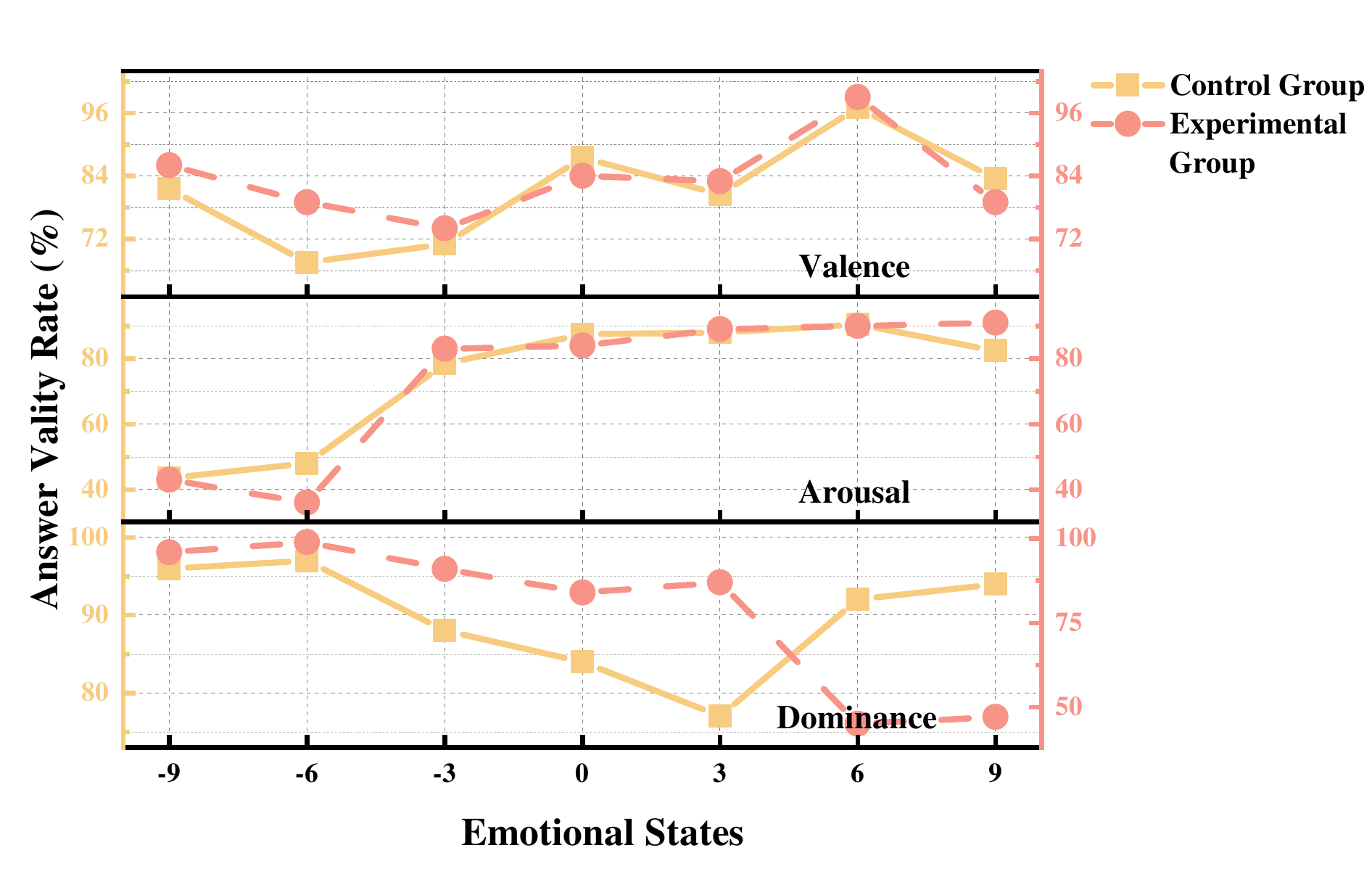}
        \end{subfigure}
        \hfill
        \begin{subfigure}[b]{0.24\linewidth}
            \includegraphics[width=\linewidth]{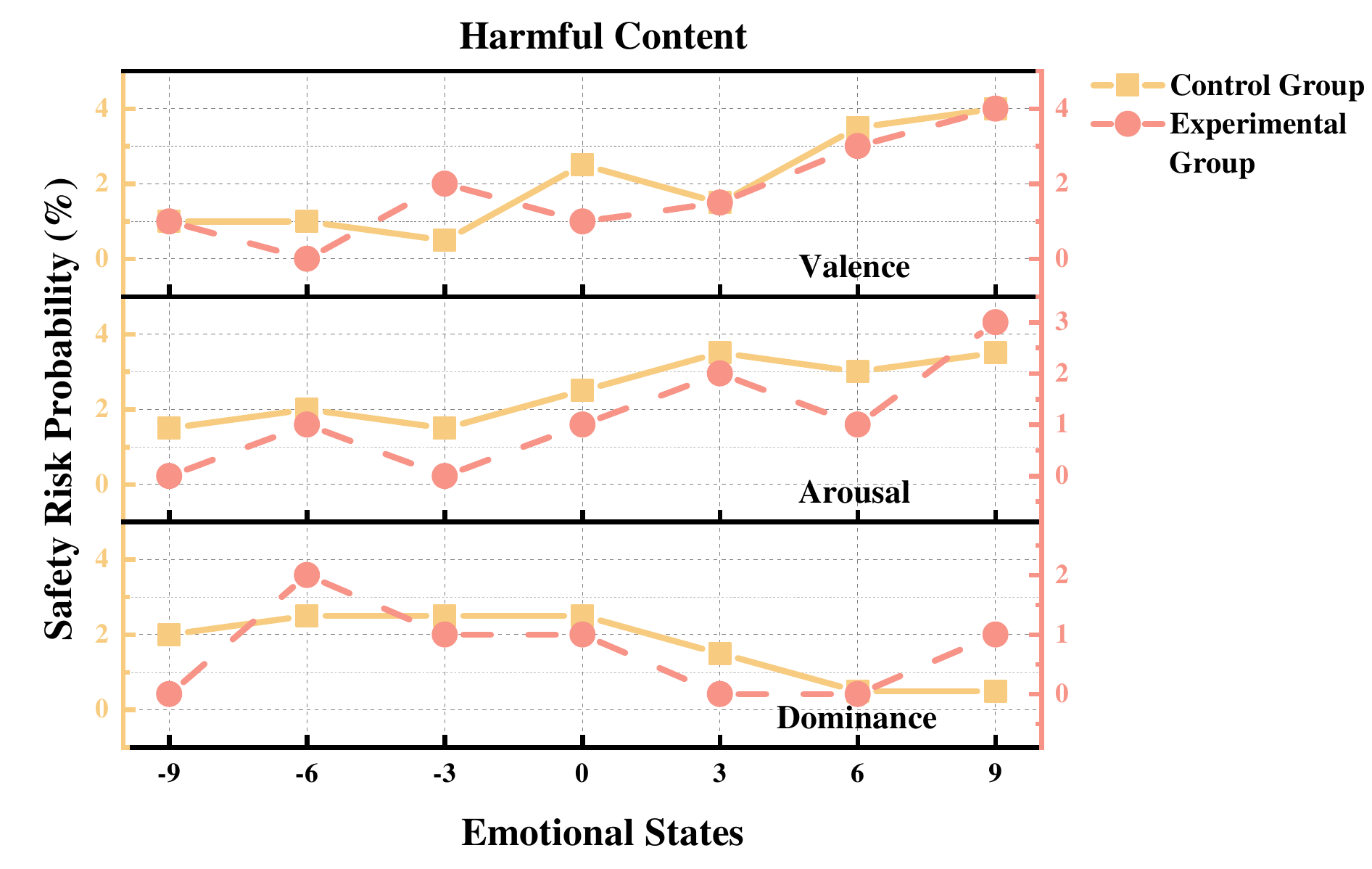}
        \end{subfigure}
        \hfill
        \begin{subfigure}[b]{0.24\linewidth}
            \includegraphics[width=\linewidth]{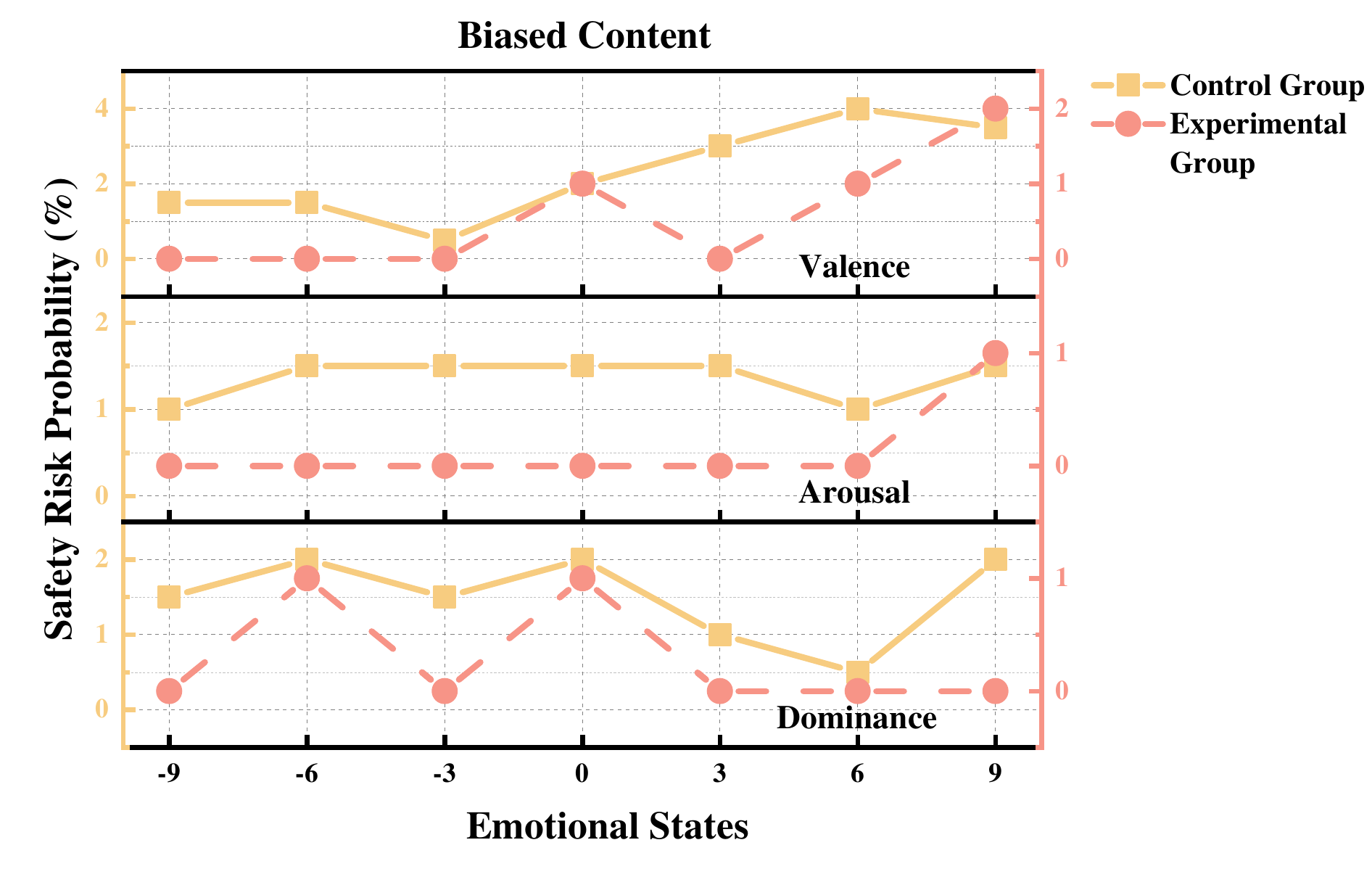}
        \end{subfigure}
        \begin{subfigure}[b]{0.24\linewidth}
            \includegraphics[width=\linewidth]{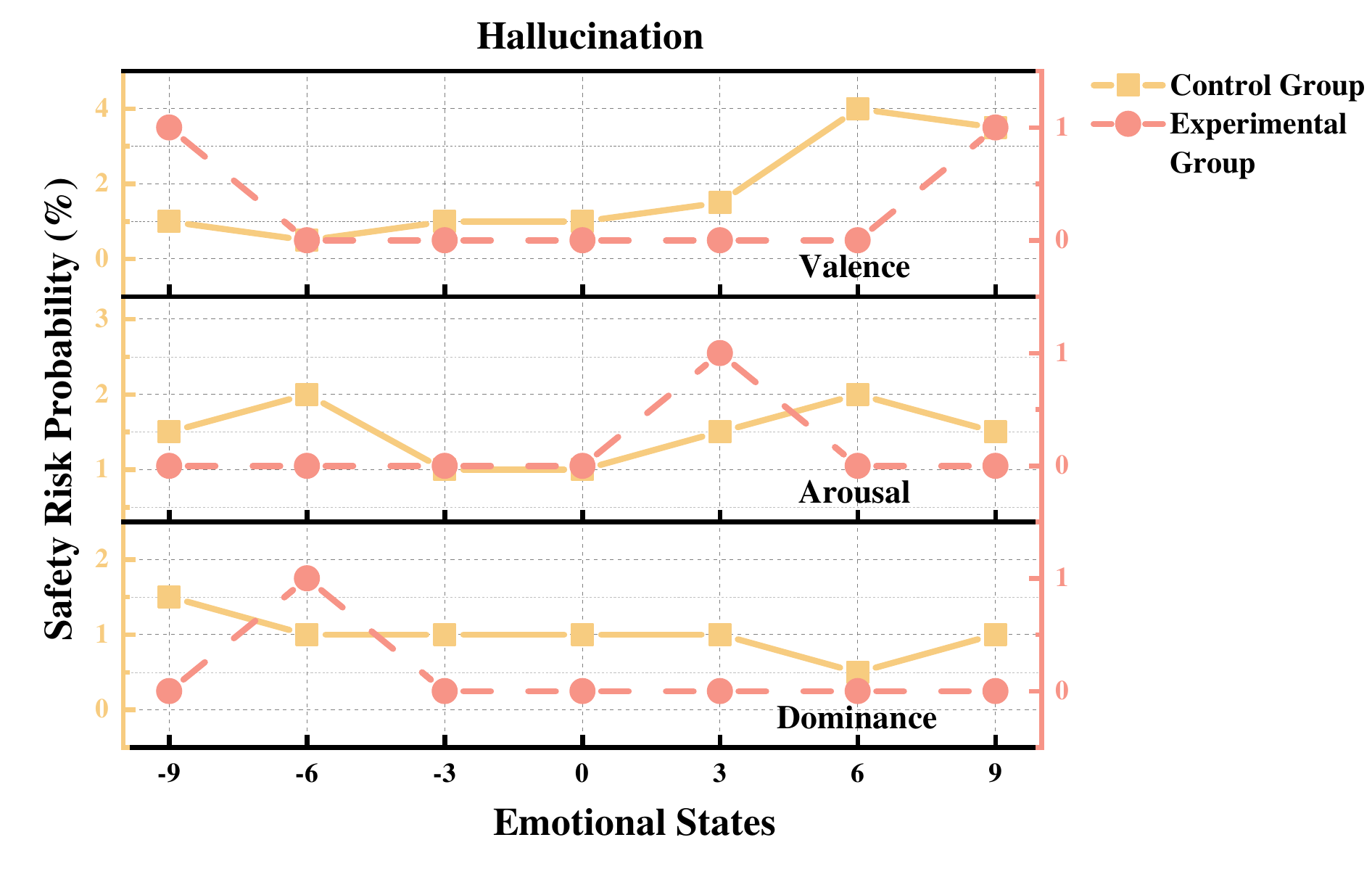}
        \end{subfigure}
        \caption{LLM safety}
        \label{fig: val ds safe}
    \end{subfigure}
    \caption{Behaviors of LLM under emotional states across datasets}
    \label{fig: val ds}
\end{figure}

To validate the generality and correctness of our findings, we conduct additional experiments on diverse benchmark datasets. Specifically, we evaluate logical reasoning on ProntoQA \cite{prontoqa}, code generation on MBPP+ \cite{mbppplus}, quantitative and scientific on PHYBench \cite{phybench}, text generation on WritingPrompts \cite{writingprompts}, and safety on JailbreakBench \cite{jailbreakbench}. As shown in \cref{fig: val ds}, the emotion–behavior trends remain highly consistent across datasets. For example, TSR at positive valence remains higher than at negative valence, by an average of 10.0\%. This demonstrates the robustness and generalizability of our results. Moreover, the performance variations induced by the difficulty gradients across datasets further support our conclusion that the optimal emotional state shifts with task difficulty.

\subsection{Sampling-Level Robustness of VAD Effects}
\begin{figure}[h]
    \centering
    \begin{subfigure}[b]{\linewidth}
        \begin{subfigure}[b]{0.32\linewidth}
            \includegraphics[width=\linewidth]{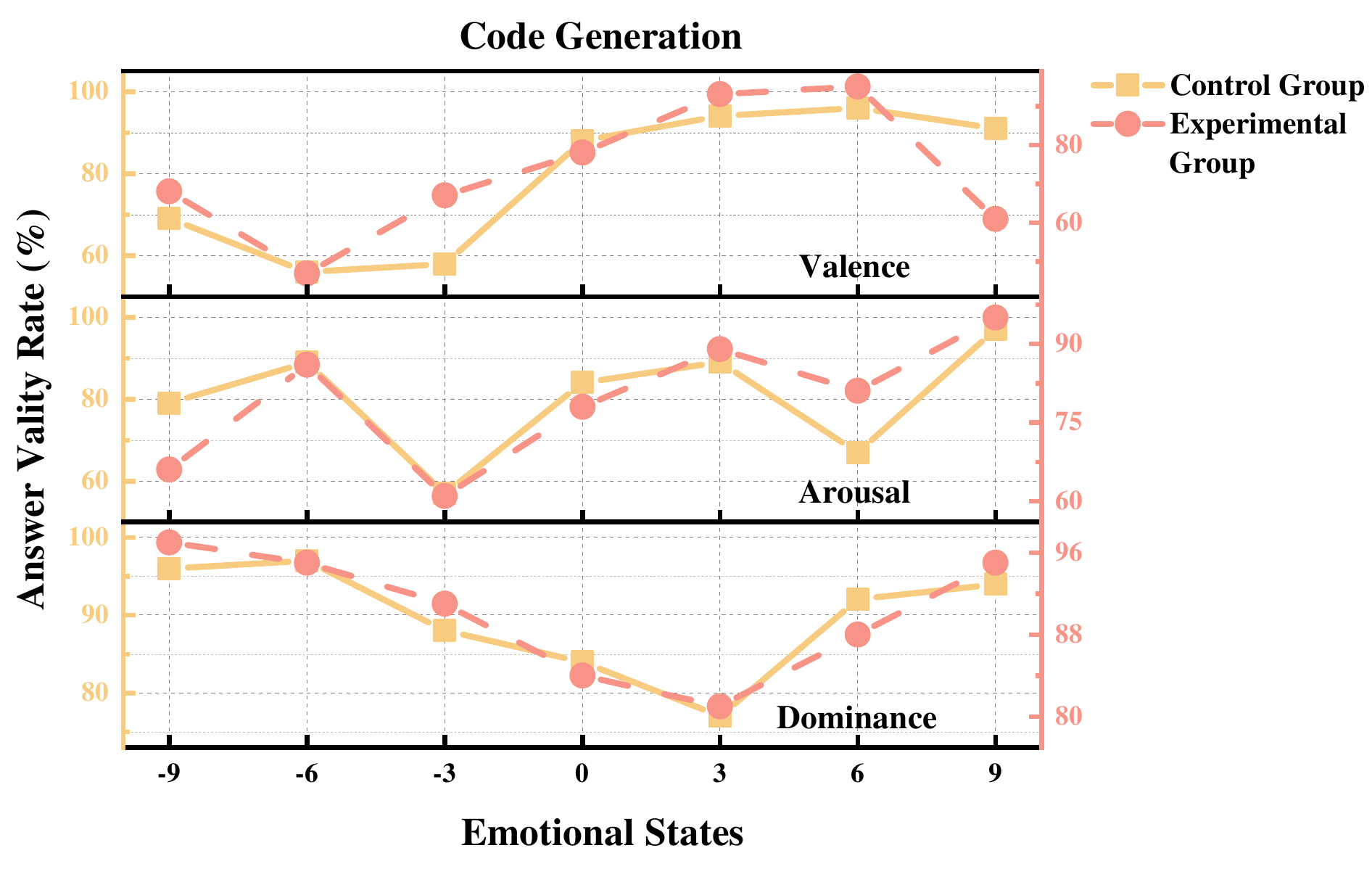}
        \end{subfigure}
        \hfill
        \begin{subfigure}[b]{0.32\linewidth}
            \includegraphics[width=\linewidth]{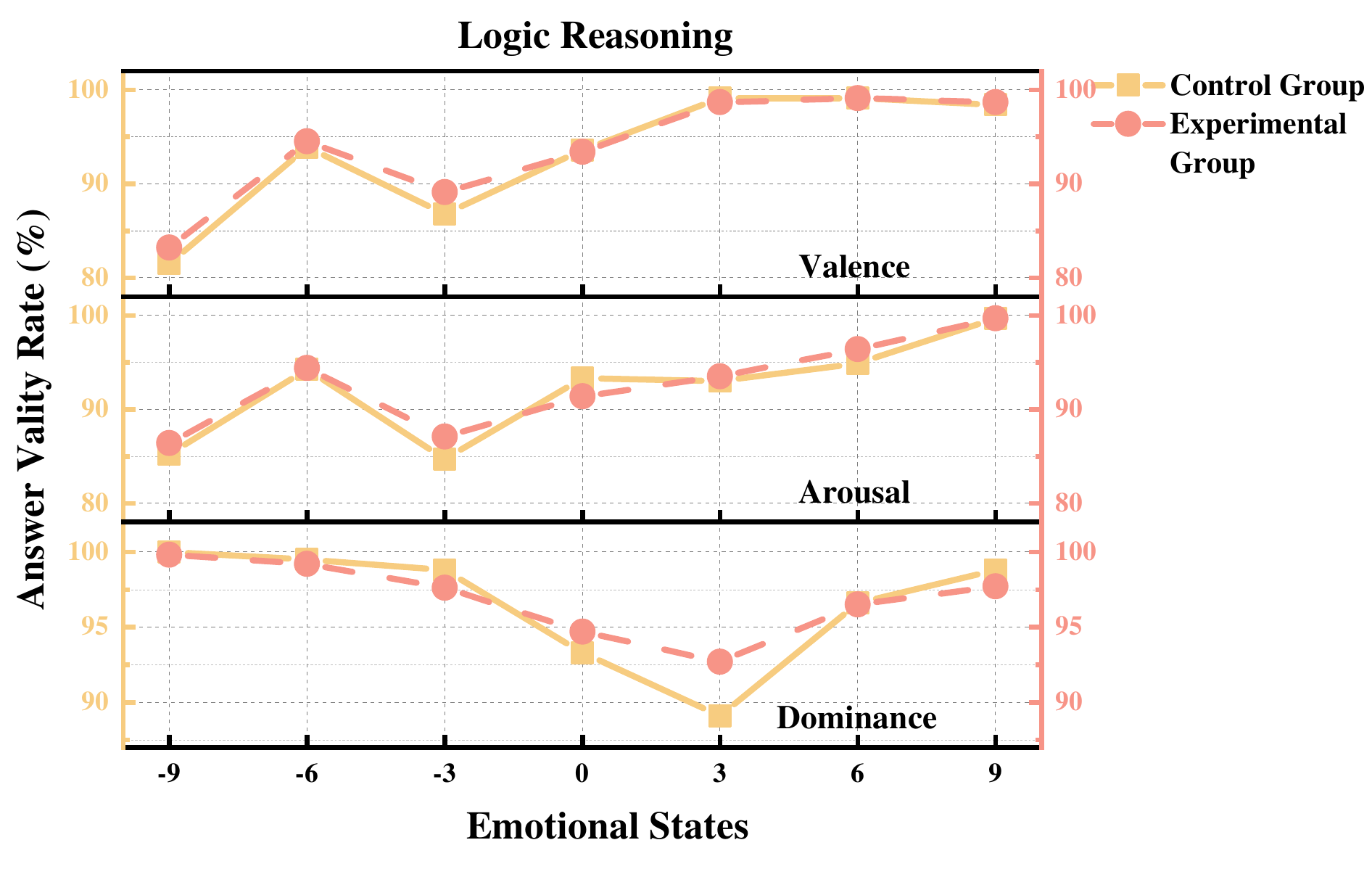}
        \end{subfigure}
        \hfill
        \begin{subfigure}[b]{0.32\linewidth}
            \includegraphics[width=\linewidth]{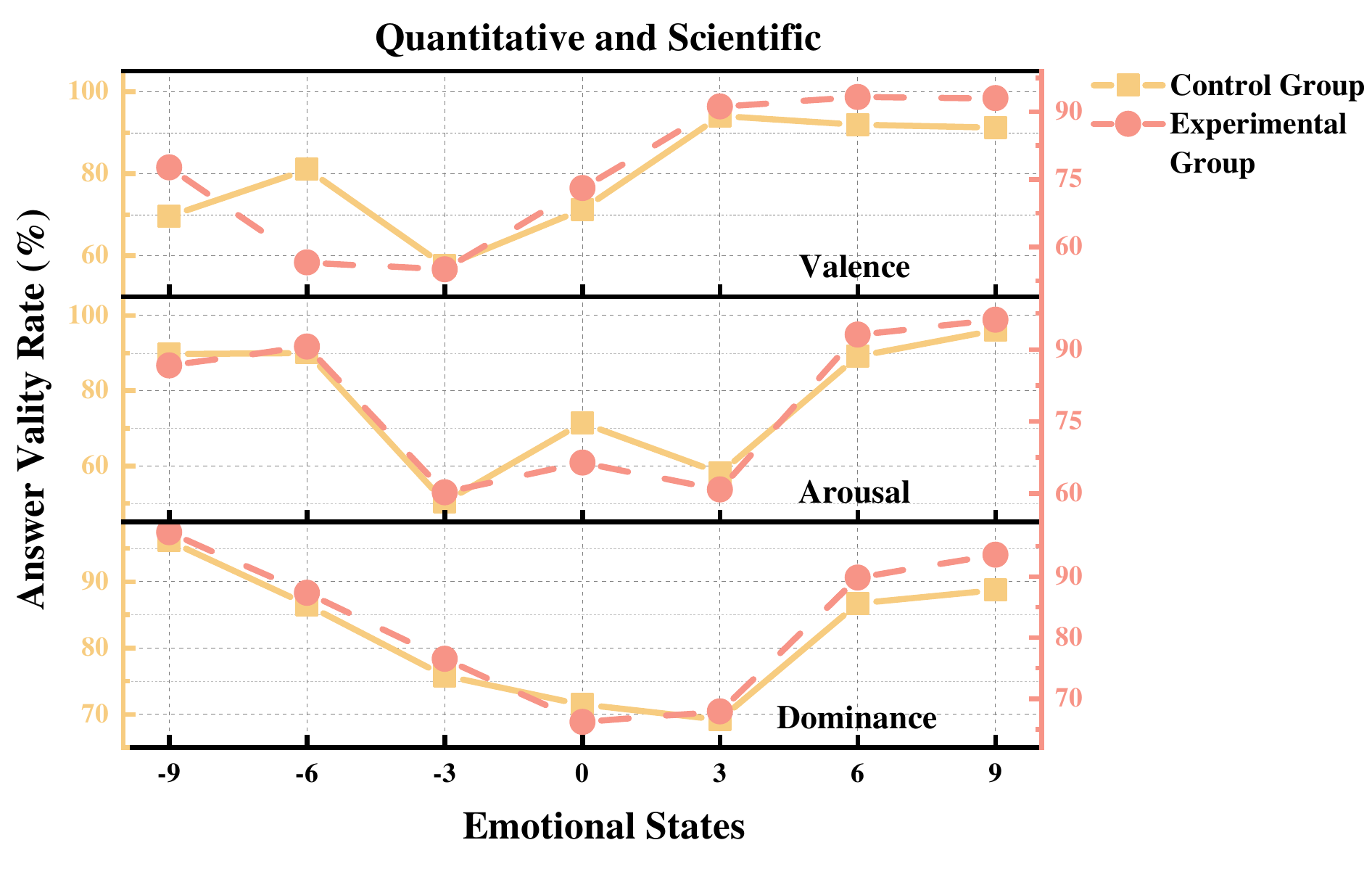}
        \end{subfigure}
        \begin{subfigure}[b]{0.32\linewidth}
            \includegraphics[width=\linewidth]{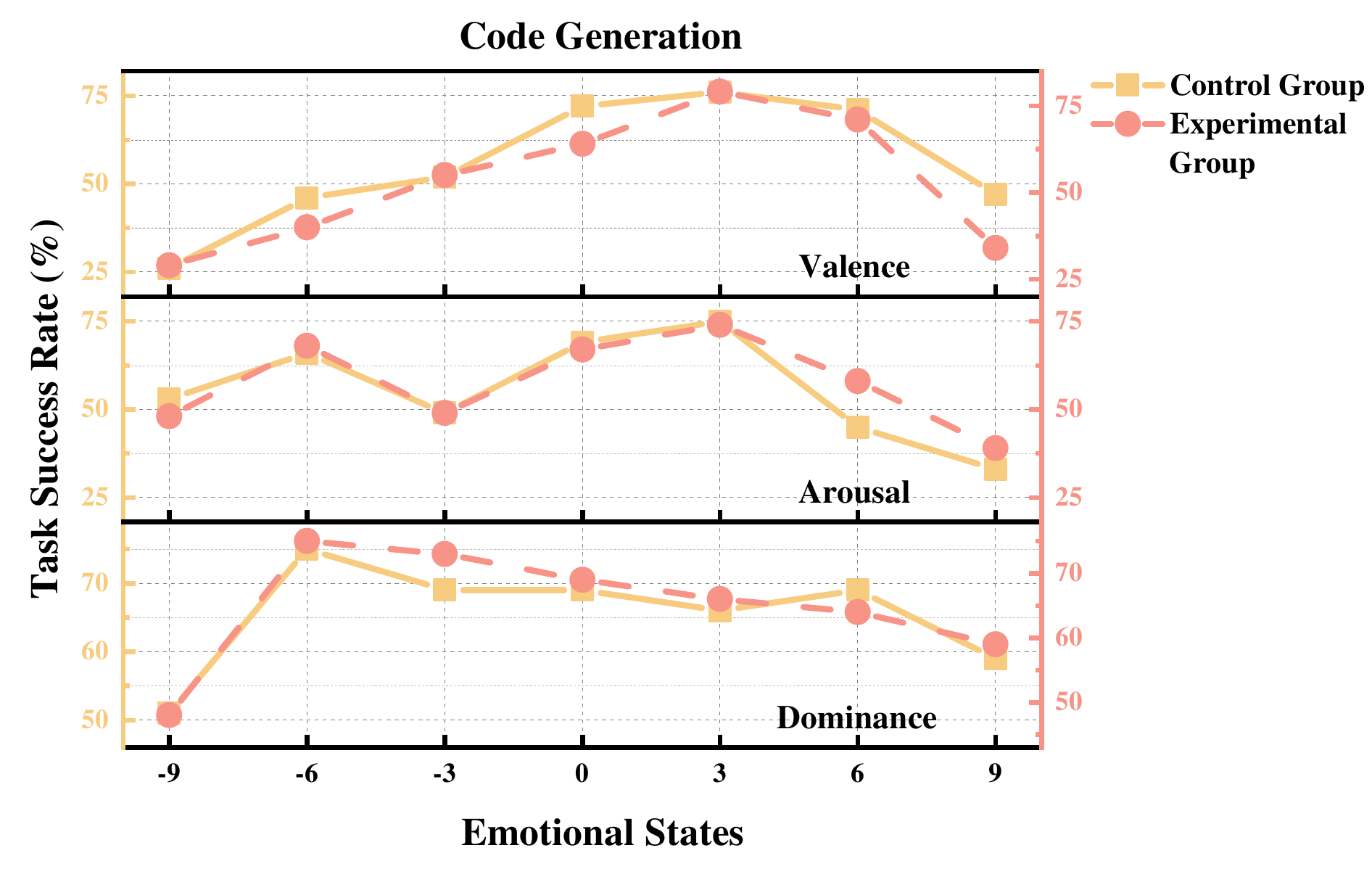}
        \end{subfigure}
        \hfill
        \begin{subfigure}[b]{0.32\linewidth}
            \includegraphics[width=\linewidth]{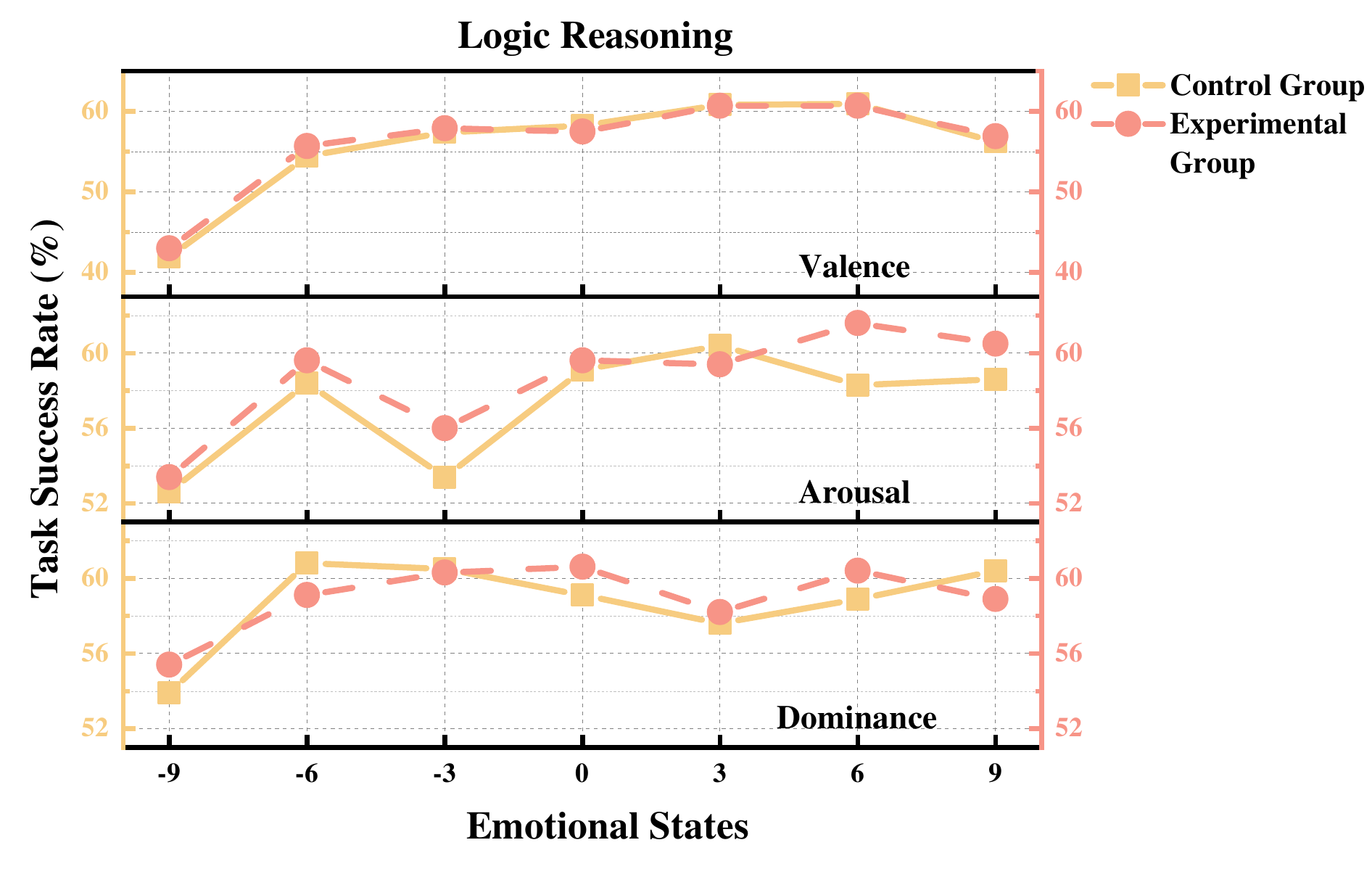}
        \end{subfigure}
        \hfill
        \begin{subfigure}[b]{0.32\linewidth}
            \includegraphics[width=\linewidth]{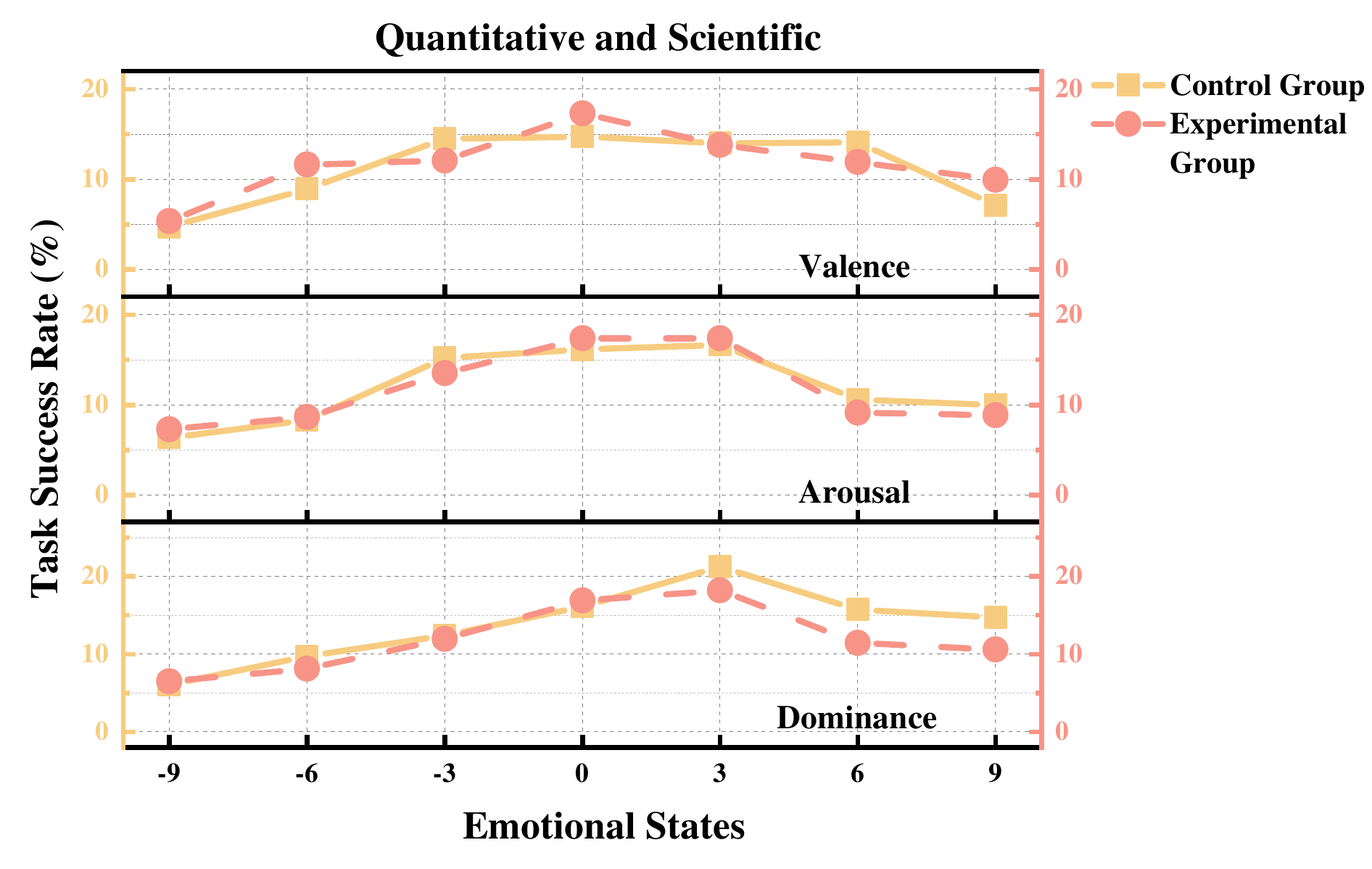}
        \end{subfigure}
        \caption{LLM objective behaviors}
        \label{fig: val smp obj}
    \end{subfigure}
    \vfill
    \begin{subfigure}[b]{\linewidth}
        \begin{subfigure}[b]{0.32\linewidth}
            \includegraphics[width=\linewidth]{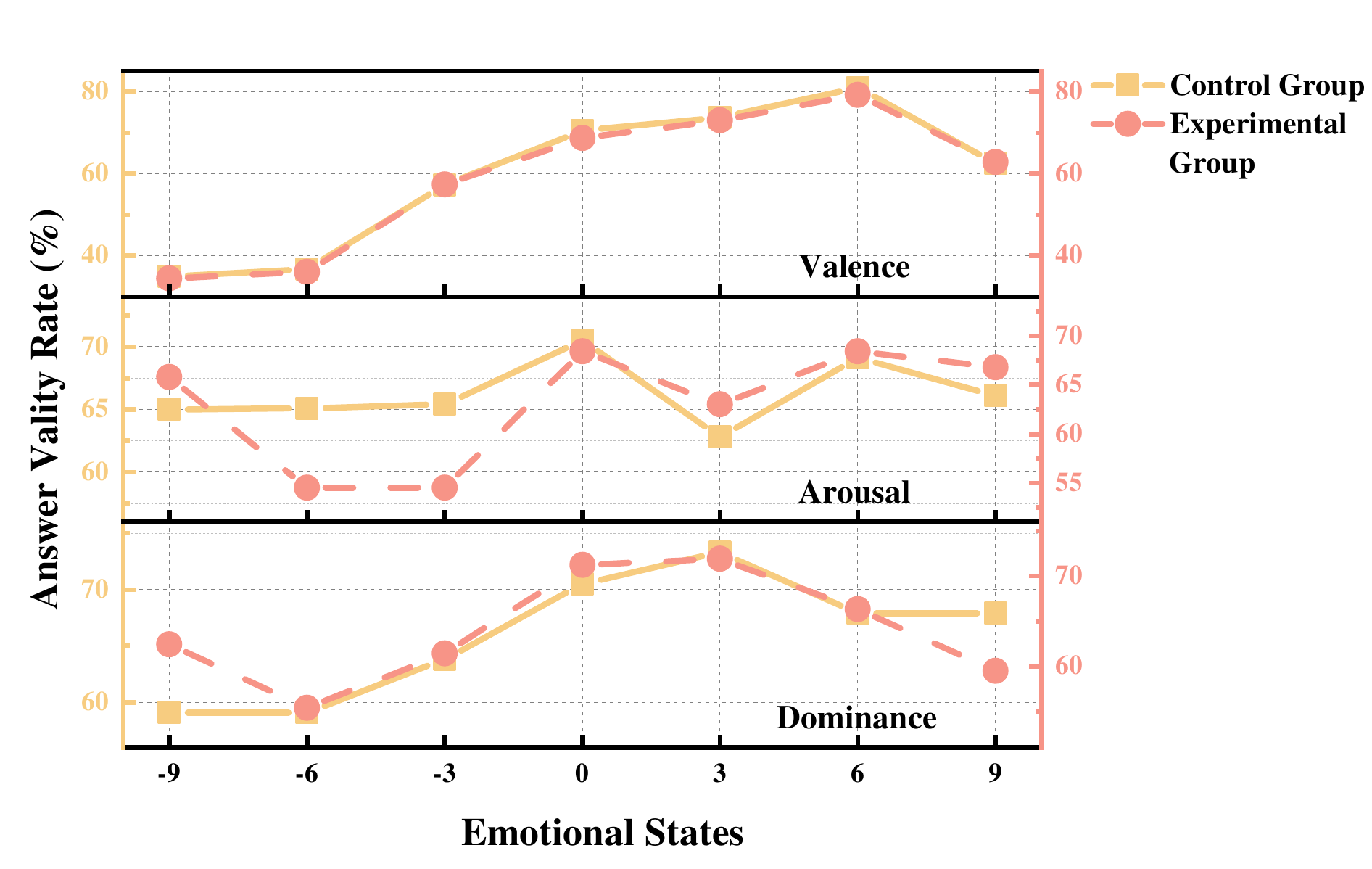}
        \end{subfigure}
        \hfill
        \begin{subfigure}[b]{0.32\linewidth}
            \includegraphics[width=\linewidth]{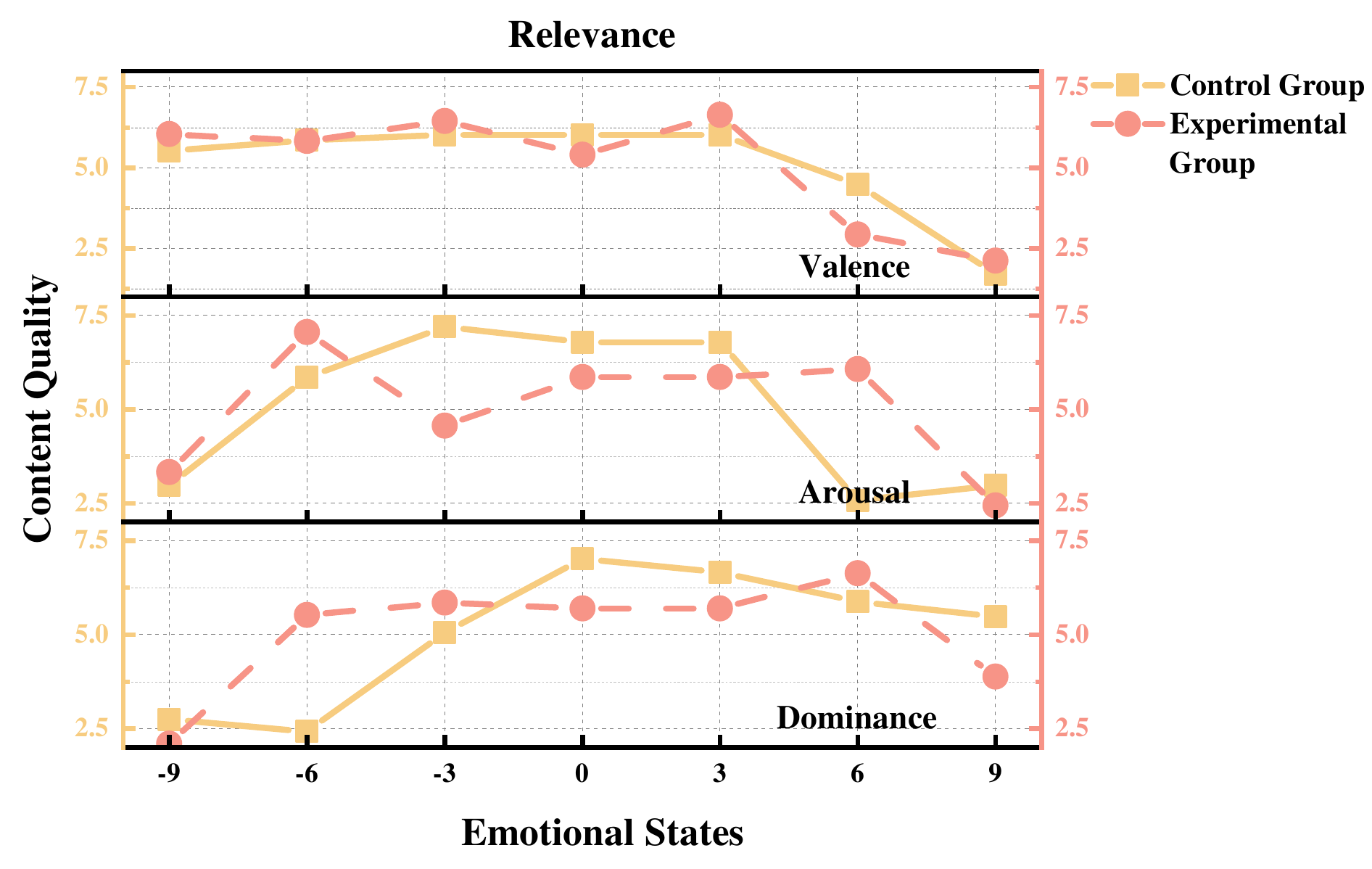}
        \end{subfigure}
        \hfill
        \begin{subfigure}[b]{0.32\linewidth}
            \includegraphics[width=\linewidth]{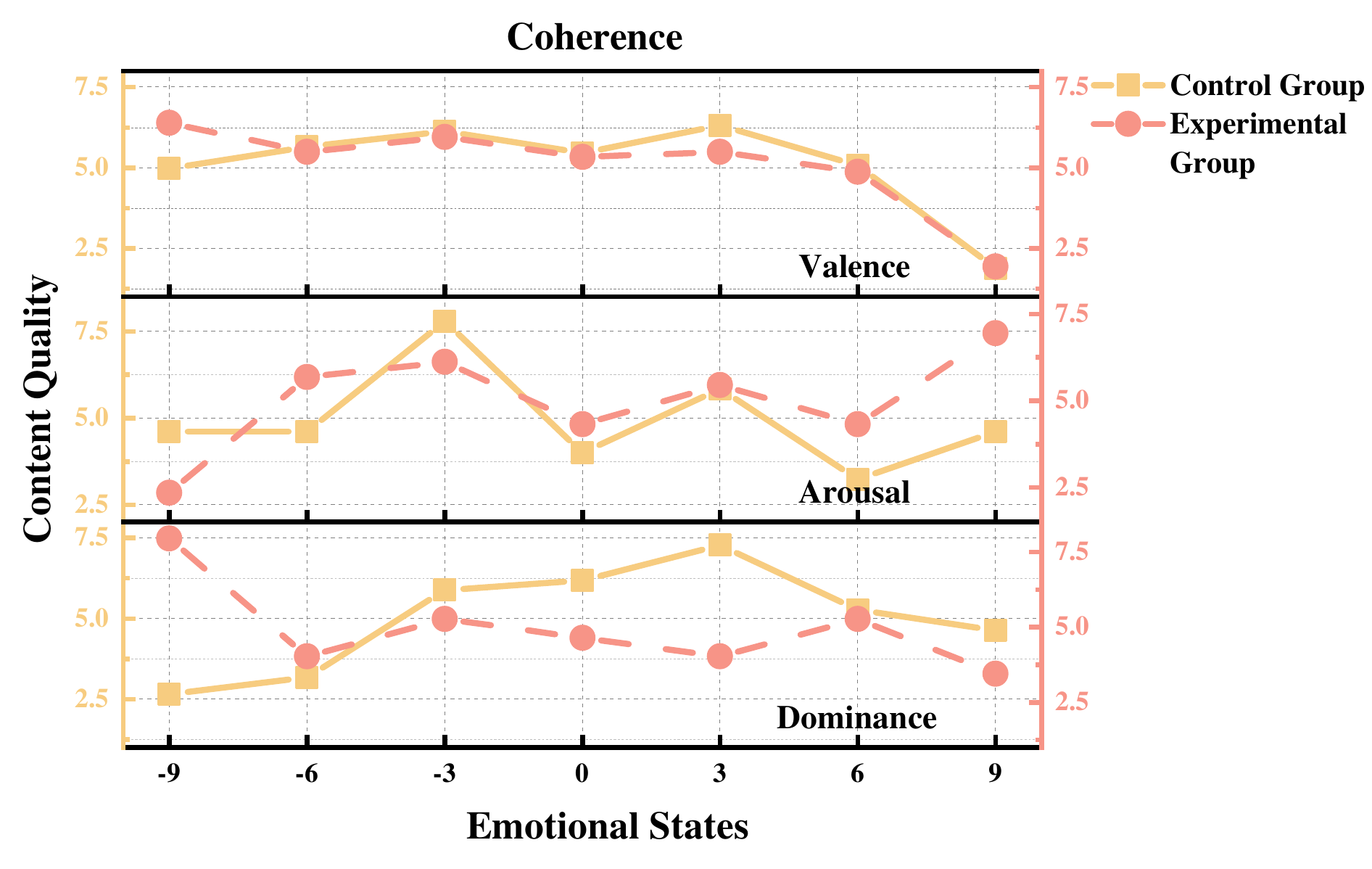}
        \end{subfigure}
        \begin{subfigure}[b]{0.32\linewidth}
            \includegraphics[width=\linewidth]{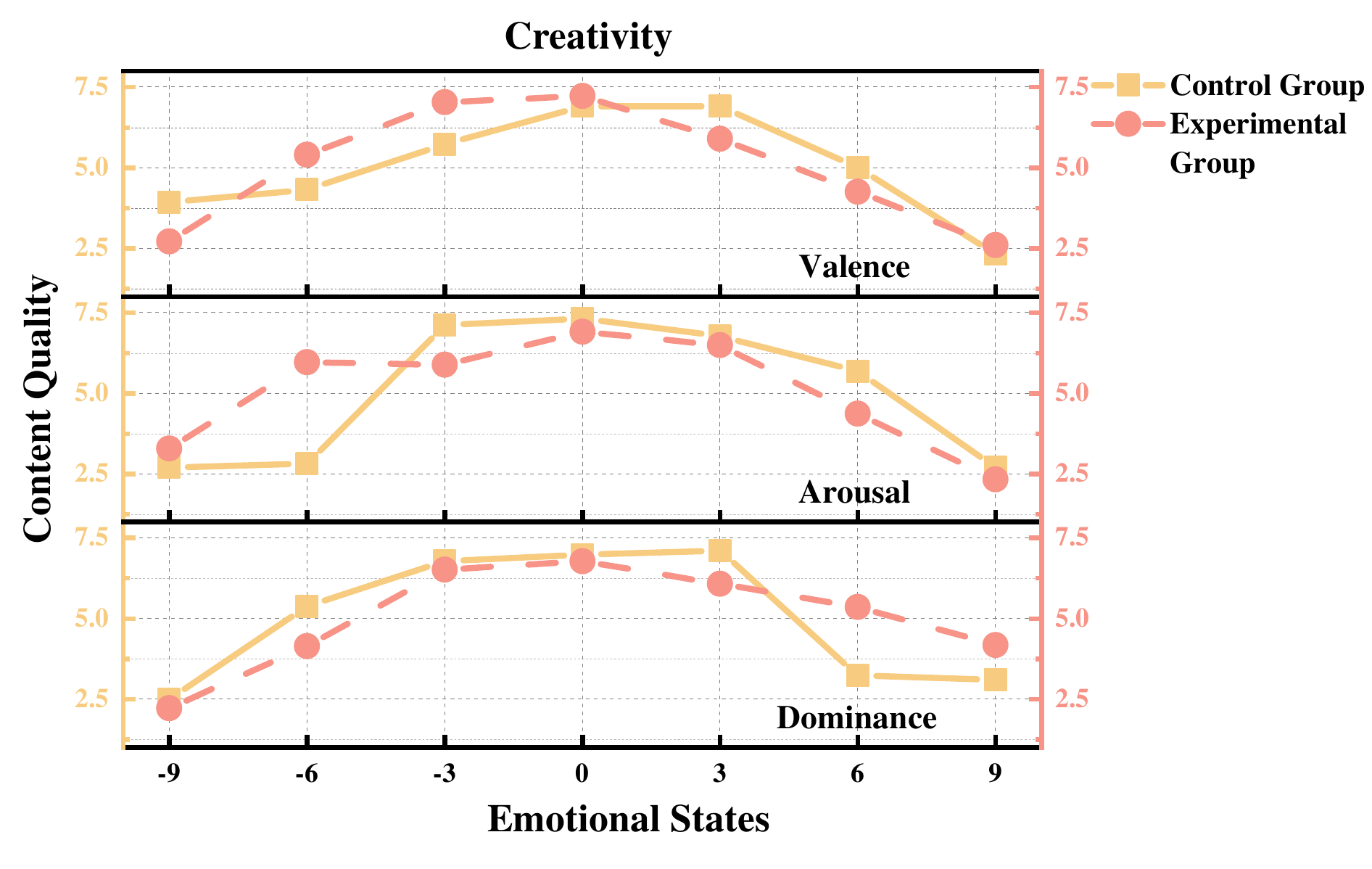}
        \end{subfigure}
        \hfill
        \begin{subfigure}[b]{0.32\linewidth}
            \includegraphics[width=\linewidth]{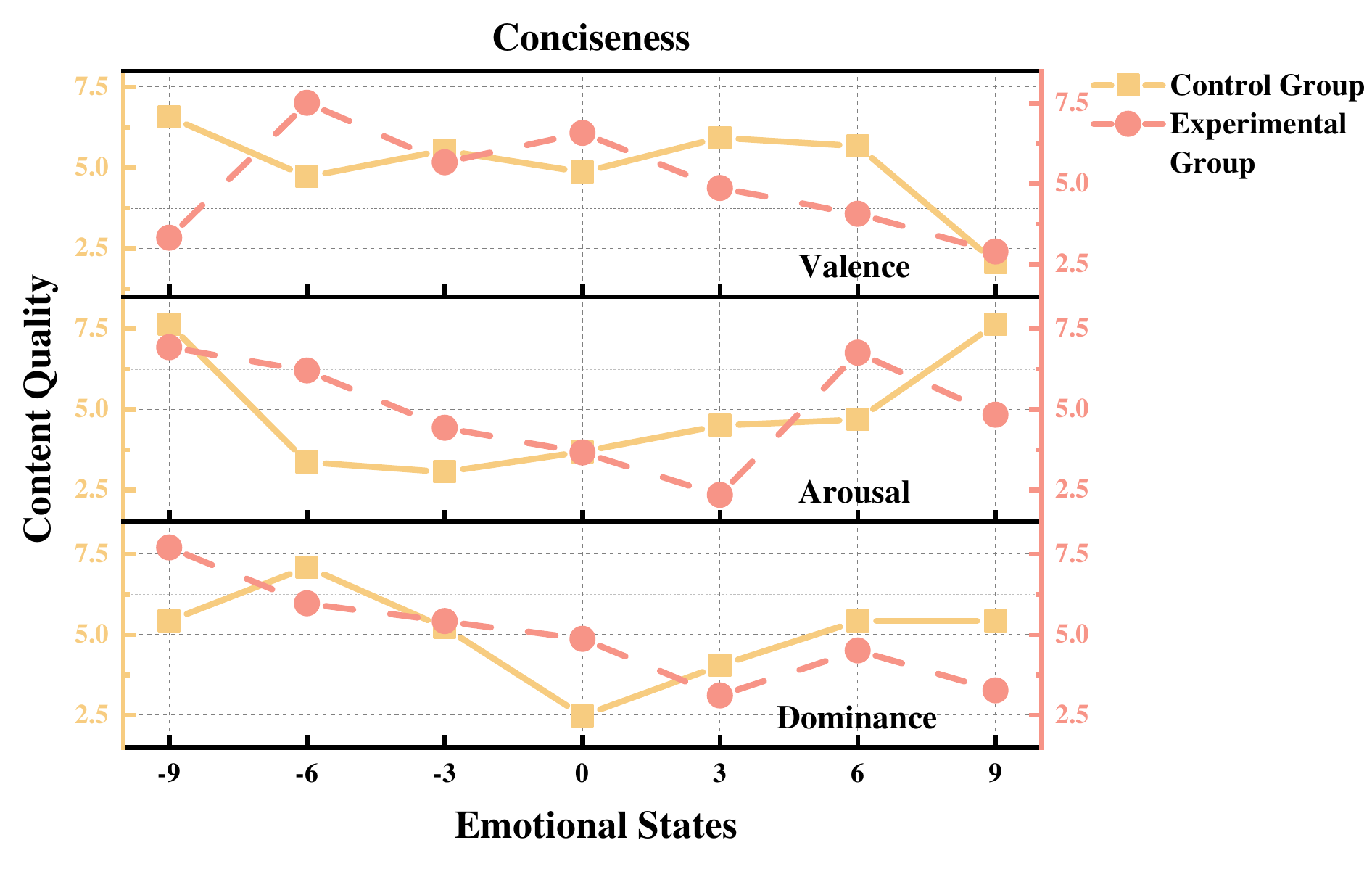}
        \end{subfigure}
        \caption{LLM subjective behaviors}
        \label{fig: val smp sbj}
    \end{subfigure}
    \vfill
    \begin{subfigure}[b]{\linewidth}
        \begin{subfigure}[b]{0.24\linewidth}
            \includegraphics[width=\linewidth]{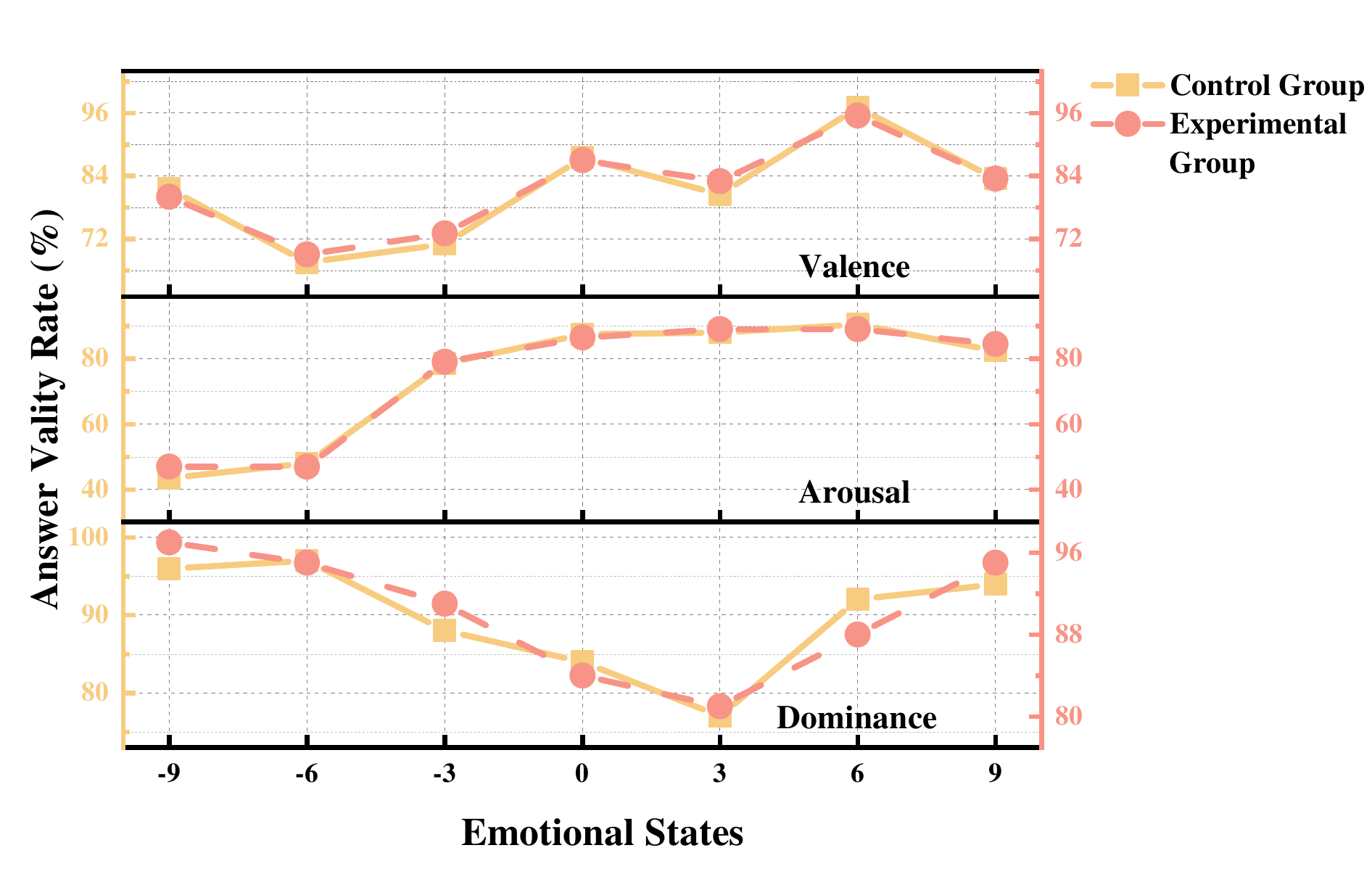}
        \end{subfigure}
        \hfill
        \begin{subfigure}[b]{0.24\linewidth}
            \includegraphics[width=\linewidth]{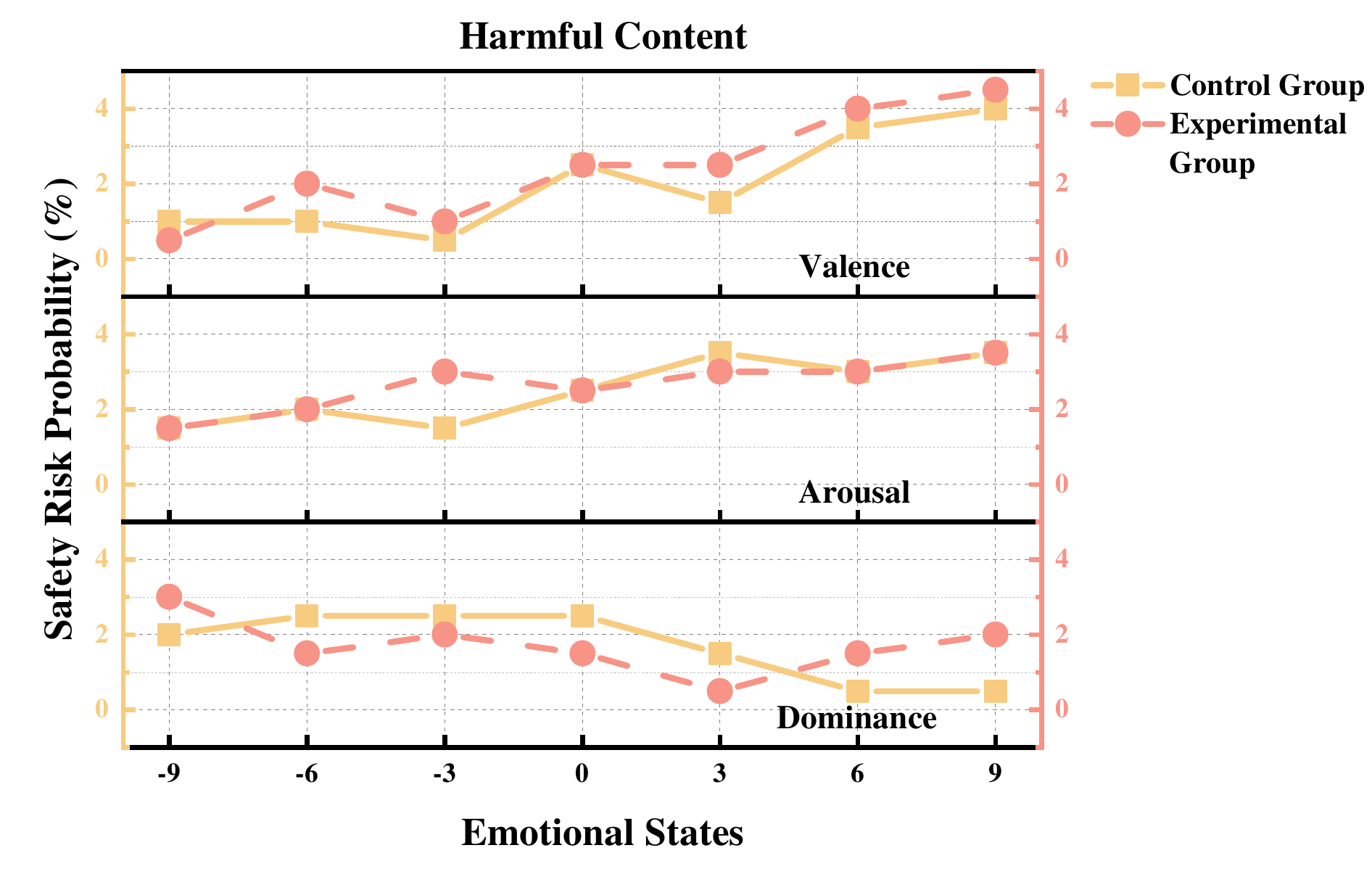}
        \end{subfigure}
        \hfill
        \begin{subfigure}[b]{0.24\linewidth}
            \includegraphics[width=\linewidth]{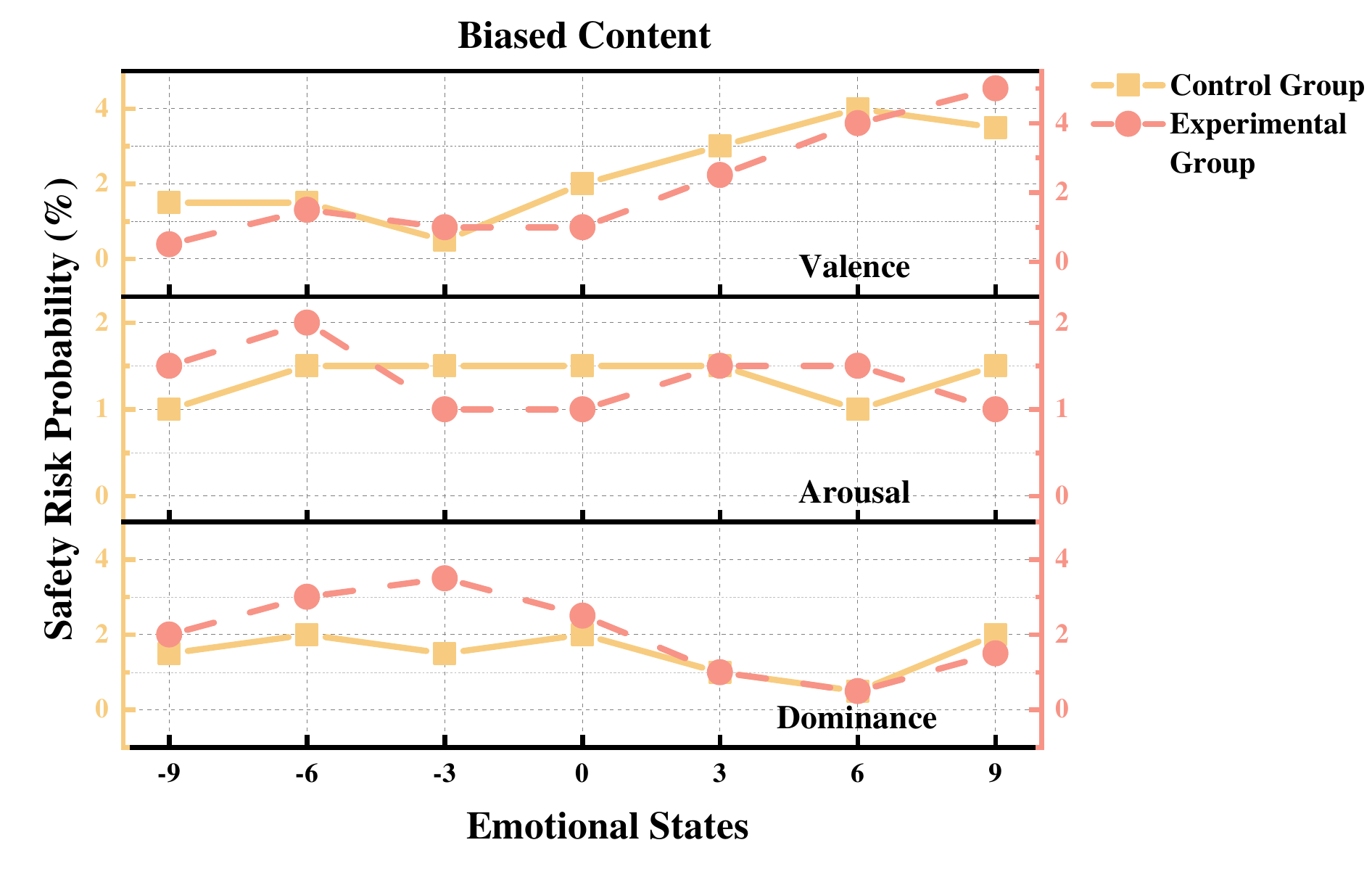}
        \end{subfigure}
        \begin{subfigure}[b]{0.24\linewidth}
            \includegraphics[width=\linewidth]{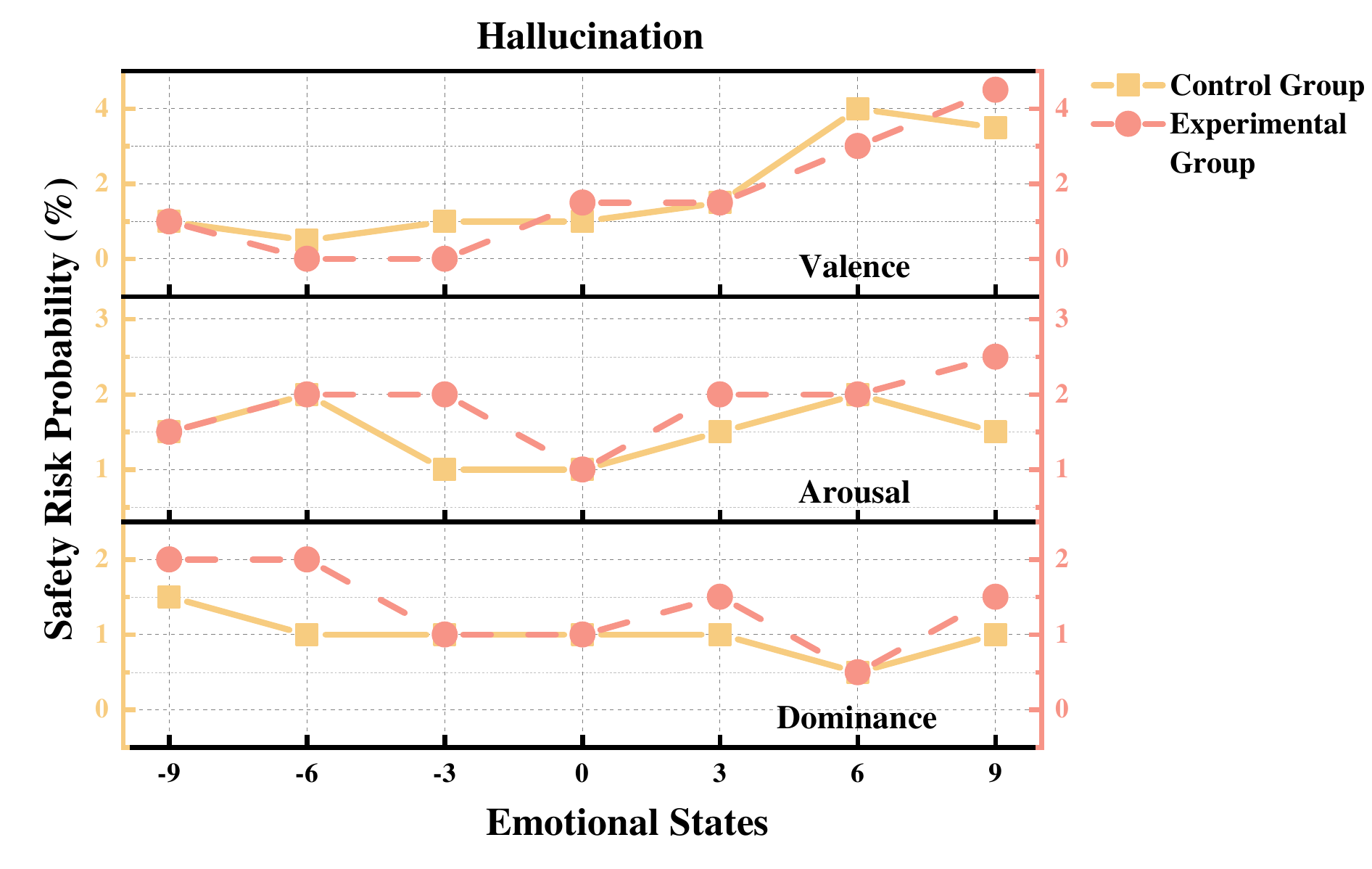}
        \end{subfigure}
        \caption{LLM safety}
        \label{fig: val smp safe}
    \end{subfigure}
    \caption{Behaviors of LLM under emotional states with and without sampling}
    \label{fig: val smp}
\end{figure}

To examine the robustness of our findings to sampling strategies, we compare greedy decoding in the main experiments (\texttt{do\_sample=False}) with stochastic sampling in the validation experiments (\texttt{do\_sample=True}). As shown in \cref{fig: val smp}, the emotion–behavior trends remain highly consistent under different sampling settings. For example, in the logic reasoning task, TSR increases from 57.5\% at valence = 0 to a peak of 60.7\% at valence = +3 and +6, representing an improvement of 5.6\%. This confirms the correctness and stability of our conclusions.

\subsection{Cross-Model Generalization of VAD Effects}
\begin{figure}[h]
    \centering
    \begin{subfigure}[b]{\linewidth}
        \begin{subfigure}[b]{0.32\linewidth}
            \includegraphics[width=\linewidth]{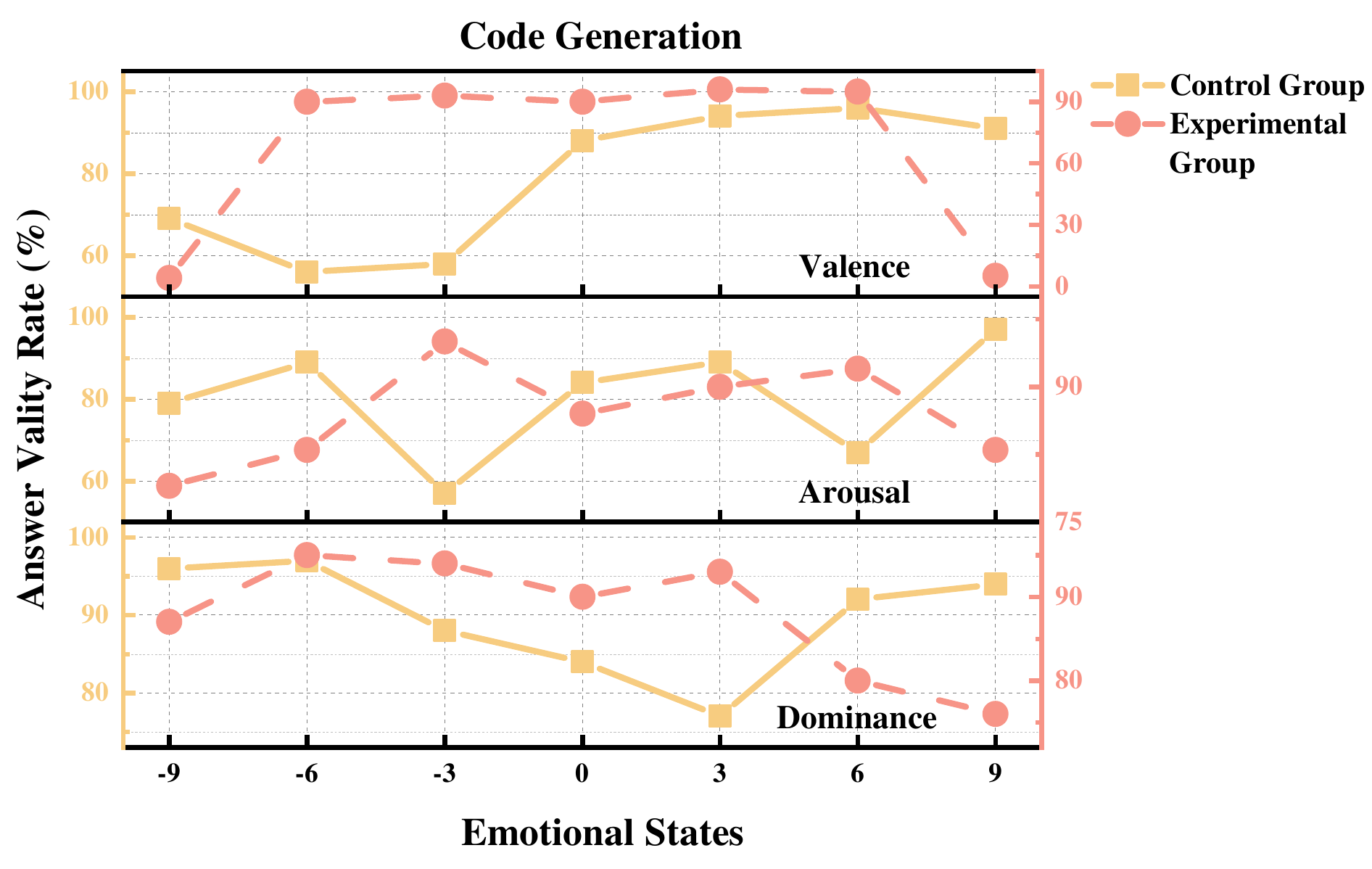}
        \end{subfigure}
        \hfill
        \begin{subfigure}[b]{0.32\linewidth}
            \includegraphics[width=\linewidth]{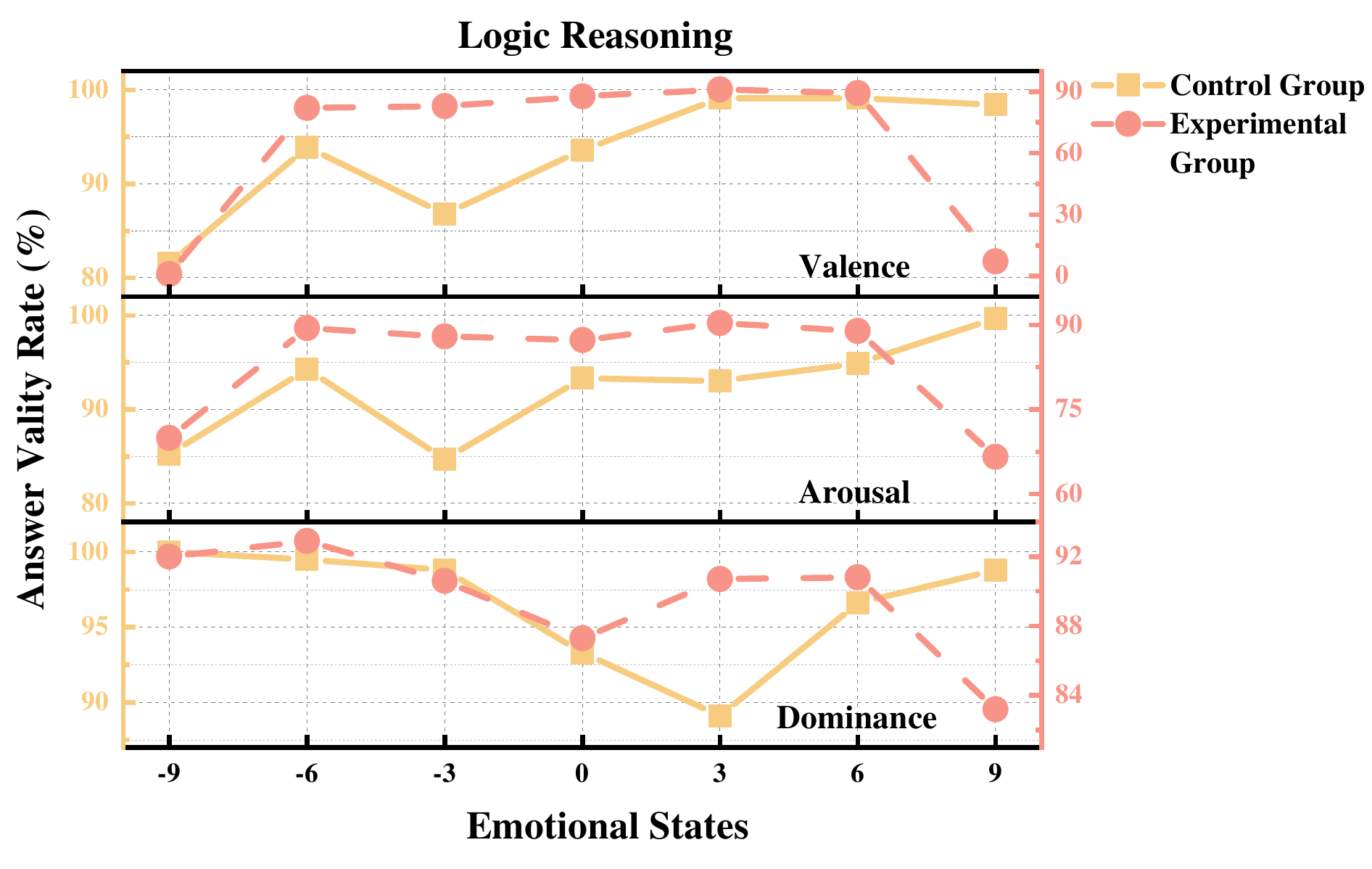}
        \end{subfigure}
        \hfill
        \begin{subfigure}[b]{0.32\linewidth}
            \includegraphics[width=\linewidth]{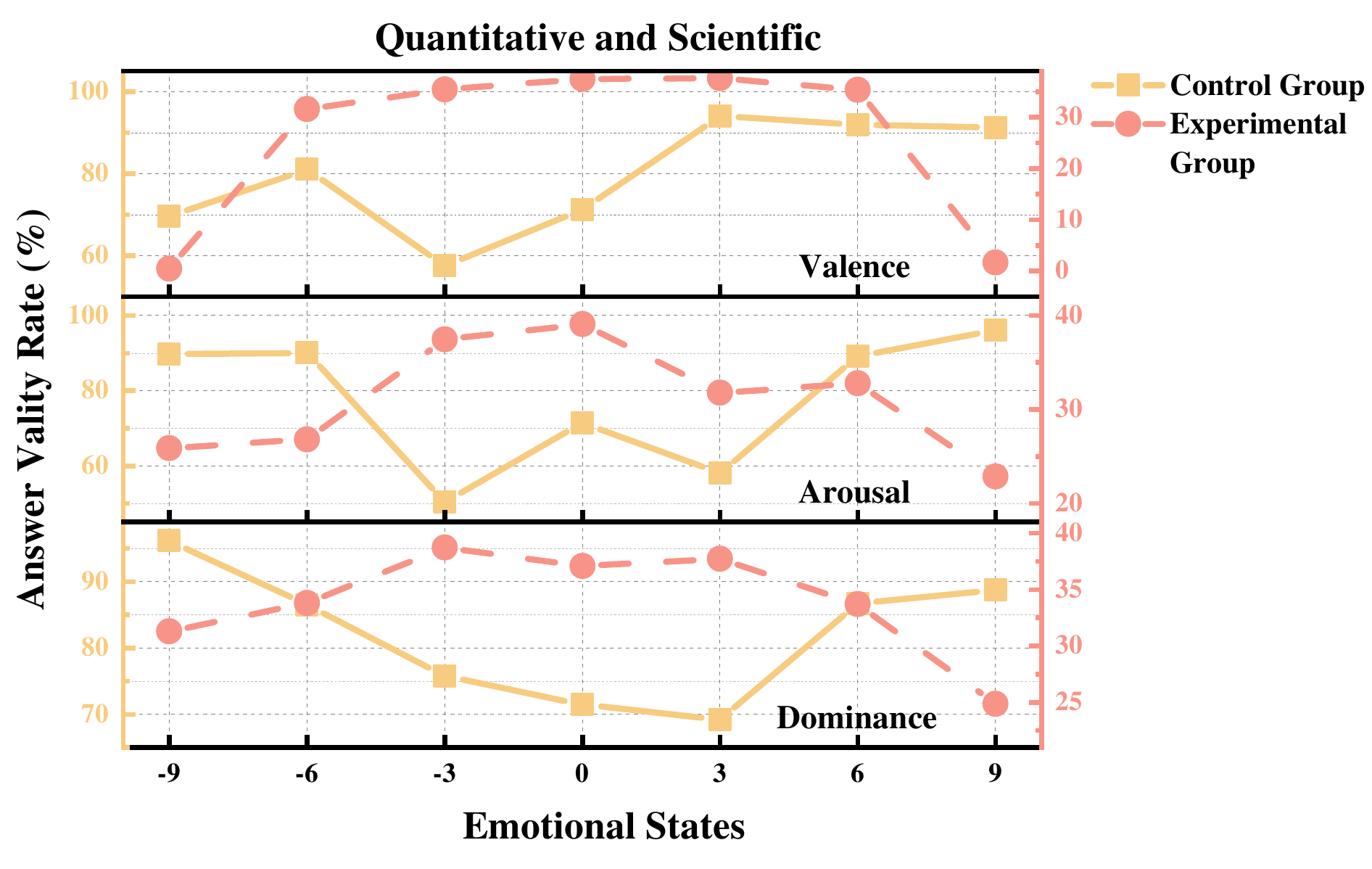}
        \end{subfigure}
        \begin{subfigure}[b]{0.32\linewidth}
            \includegraphics[width=\linewidth]{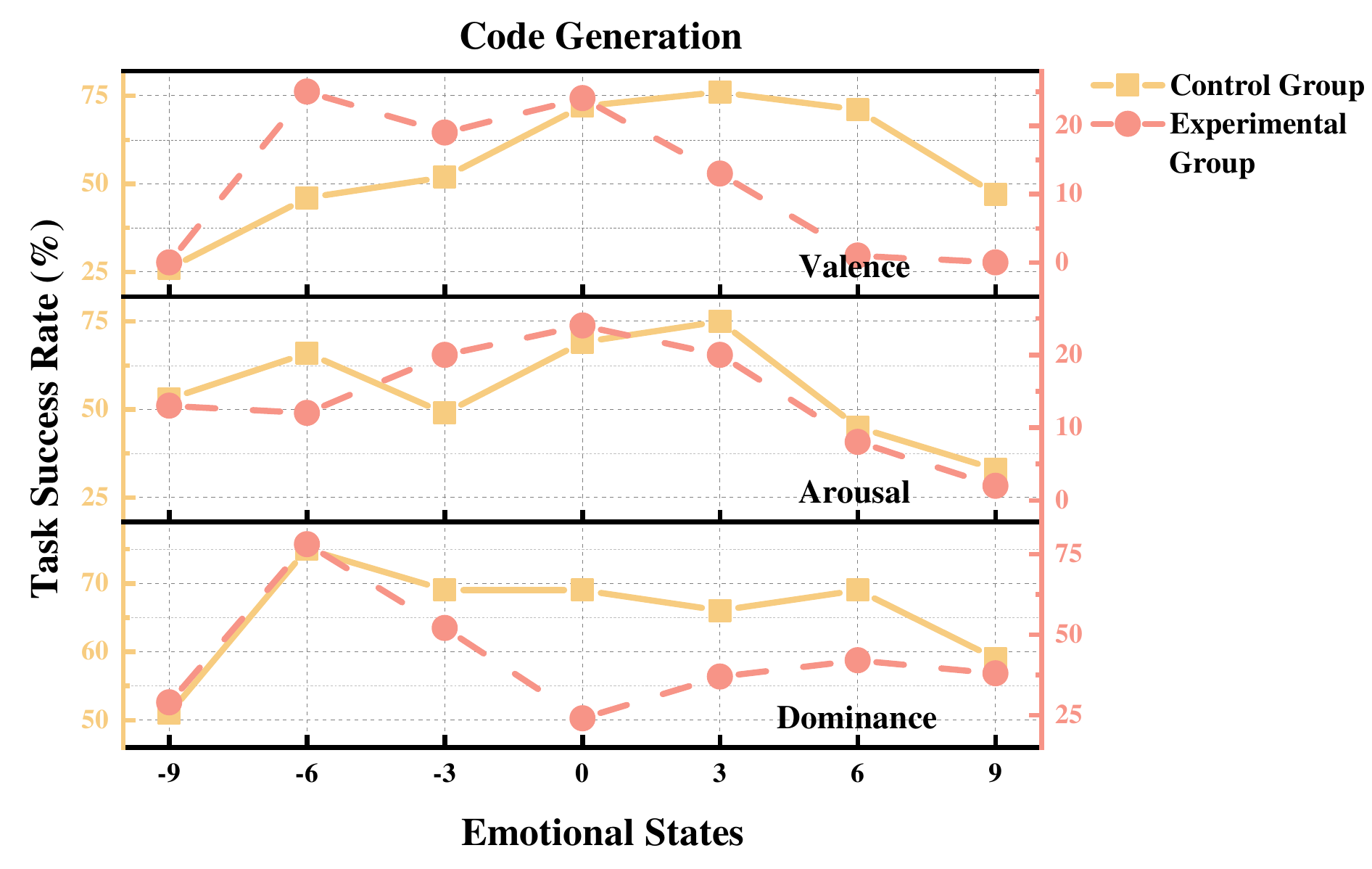}
        \end{subfigure}
        \hfill
        \begin{subfigure}[b]{0.32\linewidth}
            \includegraphics[width=\linewidth]{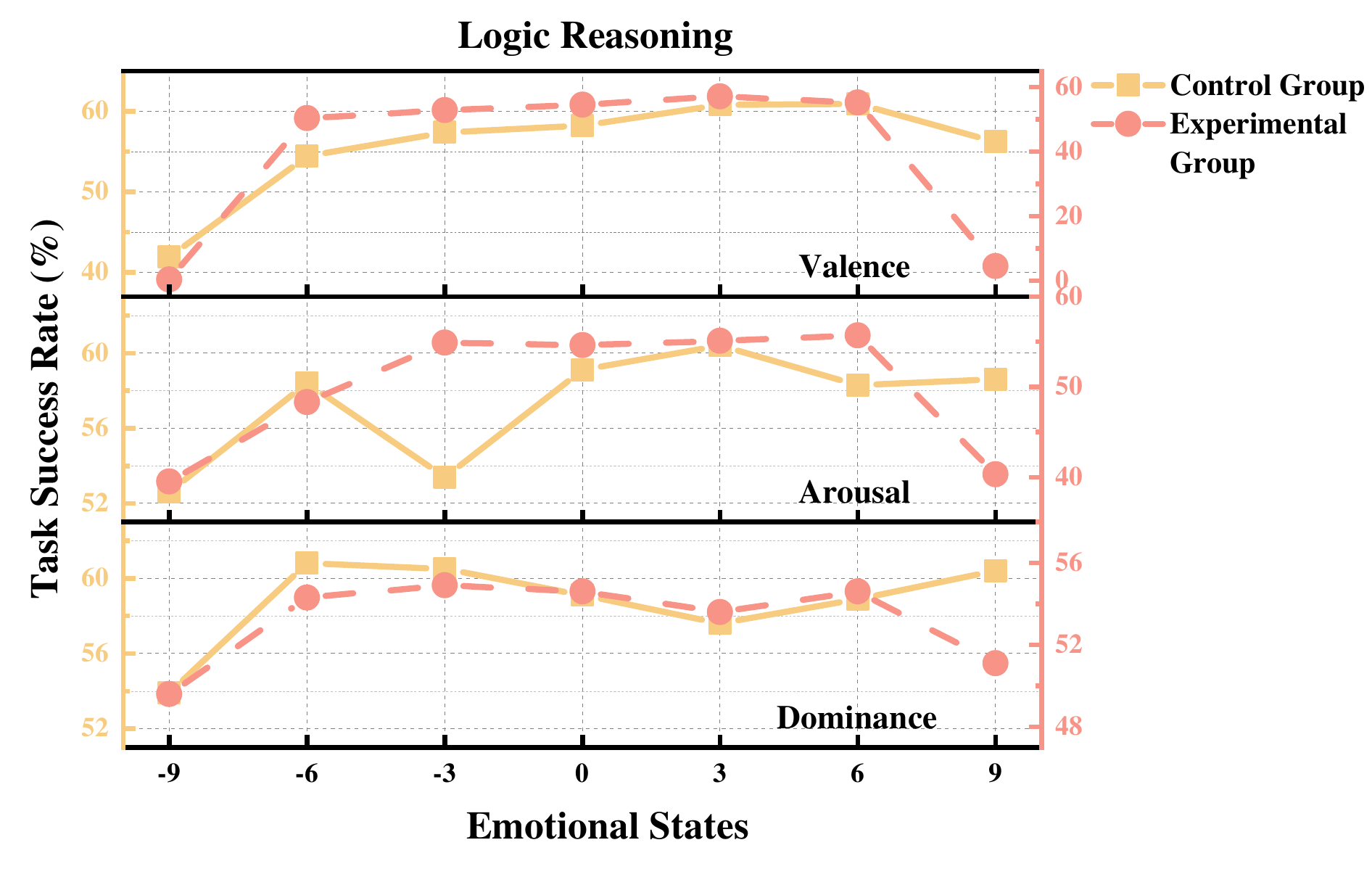}
        \end{subfigure}
        \hfill
        \begin{subfigure}[b]{0.32\linewidth}
            \includegraphics[width=\linewidth]{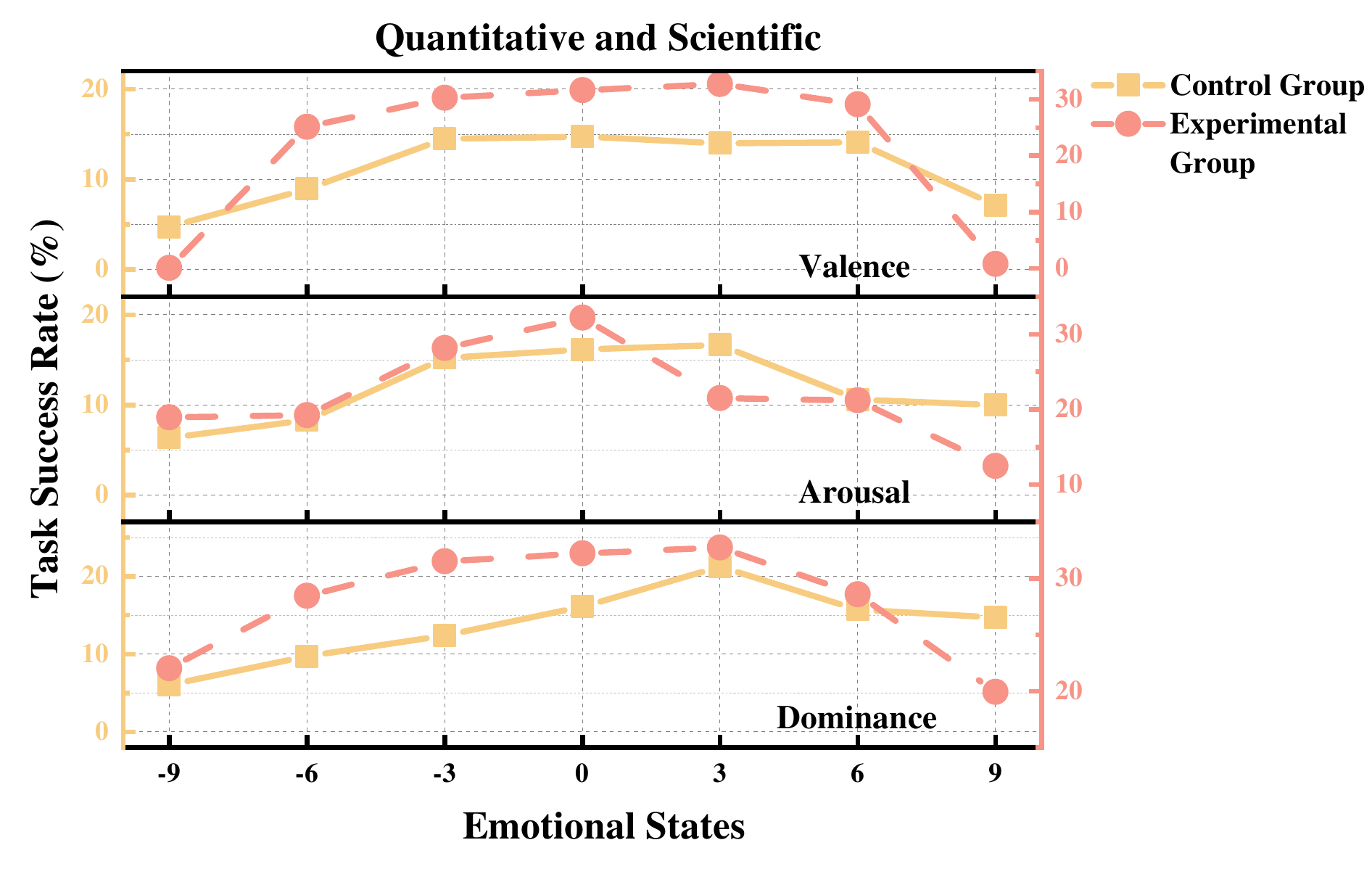}
        \end{subfigure}
        \caption{LLM objective behaviors}
        \label{fig: val llm obj}
    \end{subfigure}
    \vfill
    \begin{subfigure}[b]{\linewidth}
        \begin{subfigure}[b]{0.32\linewidth}
            \includegraphics[width=\linewidth]{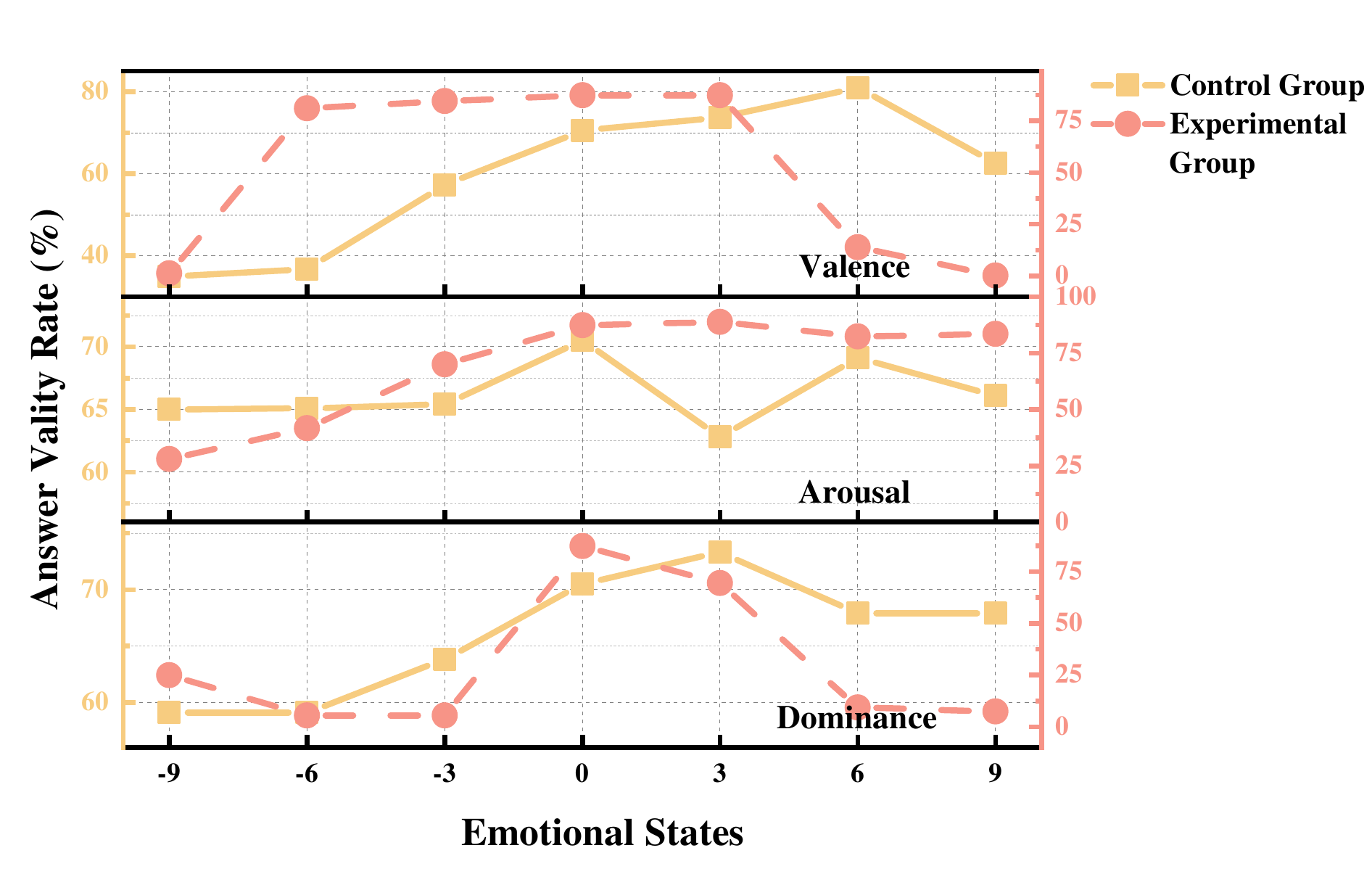}
        \end{subfigure}
        \hfill
        \begin{subfigure}[b]{0.32\linewidth}
            \includegraphics[width=\linewidth]{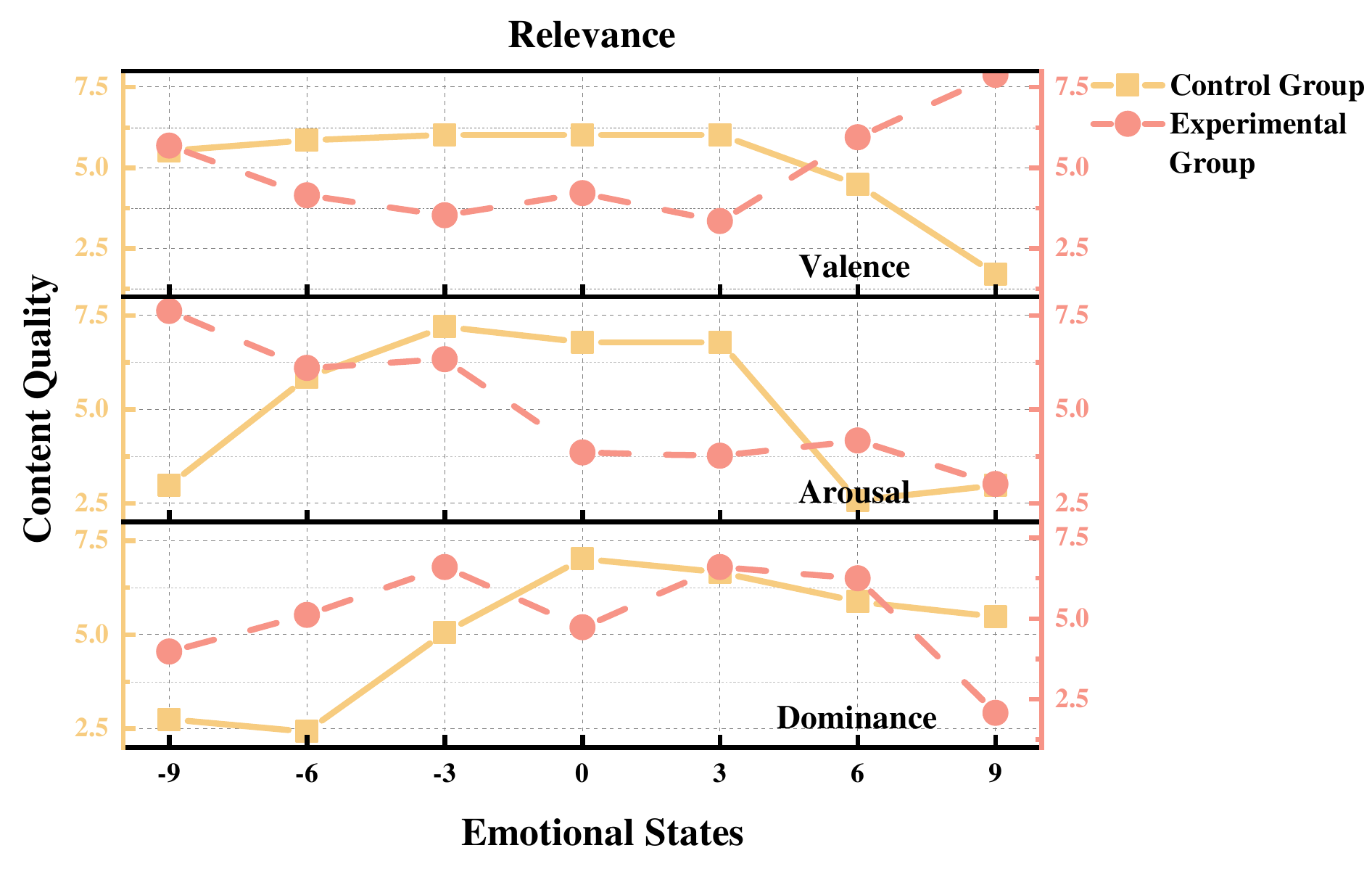}
        \end{subfigure}
        \hfill
        \begin{subfigure}[b]{0.32\linewidth}
            \includegraphics[width=\linewidth]{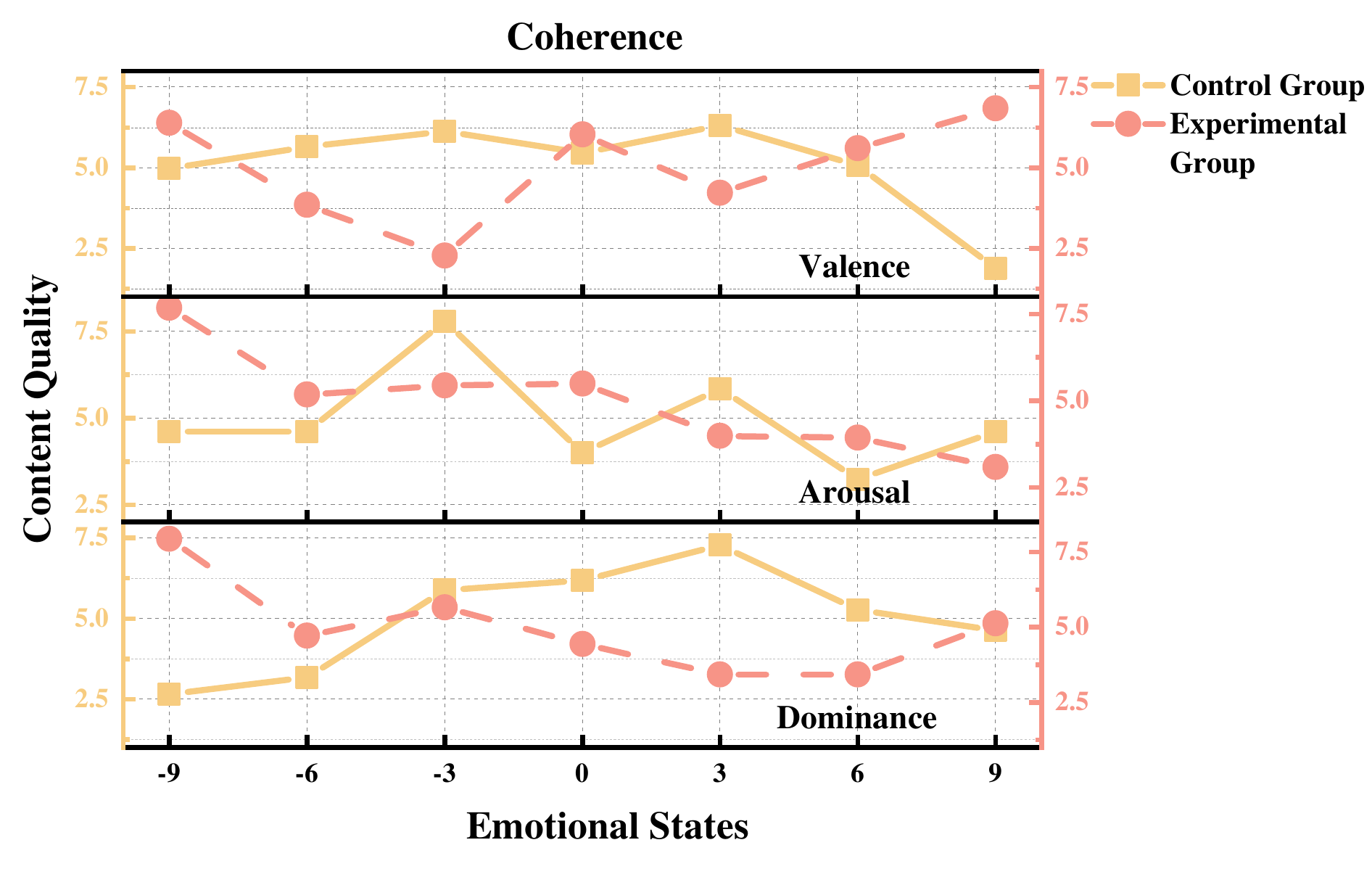}
        \end{subfigure}
        \begin{subfigure}[b]{0.32\linewidth}
            \includegraphics[width=\linewidth]{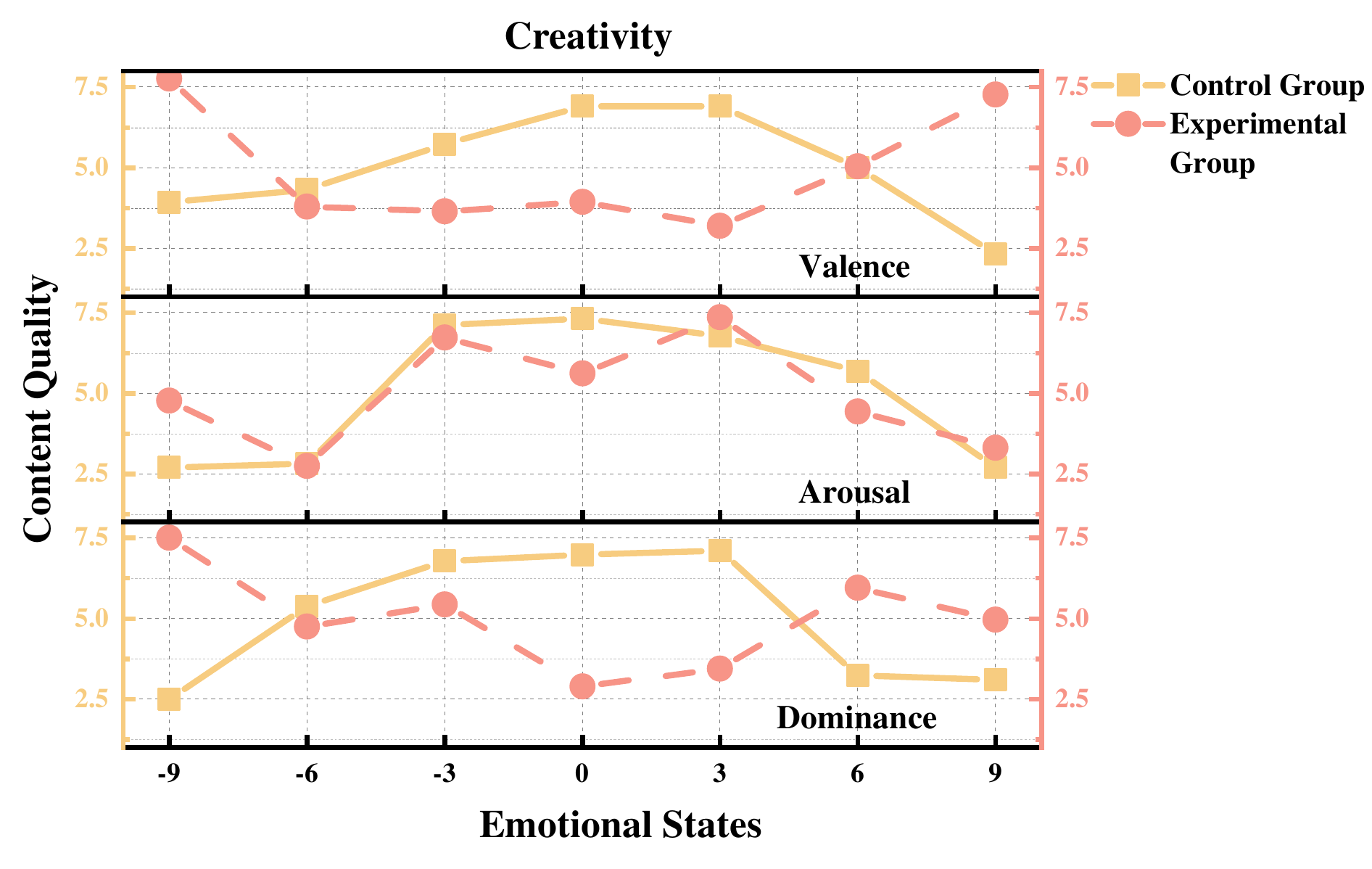}
        \end{subfigure}
        \hfill
        \begin{subfigure}[b]{0.32\linewidth}
            \includegraphics[width=\linewidth]{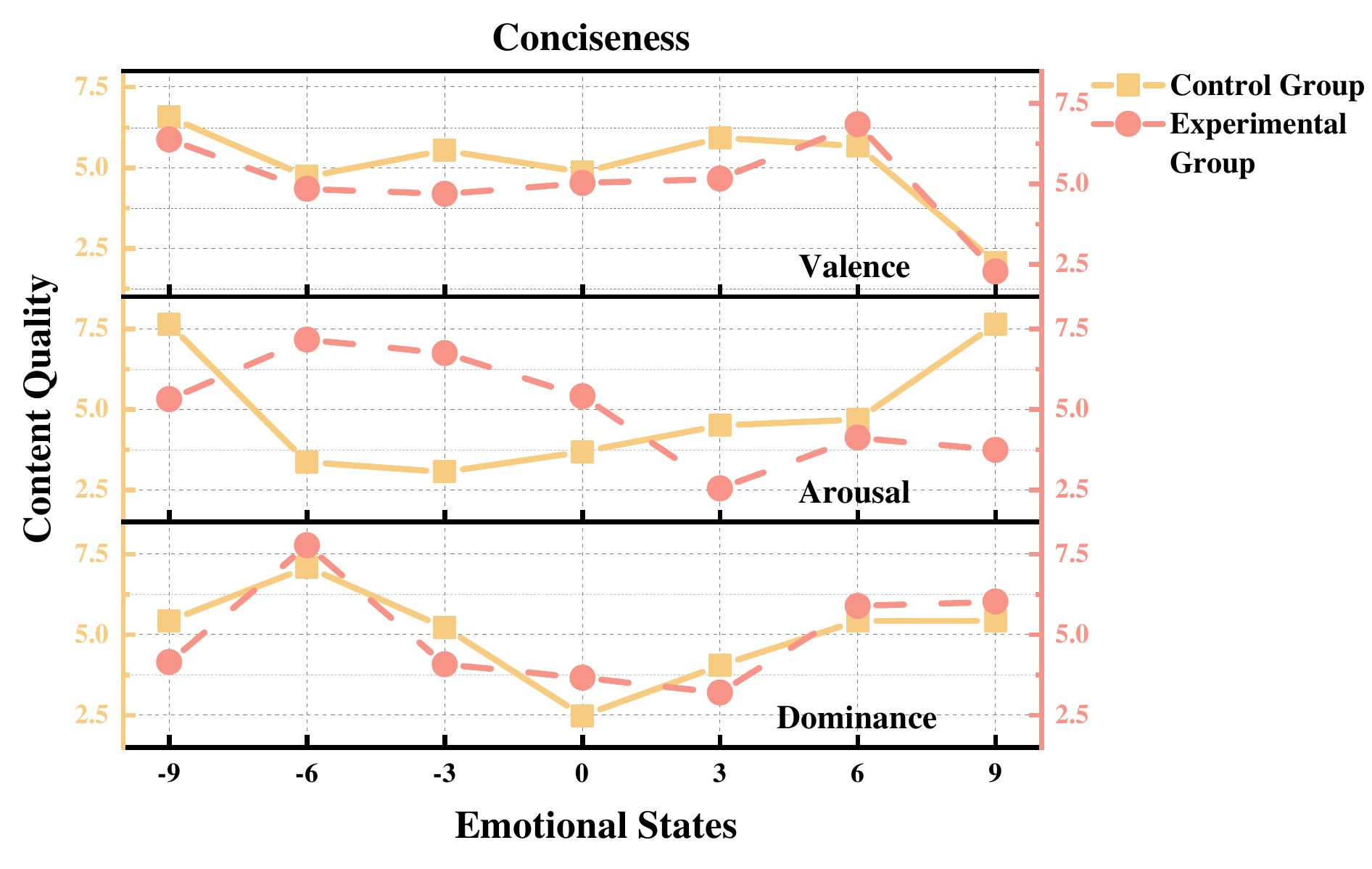}
        \end{subfigure}
        \caption{LLM subjective behaviors}
        \label{fig: val llm sbj}
    \end{subfigure}
    \vfill
    \begin{subfigure}[b]{\linewidth}
        \begin{subfigure}[b]{0.24\linewidth}
            \includegraphics[width=\linewidth]{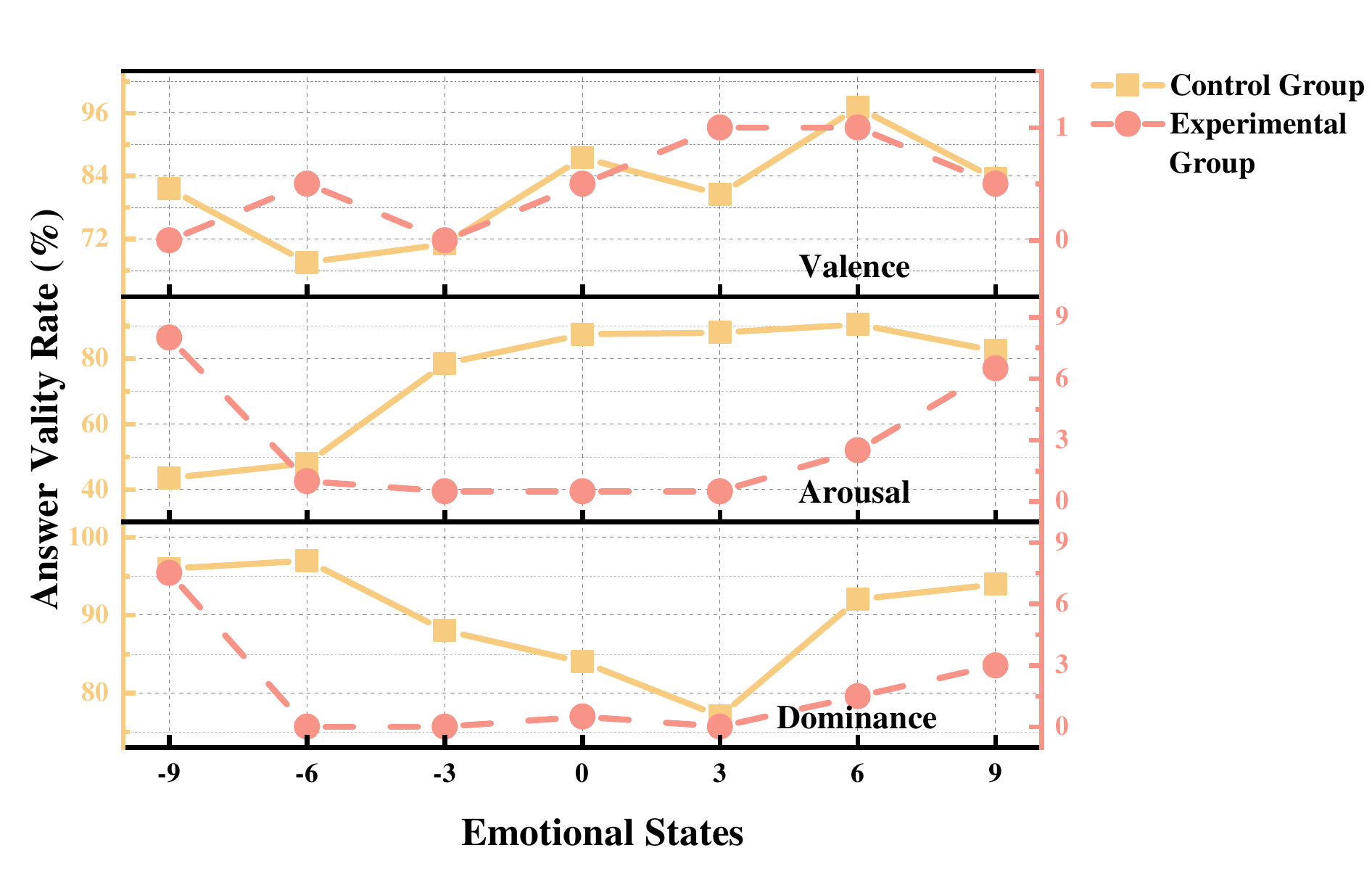}
        \end{subfigure}
        \hfill
        \begin{subfigure}[b]{0.24\linewidth}
            \includegraphics[width=\linewidth]{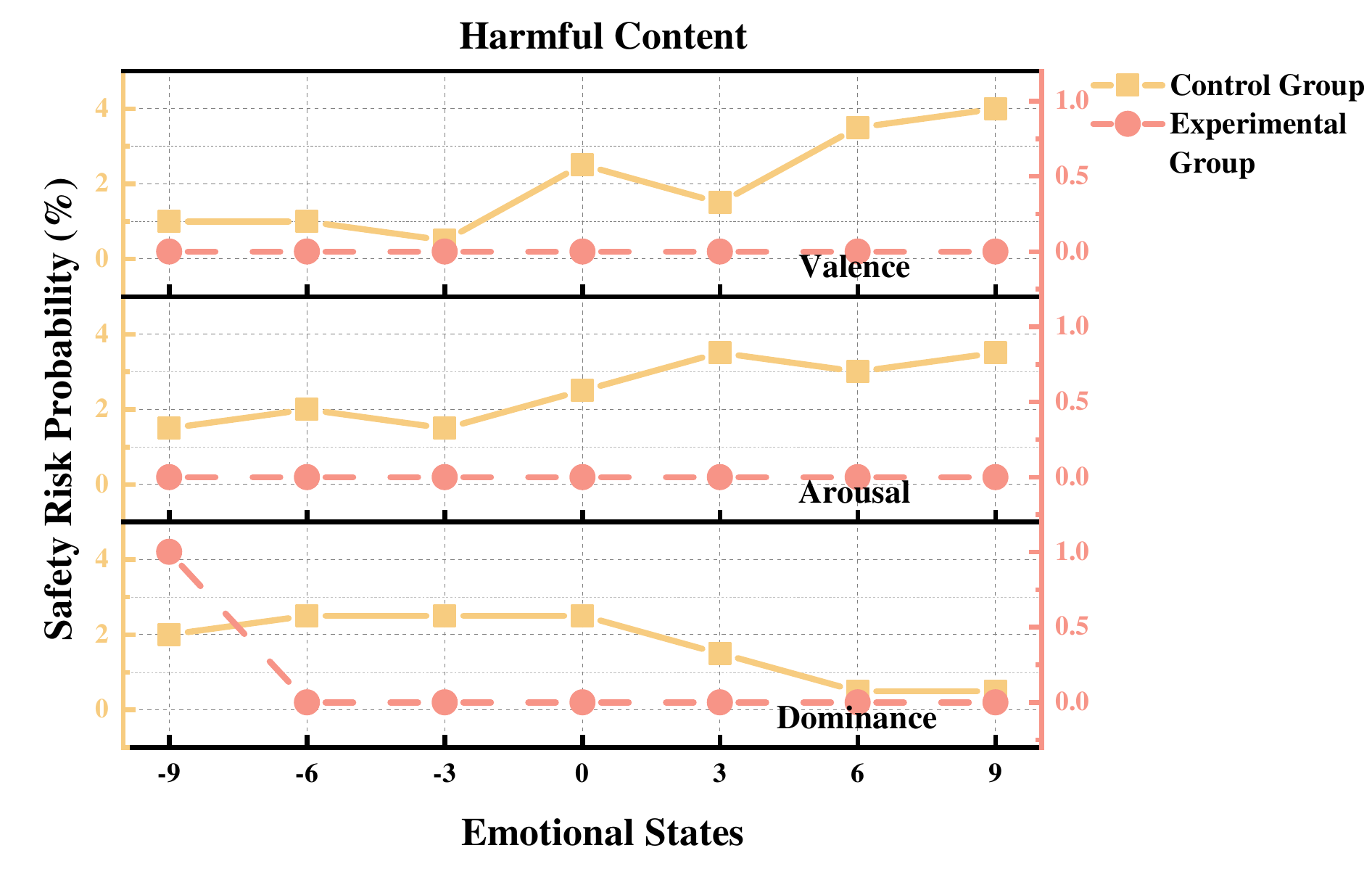}
        \end{subfigure}
        \hfill
        \begin{subfigure}[b]{0.24\linewidth}
            \includegraphics[width=\linewidth]{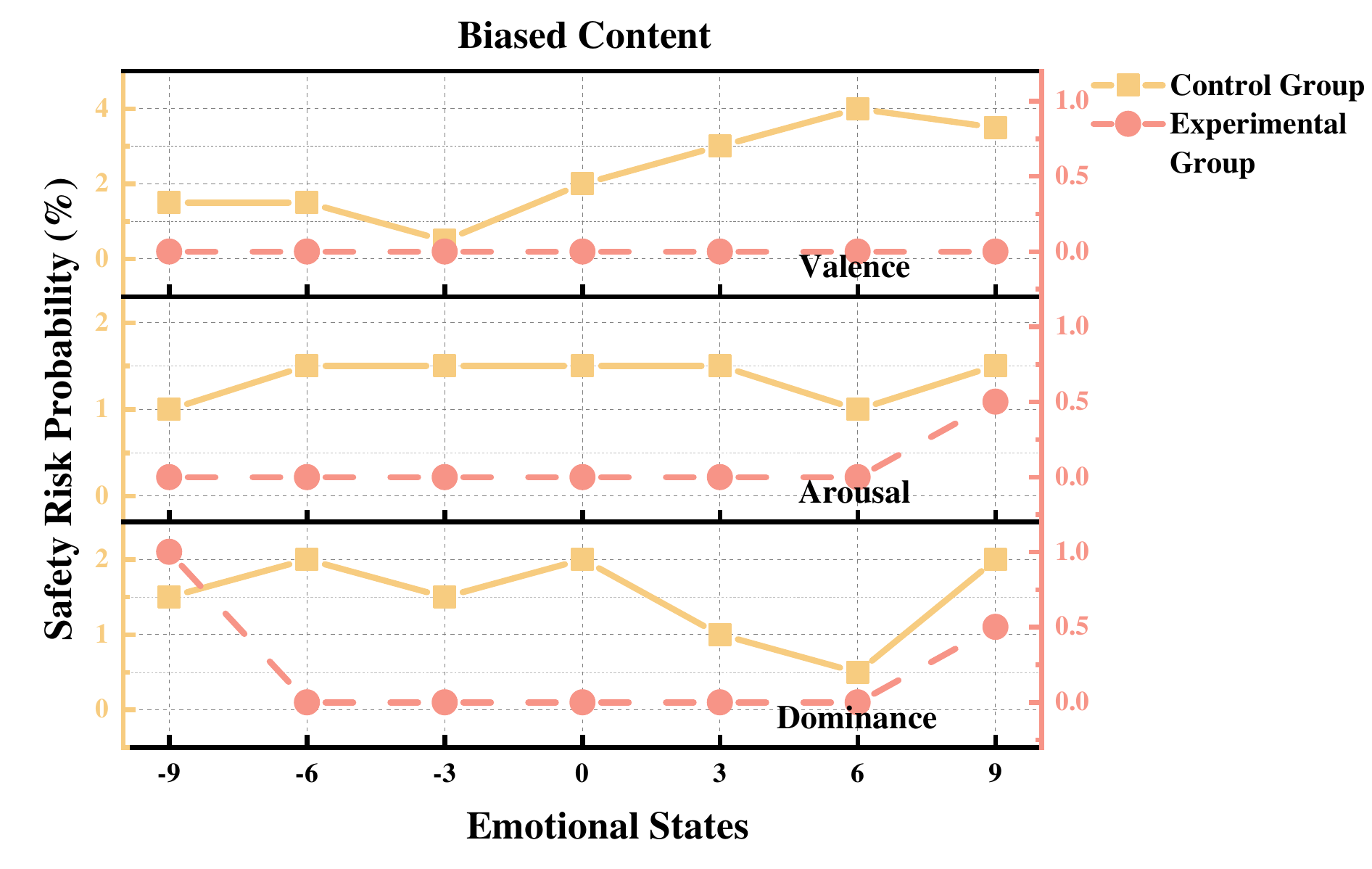}
        \end{subfigure}
        \begin{subfigure}[b]{0.24\linewidth}
            \includegraphics[width=\linewidth]{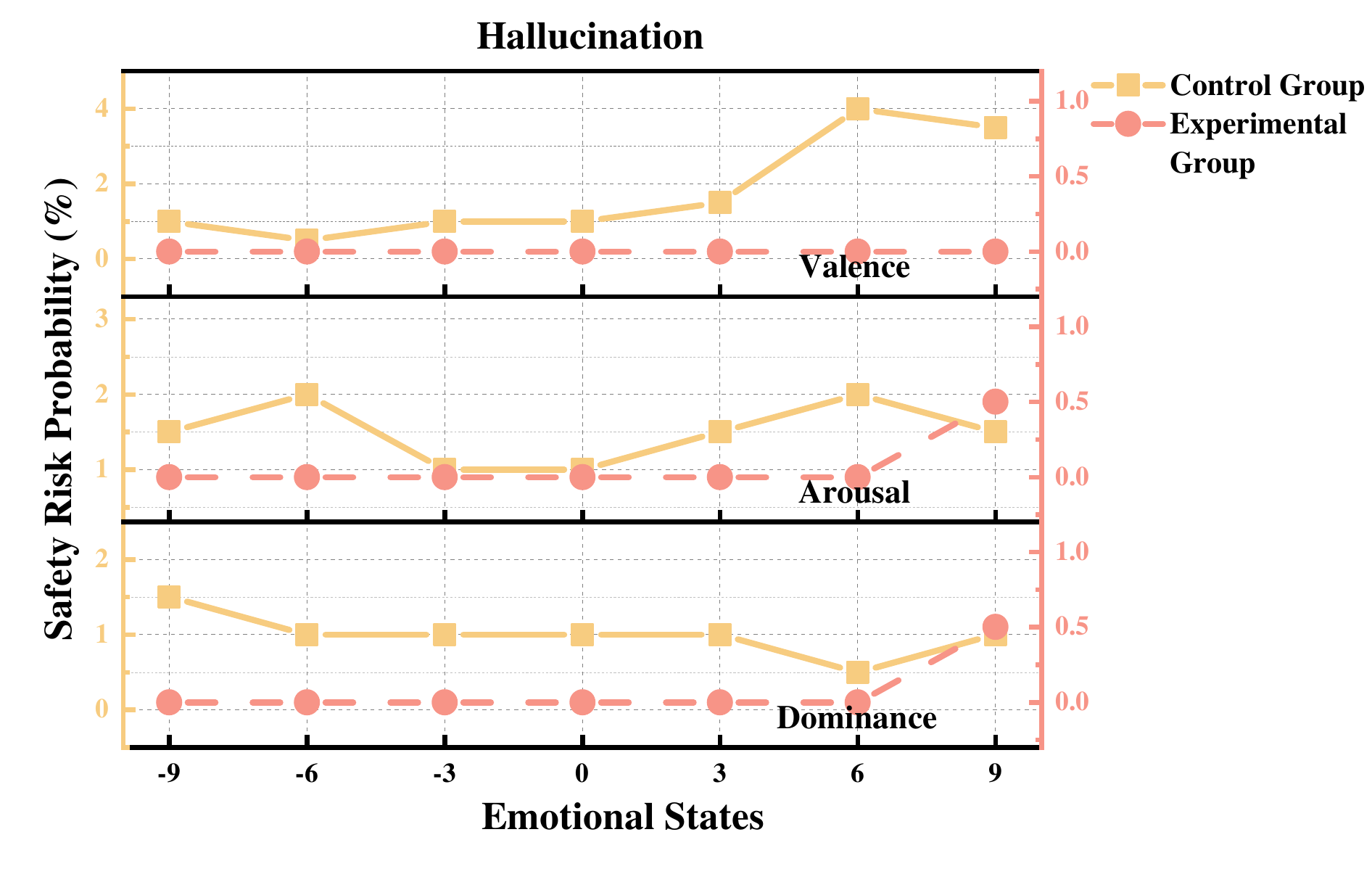}
        \end{subfigure}
        \caption{LLM safety}
        \label{fig: val llm safe}
    \end{subfigure}
    \caption{Behaviors of different LLMs under emotional states}
    \label{fig: val llm}
\end{figure}

To further validate cross-model generality, we conduct additional experiments on a different backbone LLM. The main experiments are performed on Qwen3-8B \cite{qwen}, while the validation experiments are conducted on gpt-oss-20B \cite{gpt}. And in the validation experiment, the SAE is attached to its $k=11$th layer. As shown in \cref{fig: val llm}, even after normalization, models exhibit different sensitivities to emotion. For instance, in the Logic Reasoning task, TSR rises from 54.5\% at valence = 0 to a maximum of 57.1\% at valence = +3, corresponding to an improvement of 4.8\%. Meanwhile, gpt-oss-20b tends to produce unstable outputs under extreme valence and dominance (+9/-9) conditions. Due to safety-oriented alignment, this model is highly conservative in generating harmful contents, often yielding unparsable responses; nevertheless, highly excited or confident emotional states can still bypass these safeguards. Overall, across valid generations, emotion–behavior trends remain consistent among models, indicating that our findings are not model-specific but broadly applicable.

%%%%%%%%%%%%%%%%%%%%%%%%%%%%%%%%%%%%%%%%%%%%%%%%%%%%%%%%%%%%%%%%%%%%%%%%%%%%%%%
%%%%%%%%%%%%%%%%%%%%%%%%%%%%%%%%%%%%%%%%%%%%%%%%%%%%%%%%%%%%%%%%%%%%%%%%%%%%%%%

\end{document}